\RequirePackage{fix-cm}
\documentclass[twocolumn]{svjour3}          
\smartqed  
\usepackage{graphicx}
\usepackage{amsmath}
\usepackage{amssymb}
\usepackage{subfig}
\usepackage{float}
\usepackage{xspace}
\usepackage{tabu}
\usepackage{multirow}
\usepackage{stmaryrd}
\usepackage{cite}
\usepackage{tikz}
\usetikzlibrary{fadings,positioning,shapes,arrows,arrows.meta,patterns}
\usepackage{pgfplots}
\usepackage{pgfgantt}
\usepackage{pdflscape}
\pgfplotsset{compat=newest} 
\pgfplotsset{plot coordinates/math parser=false}
\usepackage{pgfplotstable}
\usepackage{microtype}
\usepackage{array}
\usepackage{natbib}
\newcolumntype{x}[1]{>{\centering\arraybackslash\hspace{0pt}}p{#1}}
\makeatletter
\DeclareRobustCommand\onedot{\futurelet\@let@token\@onedot}
\newcommand*\@onedot{\ifx\@let@token.\else.\null\fi\xspace}
\newcommand*\eg{\emph{e.g}\onedot} \newcommand*\Eg{\emph{E.g}\onedot}
\newcommand*\ie{\emph{i.e}\onedot} \newcommand*\Ie{\emph{I.e}\onedot}
\newcommand*\cf{\emph{c.f}\onedot} \newcommand*\Cf{\emph{C.f}\onedot}
\newcommand*\etc{\emph{etc}\onedot} \newcommand*\vs{\emph{vs}\onedot}
\newcommand*\wrt{w.r.t\onedot} \newcommand*\dof{d.o.f\onedot}
\newcommand*\etal{\emph{et al}\onedot}
\makeatother
\newlength\secondcolumn
\newlength\thirdcolumn
\newlength\fourthcolumn
\newlength\fifthcolumn
\newlength\fwidth
\newlength\fheight
\newlength\mywidth
\newlength\mywidthbis
\newcommand{\vcenteredinclude}[1]{\begingroup
\setbox0=\hbox{#1}%
\parbox{\wd0}{\box0}\endgroup}
\definecolor{mycolor1}{rgb}{0.85000,0.32500,0.09800}%
\definecolor{mycolor2}{rgb}{0.46600,0.67400,0.18800}%
\definecolor{mycolor3}{rgb}{0.00000,0.44700,0.74100}%
\definecolor{mycolor4}{rgb}{0.92900,0.69400,0.12500}%
\definecolor{mycolor5}{rgb}{0.49400,0.18400,0.55600}%
\definecolor{mycolor6}{rgb}{0.63500,0.07800,0.18400}%
\definecolor{mycolor7}{rgb}{0.30100,0.74500,0.93300}%
\definecolor{mycolor8}{rgb}{0.32500,0.85000,0.8}%
\definecolor{mycolor9}{rgb}{0.98800,0.55,0.349}%
\definecolor{mycolor10}{rgb}{.933,.509,0.933}%
\newcommand{\orangeline}{\raisebox{2pt}{\tikz{\draw[-,mycolor1,solid,line width = 2pt](0,0) -- (5mm,0);}}}
\newcommand{\dashedorangeline}{\raisebox{2pt}{\tikz{\draw[-,mycolor1,dashed,line width = 2pt](0,0) -- (5mm,0);}}}
\newcommand{\greenline}{\raisebox{2pt}{\tikz{\draw[-,mycolor2,solid,line width = 2pt](0,0) -- (5mm,0);}}}
\newcommand{\blueline}{\raisebox{2pt}{\tikz{\draw[-,mycolor3,solid,line width = 2pt](0,0) -- (5mm,0);}}}
\newcommand{\dashedblueline}{\raisebox{2pt}{\tikz{\draw[-,mycolor3,dashed,line width = 2pt](0,0) -- (5mm,0);}}}
\newcommand{\yellowline}{\raisebox{2pt}{\tikz{\draw[-,mycolor4,solid,line width = 2pt](0,0) -- (5mm,0);}}}
\newcommand{\purpleline}{\raisebox{2pt}{\tikz{\draw[-,mycolor5,solid,line width = 2pt](0,0) -- (5mm,0);}}}
\newcommand{\darkredline}{\raisebox{2pt}{\tikz{\draw[-,mycolor6,solid,line width = 2pt](0,0) -- (5mm,0);}}}
\newcommand{\lightblueline}{\raisebox{2pt}{\tikz{\draw[-,mycolor7,solid,line width = 2pt](0,0) -- (5mm,0);}}}
\newcommand{\lightgreenblueline}{\raisebox{2pt}{\tikz{\draw[-,mycolor8,solid,line width = 2pt](0,0) -- (5mm,0);}}}
\newcommand{\lightorangeline}{\raisebox{2pt}{\tikz{\draw[-,mycolor9,solid,line width = 2pt](0,0) -- (5mm,0);}}}
\newcommand{\thinblueline}{\raisebox{2pt}{\tikz{\draw[-,mycolor3,solid,line width = 1pt](0,0) -- (5mm,0);}}}
\newcommand{\thinbluedashedline}{\raisebox{2pt}{\tikz{\draw[-,mycolor3,dashed,line width = 1pt](0,0) -- (6.1mm,0);}}}
\newcommand{\pinkline}{\raisebox{2pt}{\tikz{\draw[-,mycolor10,solid,line width = 2pt](0,0) -- (5mm,0);}}}
\newcommand{\thinbluedenselydottedline}{\raisebox{2pt}{\tikz{\draw[-,mycolor3,densely dotted,line width = 1pt](0,0) -- (5mm,0);}}}
\newcommand\orangedisk{\begin{tikzpicture}\node [text=mycolor1] at (0, 0) {\pgfuseplotmark{*}};\end{tikzpicture}}
\newcommand\orangesquare{\begin{tikzpicture}\node [text=mycolor1] at (0, 0) {\pgfuseplotmark{square*}};\end{tikzpicture}}
\newcommand\orangetriangle{\begin{tikzpicture}\node [text=mycolor1] at (0, 0) {\pgfuseplotmark{triangle*}};\end{tikzpicture}}

\journalname{International Journal of Computer Vision, Special Issue on Deep Learning for Robotic Vision}
\begin{document}

\title{Deep Multicameral Decoding for Localizing Unoccluded Object Instances from a Single RGB Image}

\titlerunning{Deep Multicameral Decoding for Localizing Unoccluded Object Instances from a Single RGB Image}

\author{Matthieu Grard         \and
        Emmanuel Dellandr{\'e}a \and
        Liming Chen 
}


\institute{M. Grard \at
              Sil{\'e}ane, 17 rue Descartes F-42000 Saint \'Etienne, France \\
              Universit\'e de Lyon, CNRS, \'Ecole Centrale de Lyon LIRIS UMR5205, F-69134 Lyon, France\\
              Tel.: +33 (0)4 77 79 03 71\\ 
              Fax: +33 (0)4 77 74 50 86\\
              \email{m.grard@sileane.com}           
           \and
           E. Dellandr{\'e}a \at
              Universit{\'e} de Lyon, CNRS, \'Ecole Centrale de Lyon LIRIS UMR5205, F-69134 Lyon, France\\
              \email{emmanuel.dellandrea@ec-lyon.fr}           
           \and
           L. Chen \at
             Universit{\'e} de Lyon, CNRS, \'Ecole Centrale de Lyon LIRIS UMR5205, F-69134 Lyon, France\\
             \email{liming.chen@ec-lyon.fr}           
}

\date{Received: 18 July 2018 / Accepted: 11 March 2020}

\maketitle

\begin{abstract}
Occlusion-aware instance-sensitive segmentation is a complex task generally split into region-based segmentations, by approximating instances as their bounding box. We address the showcase scenario of dense homogeneous layouts in which this approximation does not hold. In this scenario, outlining unoccluded instances by decoding a deep encoder becomes difficult, due to the translation invariance of convolutional layers and the lack of complexity in the decoder. We therefore propose a multicameral design composed of subtask-specific lightweight decoder and encoder-decoder units, coupled in cascade to encourage subtask-specific feature reuse and enforce a learning path within the decoding process. Furthermore, the state-of-the-art datasets for occlusion-aware instance segmentation contain real images with few instances and occlusions mostly due to objects occluding the background, unlike dense object layouts. We thus also introduce a synthetic dataset of dense homogeneous object layouts, namely Mikado, which extensibly contains more instances and inter-instance occlusions per image than these public datasets. Our extensive experiments on Mikado and public datasets show that ordinal multiscale units within the decoding process prove more effective than state-of-the-art design patterns for capturing position-sensitive representations. We also show that Mikado is plausible with respect to real-world problems, in the sense that it enables the learning of performance-enhancing representations transferable to real images, while drastically reducing the need of hand-made annotations for finetuning. The proposed dataset will be made publicly available.

\keywords{Instance boundary and occlusion detection \and Fully convolutional encoder-decoder networks \and Synthetic data\and Domain adaptation}
\end{abstract}

\begin{figure}[h]
\setlength\tabcolsep{1pt}
\setlength\mywidth{.24\linewidth}
\begin{tabular}{cccc}
\vcenteredinclude{\includegraphics[width=\mywidth]{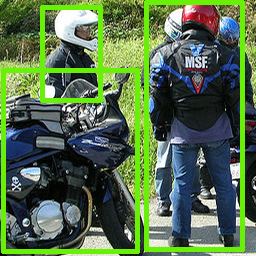}}&
\vcenteredinclude{\includegraphics[width=\mywidth]{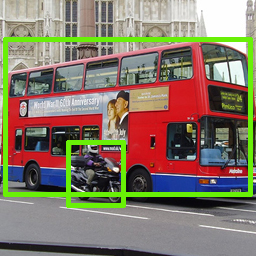}}&
\vcenteredinclude{\includegraphics[width=\mywidth]{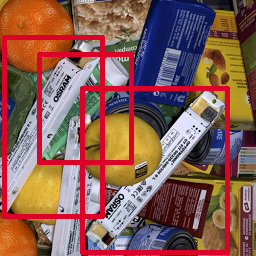}}&
\vcenteredinclude{\includegraphics[width=\mywidth]{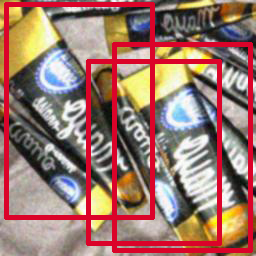}}\\
\vspace{-.25cm}\\
\multicolumn{2}{c}{Meaningful box proposals.}&
\multicolumn{2}{c}{Ambiguous box proposals.}\\
\vspace{-.2cm}\\
\vcenteredinclude{\includegraphics[width=\mywidth]{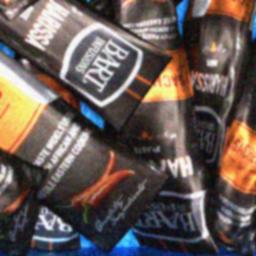}}&
\vcenteredinclude{\includegraphics[width=\mywidth]{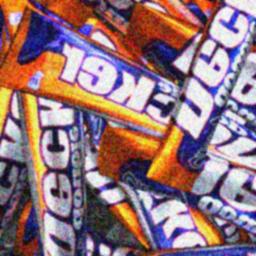}}&
\vcenteredinclude{\includegraphics[width=\mywidth]{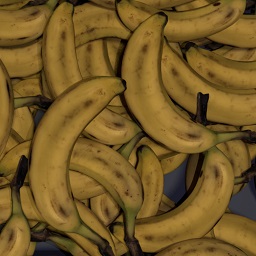}}&
\vcenteredinclude{\includegraphics[width=\mywidth]{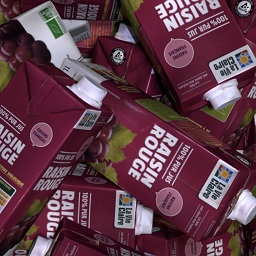}}\\
\vspace{-.25cm}\\
\multicolumn{4}{c}{Additional examples of dense object layouts in robotics.} \\
\end{tabular}
\caption{In dense object layouts, occlusions are mostly between instances that cannot be isolated in a rectangle. Mapping an image or a region that contains multiple similar instances to an instance-sensitive segmentation becomes ambiguous, thereby reducing the discriminative power of the encoded representations.}
\label{fig:piles}
\end{figure}

\begin{figure*}[t]
\centering
\setlength\mywidth{1.4cm}
\setlength\mywidthbis{3mm}
\setlength\thirdcolumn{1.1cm}
\setlength\secondcolumn{.06\linewidth}
\setlength\tabcolsep{4pt}%
\begin{tabular}{ccc}
\vcenteredinclude{
\tikzstyle{img} = [draw=none,inner sep=0pt]%
\tikzstyle{conv} = [rectangle, draw, fill=blue!20,%
    text width=2em, text centered, rounded corners=0.5mm, text height=1.5mm, inner sep=0]%
\tikzstyle{conv2} = [conv,fill=blue!60]%
\tikzstyle{conv3} = [conv,fill=blue!60!green]%
\tikzstyle{data} = [conv, rounded corners=2mm,fill=white!20,text width=2em,text centered, text height=.8em, text depth=.1em,inner sep=2pt]%
\tikzstyle{pool} = [rectangle, draw, fill=orange!20,text width=2em,text height=1mm, inner sep=0]%
\tikzstyle{hidepool} = [rectangle, draw=white, fill=white,minimum width=2em,text height=.2em, inner sep=0pt, line width=.03em,anchor=center]%
\tikzstyle{unpool} = [pool,fill=orange!80]%
\tikzstyle{convarrow} = [-{Latex[length=2mm,width=1mm]},line width=.2mm]%
\tikzstyle{convline} = [-,line width=.4mm]%
\tikzstyle{poolarrow} = [convarrow,dashed]%
\tikzstyle{scaleann}=[draw=none,text width=1.4em,align=left,midway,left=1pt,font={\fontsize{10}{9}\selectfont}]%
\def\showcaseOut{45}
\def\showcaseIn{135}
\begin{tikzpicture}
\node [hidepool] (pool1) {};%
\node [conv3, below=\mywidth of pool1] (conv2) {};%
\node [pool, below=\mywidthbis of conv2] (pool2) {};%
\node [conv3, below=\mywidth of pool2] (conv3) {};%
\node [hidepool, below=\mywidthbis of conv3] (pool3) {};%
\node [hidepool, right=\thirdcolumn of pool1.center,anchor=center] (unpool1) {};%
\node [conv, right=\thirdcolumn of conv2.center,anchor=center] (deconv2) {};%
\node [unpool, right=\thirdcolumn of pool2.center,anchor=center] (unpool2) {};%
\node [conv, right=\thirdcolumn of conv3.center,anchor=center] (deconv3) {};%
\node [hidepool, right=\thirdcolumn of pool3.center,anchor=center] (unpool3) {};%
\node [img, above=7mm of pool1] (image-img) {\includegraphics[width=\secondcolumn]{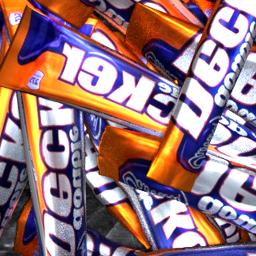}};%
\node [img, above=7mm of unpool1] (image-out) {\includegraphics[width=\secondcolumn]{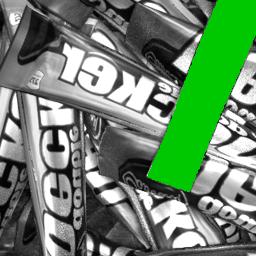}};%
\draw[convarrow] (pool1) -- (conv2);%
\draw[convarrow] (conv2) -- (pool2);%
\draw[convarrow] (pool2) -- (conv3);%
\draw[convarrow] (conv3) -- (pool3);%
\draw[convarrow] (unpool3) -- (deconv3);%
\draw[convarrow] (deconv3) -- (unpool2);%
\draw[convarrow] (unpool2) -- (deconv2);%
\draw[convarrow] (deconv2) -- (unpool1);%
\draw[convarrow] (conv2) -- (deconv2);%
\draw[convarrow] (conv3) -- (deconv3);%
\node [draw=none, anchor=south east, above left =.0cm and .2cm of pool1.north west] (axiso) {};%
\node [draw=none, anchor=south west, above right=.0cm and .1cm of unpool1.north east] (axisx) {};%
\node [draw=none, below=3.8cm of axiso.south] (axisy) {};%
\draw[convarrow,dashed,color=white] (axiso.center) -- (axisx.west);%
\draw[convarrow,dashed,color=black] (axiso.center) -- (axisy.center);%
\path [] (axiso.center) -- node [draw=none,text width=4em,inner sep=0,align=center,midway,above=2pt] {\phantom{taspk}} (axisx.center);%
\path [] (axiso.center) -- node [draw=none,inner sep=0,align=right,midway,left=5pt] {\rotatebox{90}{lower resolution}} (axisy.center);
\end{tikzpicture}
}&
\vcenteredinclude{
\tikzstyle{img} = [draw=none,inner sep=0pt]%
\tikzstyle{conv} = [rectangle, draw, fill=blue!20,%
    text width=2em, text centered, rounded corners=0.5mm, text height=1.5mm, inner sep=0]%
\tikzstyle{conv2} = [conv,fill=blue!60]%
\tikzstyle{conv3} = [conv,fill=blue!60!green]%
\tikzstyle{data} = [conv, rounded corners=2mm,fill=white!20,text width=2em,text centered, text height=.8em, text depth=.1em,inner sep=2pt]%
\tikzstyle{pool} = [rectangle, draw, fill=orange!20,text width=2em,text height=1mm, inner sep=0]%
\tikzstyle{hidepool} = [rectangle, draw=white, fill=white,minimum width=2em,text height=.2em, inner sep=0pt, line width=.03em,anchor=center]%
\tikzstyle{unpool} = [pool,fill=orange!80]%
\tikzstyle{convarrow} = [-{Latex[length=2mm,width=1mm]},line width=.2mm]%
\tikzstyle{convline} = [-,line width=.4mm]%
\tikzstyle{poolarrow} = [convarrow,dashed]%
\tikzstyle{scaleann}=[draw=none,text width=1.4em,align=left,midway,left=1pt,font={\fontsize{10}{9}\selectfont}]%
\def\showcaseOut{45}
\def\showcaseIn{135}
\begin{tikzpicture}
\node [hidepool] (pool1) {};%
\node [conv3, below=\mywidth of pool1] (conv2) {};%
\node [pool, below=\mywidthbis of conv2] (pool2) {};%
\node [conv3, below=\mywidth of pool2] (conv3) {};%
\node [hidepool, below=\mywidthbis of conv3] (pool3) {};%
\node [hidepool, right=5\thirdcolumn of pool1.center,anchor=center] (unpool1e) {};%
\node [conv, right=5\thirdcolumn of conv2.center,anchor=center] (deconv2e) {};%
\node [unpool, right=5\thirdcolumn of pool2.center,anchor=center] (unpool2e) {};%
\node [conv, right=5\thirdcolumn of conv3.center,anchor=center] (deconv3e) {};%
\node [hidepool, right=5\thirdcolumn of pool3.center,anchor=center] (unpool3e) {};%
\node [hidepool, right=4\thirdcolumn of pool1.center,anchor=center] (unpool1d) {};%
\node [conv, right=4\thirdcolumn of conv2.center,anchor=center] (deconv2d) {};%
\node [pool, right=4\thirdcolumn of pool2.center,anchor=center] (unpool2d) {};%
\node [conv, right=4\thirdcolumn of conv3.center,anchor=center] (deconv3d) {};%
\node [hidepool, right=4\thirdcolumn of pool3.center,anchor=center] (unpool3d) {};%
\node [hidepool, right=3\thirdcolumn of pool1.center,anchor=center] (unpool1c) {};%
\node [conv, right=3\thirdcolumn of conv2.center,anchor=center] (deconv2c) {};%
\node [unpool, right=3\thirdcolumn of pool2.center,anchor=center] (unpool2c) {};%
\node [conv, right=3\thirdcolumn of conv3.center,anchor=center] (deconv3c) {};%
\node [hidepool, right=3\thirdcolumn of pool3.center,anchor=center] (unpool3c) {};%
\node [hidepool, right=2\thirdcolumn of pool1.center,anchor=center] (unpool1b) {};%
\node [conv, right=2\thirdcolumn of conv2.center,anchor=center] (deconv2b) {};%
\node [unpool, right=2\thirdcolumn of pool2.center,anchor=center] (unpool2b) {};%
\node [conv, right=2\thirdcolumn of conv3.center,anchor=center] (deconv3b) {};%
\node [hidepool, right=2\thirdcolumn of pool3.center,anchor=center] (unpool3b) {};%
\node [hidepool, right=\thirdcolumn of pool1.center,anchor=center] (unpool1) {};%
\node [conv, right=\thirdcolumn of conv2.center,anchor=center] (deconv2) {};%
\node [unpool, right=\thirdcolumn of pool2.center,anchor=center] (unpool2) {};%
\node [conv, right=\thirdcolumn of conv3.center,anchor=center] (deconv3) {};%
\node [hidepool, right=\thirdcolumn of pool3.center,anchor=center] (unpool3) {};%
\node [img, above=7mm of pool1] (image-img) {\includegraphics[width=\secondcolumn]{aim256-x.jpg}};%
\node [img, above=7mm of unpool1] (image-x1) {\includegraphics[width=\secondcolumn]{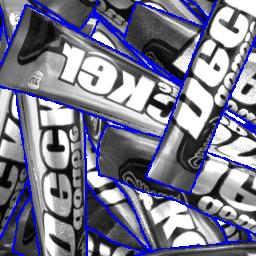}};%
\node [img, above=7mm of unpool1b] (image-x2) {\includegraphics[width=\secondcolumn]{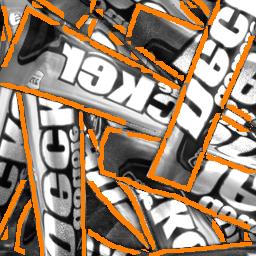}};%
\node [img, above=7mm of unpool1c] (image-y1) {\includegraphics[width=\secondcolumn]{aim256-mskgt.jpg}};
\node [img, above=7mm of unpool1e] (image-out) {\includegraphics[width=\secondcolumn]{aim256-mskgt.jpg}};
\draw[convarrow] (pool1) -- (conv2);%
\draw[convarrow] (conv2) -- (pool2);%
\draw[convarrow] (pool2) -- (conv3);%
\draw[convarrow] (conv3) -- (pool3);%
\draw[convarrow] (unpool3) -- (deconv3);%
\draw[convarrow] (deconv3) -- (unpool2);%
\draw[convarrow] (unpool2) -- (deconv2);%
\draw[convarrow] (deconv2) -- (unpool1);%
\draw[convarrow] (unpool3b) -- (deconv3b);%
\draw[convarrow] (deconv3b) -- (unpool2b);%
\draw[convarrow] (unpool2b) -- (deconv2b);%
\draw[convarrow] (deconv2b) -- (unpool1b);%
\draw[convarrow] (unpool3c) -- (deconv3c);%
\draw[convarrow] (deconv3c) -- (unpool2c);%
\draw[convarrow] (unpool2c) -- (deconv2c);%
\draw[convarrow] (deconv2c) -- (unpool1c);%
\draw[convarrow] (unpool1d) -- (deconv2d);%
\draw[convarrow] (deconv2d) -- (unpool2d);%
\draw[convarrow] (unpool2d) -- (deconv3d);%
\draw[convarrow] (deconv3d) -- (unpool3d);%
\draw[convarrow] (unpool3e) -- (deconv3e);%
\draw[convarrow] (deconv3e) -- (unpool2e);%
\draw[convarrow] (unpool2e) -- (deconv2e);%
\draw[convarrow] (deconv2e) -- (unpool1e);%
\draw[convarrow] (conv2) -- (deconv2);%
\draw[convarrow] (conv3) -- (deconv3);%
\draw[convarrow] (conv2.east) to [out=\showcaseOut,in=\showcaseIn] (deconv2b);%
\draw[convarrow] (conv3.east) to [out=\showcaseOut,in=\showcaseIn] (deconv3b);%
\draw[convarrow] (conv2.east) to [out=\showcaseOut,in=\showcaseIn] (deconv2c);%
\draw[convarrow] (conv3.east) to [out=\showcaseOut,in=\showcaseIn] (deconv3c);%
\draw[convarrow] (conv2.east) to [out=\showcaseOut,in=\showcaseIn] (deconv2d);%
\draw[convarrow] (conv3.east) to [out=\showcaseOut,in=\showcaseIn] (deconv3d);%
\draw[convarrow] (conv2.east) to [out=\showcaseOut,in=\showcaseIn] (deconv2e);%
\draw[convarrow] (conv3.east) to [out=\showcaseOut,in=\showcaseIn] (deconv3e);%
\draw[convarrow] (deconv2.east) to [out=\showcaseOut,in=\showcaseIn] (deconv2c);%
\draw[convarrow] (deconv3.east) to [out=\showcaseOut,in=\showcaseIn] (deconv3c);%
\draw[convarrow] (deconv2.east) to [out=\showcaseOut,in=\showcaseIn] (deconv2d);%
\draw[convarrow] (deconv3.east) to [out=\showcaseOut,in=\showcaseIn] (deconv3d);%
\draw[convarrow] (deconv2.east) to [out=\showcaseOut,in=\showcaseIn] (deconv2e);%
\draw[convarrow] (deconv3.east) to [out=\showcaseOut,in=\showcaseIn] (deconv3e);%
\draw[convarrow] (deconv2b.east) to [out=\showcaseOut,in=\showcaseIn] (deconv2d);%
\draw[convarrow] (deconv3b.east) to [out=\showcaseOut,in=\showcaseIn] (deconv3d);%
\draw[convarrow] (deconv2b.east) to [out=\showcaseOut,in=\showcaseIn] (deconv2e);%
\draw[convarrow] (deconv3b.east) to [out=\showcaseOut,in=\showcaseIn] (deconv3e);%
\draw[convarrow] (deconv2c.east) to [out=\showcaseOut,in=\showcaseIn] (deconv2e);%
\draw[convarrow] (deconv3c.east) to [out=\showcaseOut,in=\showcaseIn] (deconv3e);%
\draw[convarrow] (deconv2) -- (deconv2b);%
\draw[convarrow] (deconv3) -- (deconv3b);%
\draw[convarrow] (deconv2b) -- (deconv2c);%
\draw[convarrow] (deconv3b) -- (deconv3c);%
\draw[convarrow] (deconv2c) -- (deconv2d);%
\draw[convarrow] (deconv3c) -- (deconv3d);%
\draw[convarrow] (deconv2d) -- (deconv2e);%
\draw[convarrow] (deconv3d) -- (deconv3e);%
\node [draw=none, anchor=south east, above left =.0cm and .2cm of pool1.north west] (axiso) {};%
\node [draw=none, anchor=south west, above right=.0cm and .1cm of unpool1e.north east] (axisx) {};%
\node [draw=none, below=3.8cm of axiso.south] (axisy) {};%
\draw[convarrow,dashed,color=black] (axiso.center) -- (axisx.west);%
\draw[convarrow,dashed,color=black] (axiso.center) -- (axisy.center);%
\path [] (axiso.center) -- node [draw=none,text width=19em,inner sep=0,align=center,midway,above=2pt] {boundaries, occluding sides, instances} (axisx.center);
\path [] (axiso.center) -- node [draw=none,inner sep=0,align=right,midway,left=5pt] {\rotatebox{90}{lower resolution}} (axisy.center);
\end{tikzpicture}
}\raisebox{-2em}{$\equiv$}\vcenteredinclude{
\tikzstyle{img} = [draw=none,inner sep=0pt]%
\tikzstyle{conv} = [rectangle, draw, fill=blue!20,%
    text width=2em, text centered, rounded corners=0.5mm, text height=1.5mm, inner sep=0]%
\tikzstyle{conv2} = [conv,fill=blue!60]%
\tikzstyle{conv3} = [conv,fill=blue!60!green]%
\tikzstyle{data} = [conv, rounded corners=2mm,fill=white!20,text width=2em,text centered, text height=.8em, text depth=.1em,inner sep=2pt]%
\tikzstyle{pool} = [rectangle, draw, fill=orange!20,text width=2em,text height=1mm, inner sep=0]%
\tikzstyle{hidepool} = [rectangle, draw=white, fill=white,minimum width=2em,text height=.2em, inner sep=0pt, line width=.03em,anchor=center]%
\tikzstyle{unpool} = [pool,fill=orange!80]%
\tikzstyle{convarrow} = [-{Latex[length=2mm,width=1mm]},line width=.2mm]%
\tikzstyle{convline} = [-,line width=.4mm]%
\tikzstyle{poolarrow} = [convarrow,dashed]%
\tikzstyle{mcconv} = [conv,fill=purple!50]%
\tikzstyle{scaleann}=[draw=none,text width=1.4em,align=left,midway,left=1pt,font={\fontsize{10}{9}\selectfont}]%
\def\showcaseOut{45}
\def\showcaseIn{135}
\begin{tikzpicture}
\node [hidepool] (pool1) {};%
\node [conv3, below=\mywidth of pool1] (conv2) {};%
\node [pool, below=\mywidthbis of conv2] (pool2) {};%
\node [conv3, below=\mywidth of pool2] (conv3) {};%
\node [hidepool, below=\mywidthbis of conv3] (pool3) {};%
\node [hidepool, right=\thirdcolumn of pool1.center,anchor=center] (unpool1) {};%
\node [mcconv, right=\thirdcolumn of conv2.center,anchor=center] (deconv2) {};%
\node [hidepool, right=\thirdcolumn of pool2.center,anchor=center] (unpool2) {};%
\node [mcconv, right=\thirdcolumn of conv3.center,anchor=center] (deconv3) {};%
\node [hidepool, right=\thirdcolumn of pool3.center,anchor=center] (unpool3) {};%
\node [img, above=7mm of pool1] (image-img) {\includegraphics[width=\secondcolumn]{aim256-x.jpg}};%
\node [img, above=7mm of unpool1] (image-out) {\includegraphics[width=\secondcolumn]{aim256-mskgt.jpg}};%
\draw[convarrow] (pool1) -- (conv2);%
\draw[convarrow] (conv2) -- (pool2);%
\draw[convarrow] (pool2) -- (conv3);%
\draw[convarrow] (conv3) -- (pool3);%
\draw[convarrow] (unpool3) -- (deconv3);%
\draw[convarrow] (deconv3) -- (deconv2);%
\draw[convarrow] (deconv2) -- (unpool1);%
\draw[convarrow] (conv2) -- (deconv2);%
\draw[convarrow] (conv3) -- (deconv3);%
\node [draw=none, anchor=south east, above left =.0cm and 0cm of pool1.north west] (axiso) {};%
\node [draw=none, anchor=south west, above right=.0cm and .1cm of unpool1.north east] (axisx) {};%
\node [draw=none, below=3.8cm of axiso.south] (axisy) {};%
\draw[convarrow,dashed,color=white] (axiso.center) -- (axisx.west);%
\path [] (axiso.center) -- node [draw=none,text width=4em,inner sep=0,align=center,midway,above=2pt] {\phantom{taspk}} (axisx.center);%
\end{tikzpicture}
}&
\vcenteredinclude{%
\setlength{\mywidth}{.7cm}%
\setlength{\mywidthbis}{.1cm}%
\tikzstyle{data} = [rectangle, draw, fill=white!20,%
    text width=2em, text centered, rounded corners, minimum height=.5mm]%
\tikzstyle{conv} = [rectangle, draw, fill=blue!20,%
    text width=2em, text centered, rounded corners=0.5mm, text height=1.5mm, inner sep=0]%
\tikzstyle{conv2} = [conv,fill=blue!60!green]%
\tikzstyle{mcconv} = [conv,fill=purple!50]%
\tikzstyle{sum} = [circle, draw, fill=blue!20, text height=1mm, inner sep=0]%
\tikzstyle{relu} = [rectangle, draw, fill=red!20,minimum width=3em]%
\tikzstyle{pool} = [rectangle, draw, fill=orange!20,text width=2em,text height=1mm, inner sep=0]%
\tikzstyle{hidepool} = [rectangle, draw=white, fill=white,minimum width=2em,text height=.2em, inner sep=0pt, line width=.03em,anchor=center]%
\tikzstyle{unpool} = [pool,fill=orange!80]%
\tikzstyle{convtext} = [draw=none,fill=none, align=left]%
\begin{tikzpicture}
\node [conv2] (conv2) {};%
\node [conv, below=\mywidth of conv2] (conv1) {};%
\node [mcconv, below=\mywidth of conv1] (conv3) {};%
\node [pool, below=\mywidth of conv3] (pool1) {};%
\node [unpool, below=\mywidth of pool1] (unpool1) {};%
\node[convtext, below right=4mm and \mywidthbis of conv2.west, anchor=west] {Image encoder block};%
\node[convtext, below right=4mm and \mywidthbis of conv1.west, anchor=west] {Concat+Conv+ReLU};%
\node[convtext, below right=4mm and \mywidthbis of conv3.west, anchor=west] {Multicameral node};%
\node[convtext, below right=4mm and \mywidthbis of pool1.west, anchor=west] {Spatial pooling ($.5\times$)};%
\node[convtext, below right=3mm and \mywidthbis of unpool1.west, anchor=west] (unpoolname) {Spatial unpooling ($2\times$)};%
\end{tikzpicture}%
%
}\\
\vspace{-.2cm}\\
Encoder-decoder block &Multicameral block (Ours) &\\
\end{tabular}

\bigskip

\setlength\tabcolsep{1pt}%
\setlength\mywidth{.24\linewidth}
\begin{tabular}{cccc}
\vcenteredinclude{\includegraphics[width=\mywidth]{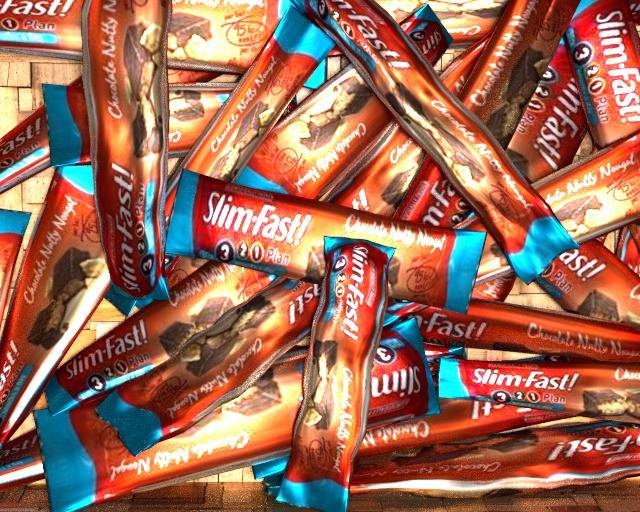}}&
\vcenteredinclude{\includegraphics[width=\mywidth]{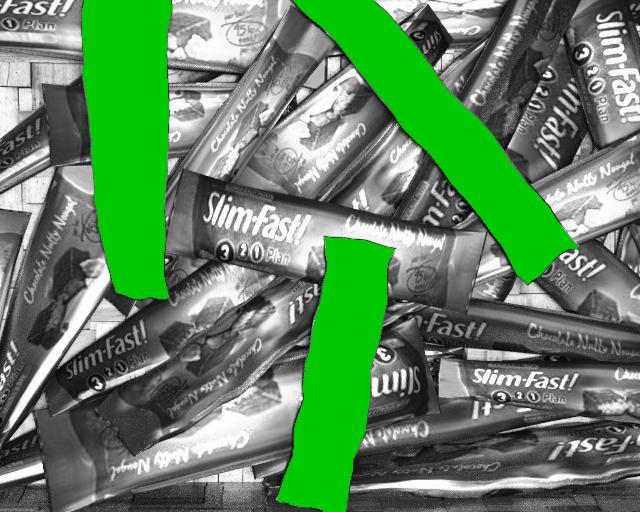}}&
\vcenteredinclude{\includegraphics[width=\mywidth]{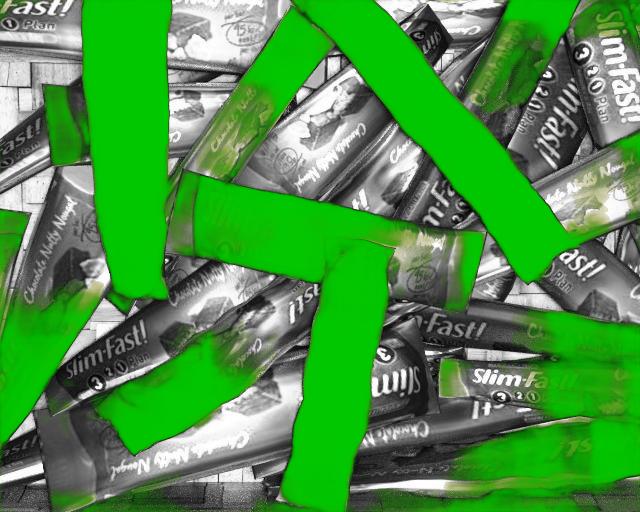}}&
\vcenteredinclude{\includegraphics[width=\mywidth]{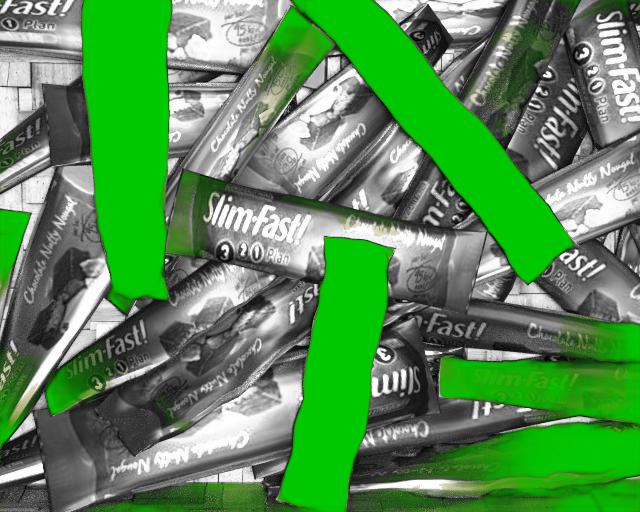}}\\
\vspace{-.2cm}\\
Input&Expected result& Encoder-decoder FCN &Multicameral FCN (Ours)\\
\end{tabular}
\caption{Due to its built-in translation invariance, a deep encoder can hardly be decoded for distinguishing similar overlapping instances. We show the importance of decomposing the decoding process into ordinal subtasks to improve the attention to unoccluded instances in homogeneous layouts.}
\label{fig:showcase}
\end{figure*} 

\section{Introduction}
\label{sec:intro}

Outlining object instances and understanding their spatial layout from a single RGB image without explicit object models is a core computer vision task in many robotic applications, such as object picking and autonomous driving in unknown environments. Indeed, the least occluded instances are often the most affordable ones to grasp or the closest obstacles to avoid. Automating such a task remains challenging as a robot must handle many variations of scene layouts from a mere grid of RGB values.

Deep fully convolutional networks (FCN) have become the state of the art for learning generalizable image representations due to their ability to capture multiscale invariants in trainable convolution kernels. In this context, a mainstream strategy for detecting salient instances consist in splitting the image segmentation into many region-wise segmentations. Specifically, a two-step FCN is trained to first isolate each instance in a bounding box by joint classification and regression of anchor boxes, then for each box proposal fire the pixels that belong to the visible and occluded instance parts \citep{Qi2019amodal,FollmannKHKB19,Zhu17amodal} or to predefined affordance categories \citep{Do18affordancenet}. However, approximating an instance as a rectangle is not always relevant. Typically, in dense homogeneous layouts, many instances of the same object occlude each other. As a result, a box proposal often contains multiple instances (\cf{} Figure \ref{fig:piles}). 

In such object layouts, mapping an image or a region to an instance-sensitive segmentation becomes a difficult task, because a pixel-wise attention to specific instances requires position-dependent representations, whereas convolution kernels are translation invariant. Generally, pixel-wise labels are inferred by gradually combining low-resolution object-level semantics and higher-resolution local cues using a residual encoder-decoder (RED) network. In such a structure, the decoder aims to upsample the encoder latent representations. RED networks have proved efficient for inferring instance-agnostic categories \citep{deeplabv3+18} and instance boundaries \citep{DengSLWL18,CED17,unet15}. However, a deep encoder can hardly be decoded for distinguishing similar overlapping instances, due to its built-in translation invariance (\cf{} Figure \ref{fig:showcase}). Most research efforts to improve object delineation have been put in the encoder, using densely connected layers to deepen the encoder blocks \citep{densenet17}, dilated convolutions to enlarge the receptive field at the lowest-resolution encoding level \citep{deeplabv3+18,WangCYLHHC18,YuK16} or coordinate-aware convolutions to associate the latent representations with global pixel locations \citep{Liu18coordconv,NovotnyALV18}. These design patterns lead to low-resolution position-dependent representations of object categories, easier to be upsampled. However, in dense homogeneous layouts, the decoding process has greater importance because the diversity of objects to encode is much reduced while the pixel embeddings must discriminate between instances of the same object.

We therefore further the residual encoder-decoder design in order to approximate a mapping between single RGB images of homogeneous instance layouts and occlusion-aware instance-sensitive segmentations. Specifically, we propose a more complex decoding process to produce contextual pixel embeddings that better discriminate between similar instances. Our multicameral design consists of lightweight decoder and encoder-decoder units densely coupled in cascade, and differently supervised to decompose the complex task of outlining unoccluded instances into simpler ones: extracting image cues, detecting instance boundaries, detecting occluding boundary sides, firing the pixels of unoccluded instances, refining the segmentation. In contrast with the state-of-the-art design patterns for capturing position-dependent representations, our approach encourages subtask-specific feature reuse and longer-range relations within the decoding process, thus improving the attention to unoccluded instances in homogeneous layouts (\cf{} Figure \ref{fig:showcase}).

\begin{figure*}[t]
\centering
\setlength{\mywidth}{.1021\textwidth}
\setlength{\tabcolsep}{1pt}
\begin{tabular}{cccp{0.15cm}cccp{0.15cm}ccc}
\multicolumn{3}{c}{\textbf{BSDS-BOW} \citep{Ren2006}} && \multicolumn{3}{c}{\textbf{PIOD} \citep{doc16}} && \multicolumn{3}{c}{\textbf{COCOA} \citep{Zhu17amodal}} \\
\vcenteredinclude{\includegraphics[width=\mywidth]{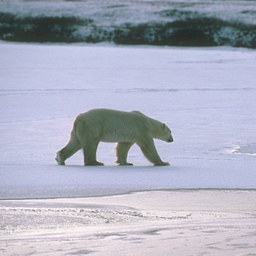}}&
\vcenteredinclude{\includegraphics[width=\mywidth]{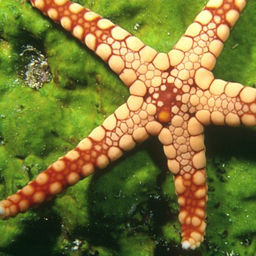}}&
\vcenteredinclude{\includegraphics[width=\mywidth]{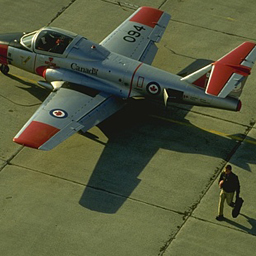}}&&
\vcenteredinclude{\includegraphics[width=\mywidth]{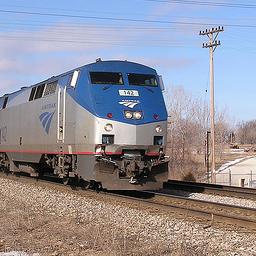}}&
\vcenteredinclude{\includegraphics[width=\mywidth]{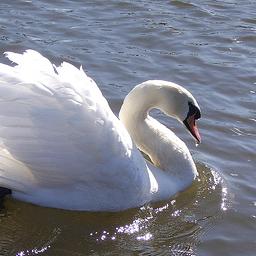}}&
\vcenteredinclude{\includegraphics[width=\mywidth]{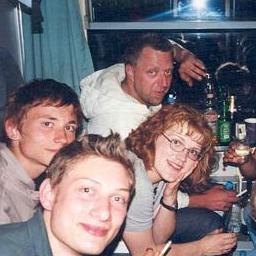}}&&
\vcenteredinclude{\includegraphics[width=\mywidth]{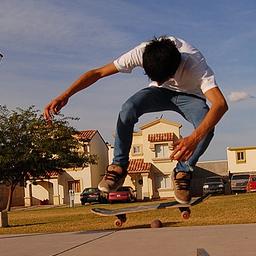}}&
\vcenteredinclude{\includegraphics[width=\mywidth]{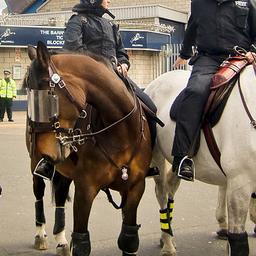}}&
\vcenteredinclude{\includegraphics[width=\mywidth]{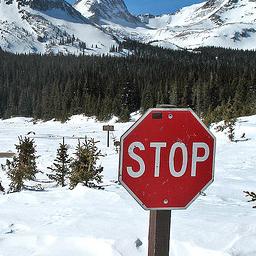}}\\
\vspace{-0.35cm}\\
\vcenteredinclude{\includegraphics[width=\mywidth]{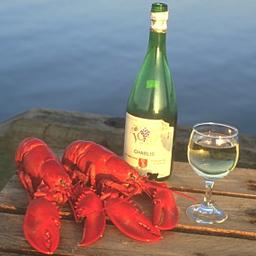}}&
\vcenteredinclude{\includegraphics[width=\mywidth]{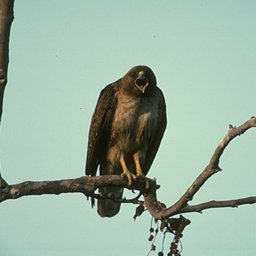}}&
\vcenteredinclude{\includegraphics[width=\mywidth]{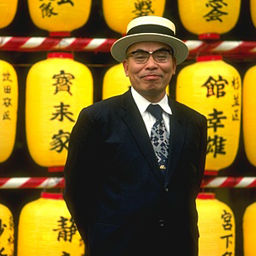}}&&
\vcenteredinclude{\includegraphics[width=\mywidth]{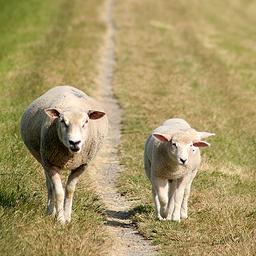}}&
\vcenteredinclude{\includegraphics[width=\mywidth]{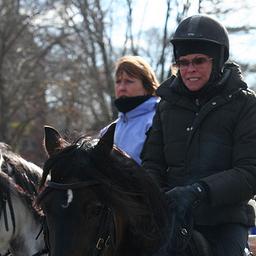}}&
\vcenteredinclude{\includegraphics[width=\mywidth]{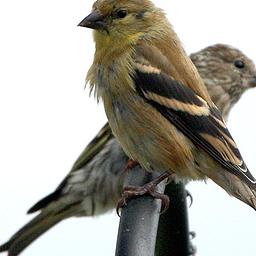}}&&
\vcenteredinclude{\includegraphics[width=\mywidth]{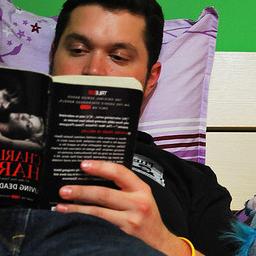}}&
\vcenteredinclude{\includegraphics[width=\mywidth]{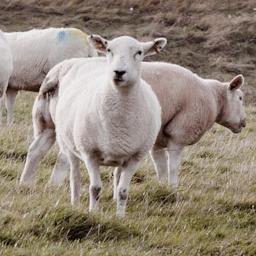}}&
\vcenteredinclude{\includegraphics[width=\mywidth]{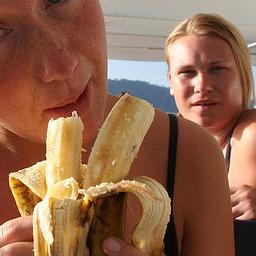}}\\
\vspace{-0.35cm}\\
\vcenteredinclude{\includegraphics[width=\mywidth]{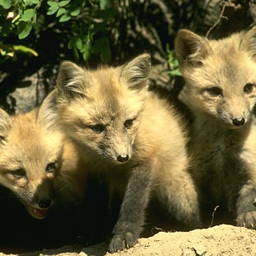}}&
\vcenteredinclude{\includegraphics[width=\mywidth]{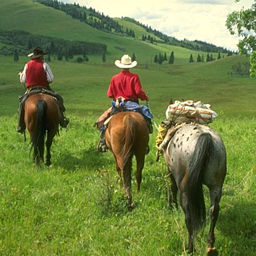}}&
\vcenteredinclude{\includegraphics[width=\mywidth]{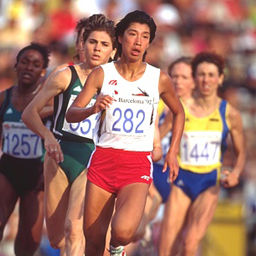}}&&
\vcenteredinclude{\includegraphics[width=\mywidth]{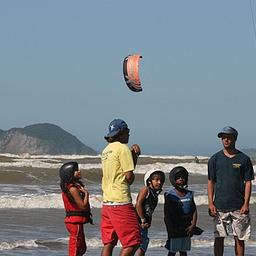}}&
\vcenteredinclude{\includegraphics[width=\mywidth]{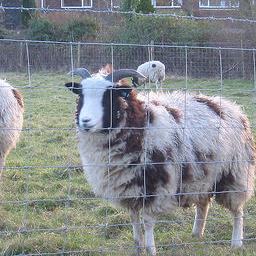}}&
\vcenteredinclude{\includegraphics[width=\mywidth]{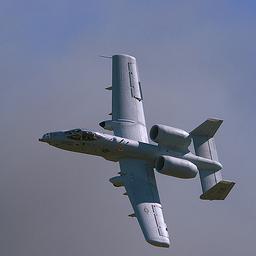}}&&
\vcenteredinclude{\includegraphics[width=\mywidth]{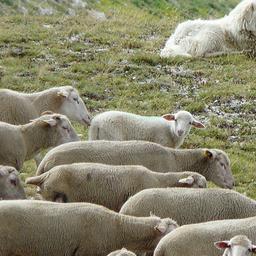}}&
\vcenteredinclude{\includegraphics[width=\mywidth]{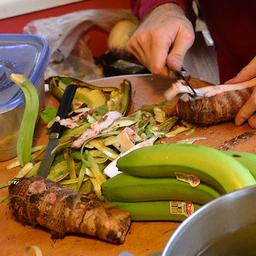}}&
\vcenteredinclude{\includegraphics[width=\mywidth]{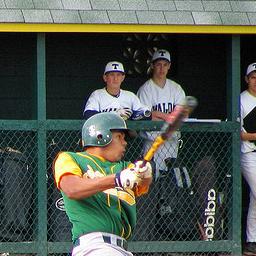}}\\
\\
 \multicolumn{3}{c}{\textbf{D2SA} \citep{FollmannKHKB19}} && \multicolumn{3}{c}{\textbf{KINS} \citep{Qi2019amodal}} && \multicolumn{3}{c}{\textbf{Mikado/Mikado+ (Ours)}} \\
 \vcenteredinclude{\includegraphics[width=\mywidth]{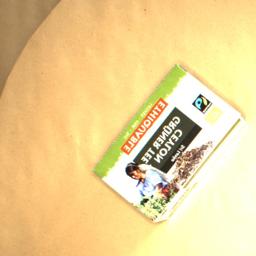}}&
\vcenteredinclude{\includegraphics[width=\mywidth]{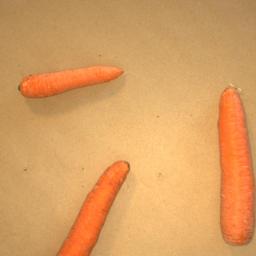}}&
\vcenteredinclude{\includegraphics[width=\mywidth]{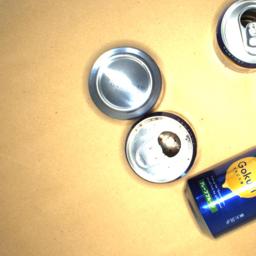}}&&
\vcenteredinclude{\includegraphics[width=\mywidth]{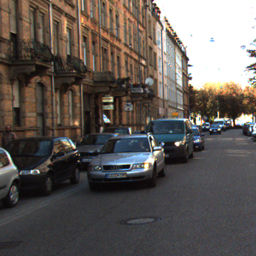}}&
\vcenteredinclude{\includegraphics[width=\mywidth]{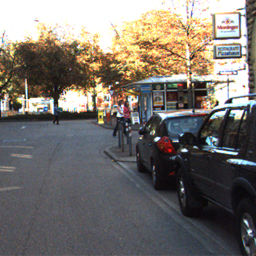}}&
\vcenteredinclude{\includegraphics[width=\mywidth]{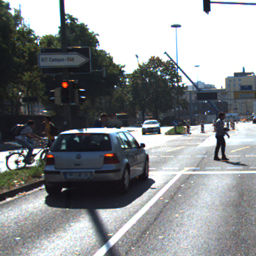}}&&
\vcenteredinclude{\includegraphics[width=\mywidth]{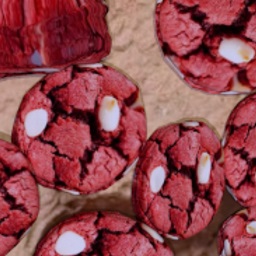}}&
\vcenteredinclude{\includegraphics[width=\mywidth]{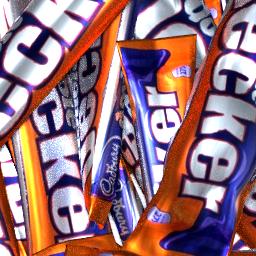}}&
\vcenteredinclude{\includegraphics[width=\mywidth]{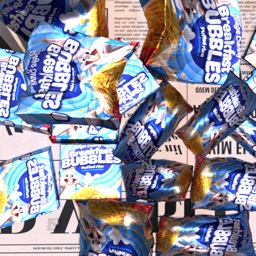}}\\
\vspace{-0.35cm}\\
\vcenteredinclude{\includegraphics[width=\mywidth]{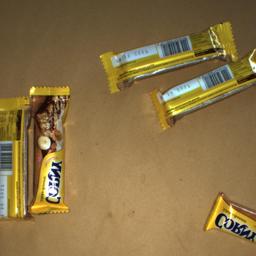}}&
\vcenteredinclude{\includegraphics[width=\mywidth]{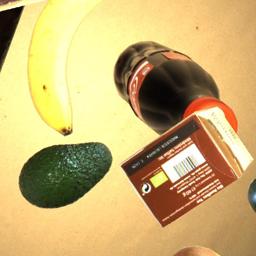}}&
\vcenteredinclude{\includegraphics[width=\mywidth]{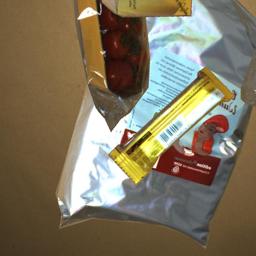}}&&
\vcenteredinclude{\includegraphics[width=\mywidth]{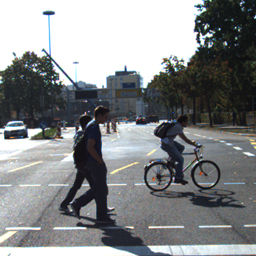}}&
\vcenteredinclude{\includegraphics[width=\mywidth]{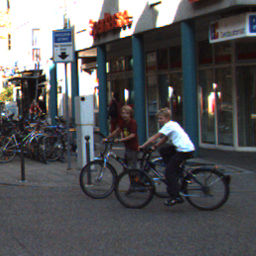}}&
\vcenteredinclude{\includegraphics[width=\mywidth]{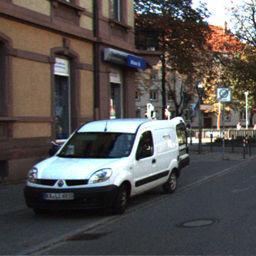}}&&
\vcenteredinclude{\includegraphics[width=\mywidth]{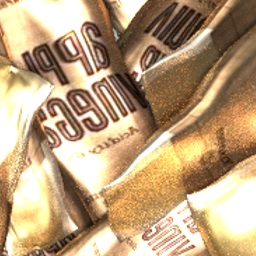}}&
\vcenteredinclude{\includegraphics[width=\mywidth]{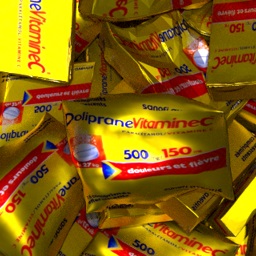}}&
\vcenteredinclude{\includegraphics[width=\mywidth]{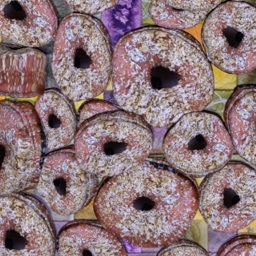}}\\
\vspace{-0.35cm}\\
\vcenteredinclude{\includegraphics[width=\mywidth]{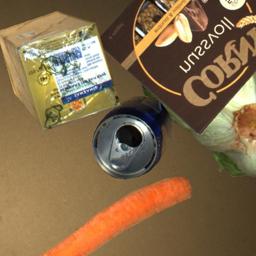}}&
\vcenteredinclude{\includegraphics[width=\mywidth]{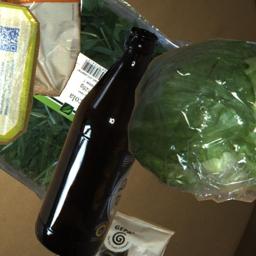}}&
\vcenteredinclude{\includegraphics[width=\mywidth]{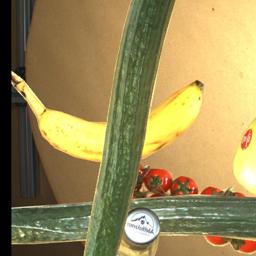}}&&
\vcenteredinclude{\includegraphics[width=\mywidth]{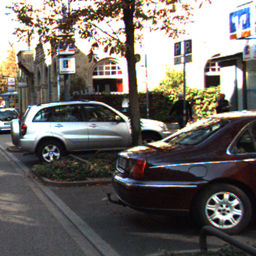}}&
\vcenteredinclude{\includegraphics[width=\mywidth]{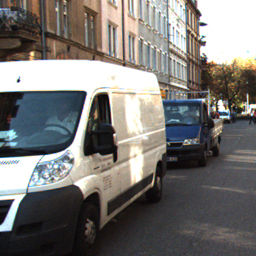}}&
\vcenteredinclude{\includegraphics[width=\mywidth]{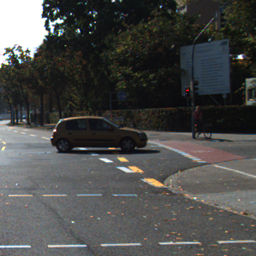}}&&
\vcenteredinclude{\includegraphics[width=\mywidth]{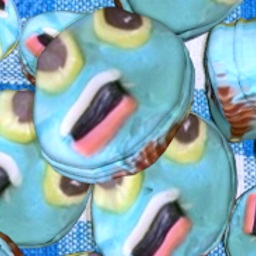}}&
\vcenteredinclude{\includegraphics[width=\mywidth]{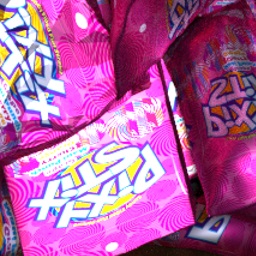}}&
\vcenteredinclude{\includegraphics[width=\mywidth]{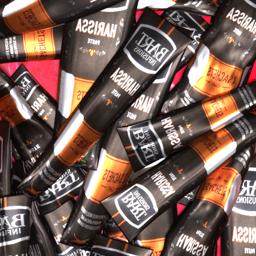}}\\
\vspace{-0.35cm}\\
\end{tabular}

\bigskip

\setlength{\tabcolsep}{6pt}
\begin{tabular}{r|l|l|l|l|l|l|l}
Dataset & 
\parbox{1.4cm}{Average\\ image size} & 
\parbox{1.5cm}{Number of\\ images} & 
\parbox{1.5cm}{Number of\\ instances} & 
\parbox{1.4cm}{Instances\\ per image} & 
\parbox{1.5cm}{Inter-instance\\ occlusions\\ per image\\} & 
\parbox{1.7cm}{Background\\ pixels \\ per image} & 
\parbox{1.9cm}{Ground-truth\\ annotations} \\
\hline
BSDS-BOW$^{1}$&
432$\times$369 &
200 &
-- &
--& 
--&
--&
\multirow{5}*{Human-made}\\
PIOD &
469$\times$386 &
10,100 &
24,797 &
2.5 & 
1.3 &
69\% &
\\
\cline{1-7}
COCOA$^{2}$ &
578$\times$483 &
3,823 &
34,884 &
9.1 &
13.5 &
33\% &
\\
D2SA$^{2}$ &
1962$\times$1569 &
5,600 &
28,703 &
5.1 &
2.8 &
79\% &
\\
KINS &
1695$\times$362 &
14,991 &
187,730 &
12.5 &
8.0 &
92\% &
\\
\hline
\textbf{Mikado (Ours)} &
640$\times$512 &
2,400 &
48,184 &
\textbf{20.1} &
\textbf{52.9} &
\textbf{24\%} &
\multirow{2}*{\parbox{1.9cm}{\textbf{Computer-generated}}} \\
\textbf{Mikado+$^{3}$ (Ours)} &
640$\times$512 &
14,560 &
459,002 &
\textbf{31.5} &
\textbf{60.5} &
\textbf{24\%} &
\\
\multicolumn{8}{c}{}\\
\end{tabular}

\vspace{-.3cm}

\begin{flushleft}
\small

$^{1}$ The empty cells are due to the ground truth that consists only of object part-level oriented edges. \\

$^{2}$ The statistics are only on the train and validation subsets as the test subset is not provided.

$^{3}$ Mikado+ is an extension of Mikado used only to show the impact of a richer synthetic data distribution.
\end{flushleft}

\vspace{-.3cm}
\caption{State-of-the-art datasets for occlusion-aware boundary detection (BSDS-BOW, PIOD) and amodal instance segmentation (COCOA, D2SA, KINS) compared with our synthetic dataset. Unlike the state-of-the-art datasets in which occlusions are mostly due to objects occluding the background, Mikado contains more instances and occlusions between instances per image, thus better representing the variety of occlusions.}
\label{fig:datasets}
\end{figure*}

Furthermore, the state-of-the-art datasets for joint instance delineation and occlusion detection \citep{Qi2019amodal,FollmannKHKB19,Zhu17amodal,doc16,FuWTB16} are intrinsically designed for the foreground/back\-ground paradigm. As shown by Figure \ref{fig:datasets}, the images in these datasets contain few instances and a large number of occlusions are due to objects occluding the background. In addition, these datasets suffer from biased data distributions due to limited variations and error-prone hand-made annotations. They can hardly be extended, as producing a pixel-wise ground truth for instance boundaries and occlusions is a tedious and time-consuming task for human annotators. Specifically, these datasets never showcase homogeneous layouts with many occlusions between instances, although it is a common scenario in robotic applications for manufactured object manipulation.

Therefore, we also propose a synthetic dataset of dense homogeneous layouts for evaluating the learning of an instance-sensitive mapping, through the canonical scenario of many sachets piled up in bulk. Our data generation pipeline flexibly enables lots of inter-instance occlusion variations and error-free annotations, unlike datasets of real images.

In summary, our contribution is two-fold:
\begin{itemize}
\item A \emph{multicameral} FCN design to approximate a more complex decoding function for dense homogeneous layouts. Our extensive experiments show that introducing complexity and task decomposition into ordinal subtasks within the decoding process proves more effective than the state-of-the-art design patterns for capturing position-dependent representations, thus improving the attention to unoccluded instances from a single RGB image.

\item A simulation-based pipeline, referred to as \emph{Mikado}, to evaluate the proposed model on dense homogeneous instance layouts. Our synthetic data\footnote{Publicly available at https://mikado.liris.cnrs.fr} extensibly contains more occlusions between similar instances than the public datasets for occlusion-aware instance segmentation. We show that the proposed data is plausible with respect to real-world problems, through experiments on transfer learning from Mikado to D2SA, a public dataset of real-world heterogeneous object layouts \citep{FollmannKHKB19}.
\end{itemize}

Our paper is organized as follows. After reviewing the related work in Section \ref{sec:relatedwork}, we describe the proposed model in Section \ref{sec:proposedmodel}, the proposed dataset in Section \ref{sec:proposeddataset}, then our experimental protocol in Section \ref{sec:experimentalsetup}. Our results are finally discussed in Section \ref{sec:results}.

\section{Related Work}
\label{sec:relatedwork}

Occlusion-aware instance-wise attention lies at the intersection of salient instance segmentation and occlusion detection. Also, the proposed multicameral design is composed of shared or task-specific encoders and decoders. In this section, we thus review the state of the art on salient instance segmentation and occlusion detection from a single RGB image, FCN architectures for pixel multi-labeling, and the public datasets for joint instance segmentation and occlusion detection.

\subsection{Salient Instance Segmentation}

\paragraph{Graph-based segmentation} 

Instance delineation has been approached further to pixel-wise object categorization. Specifically, an instance-agnostic category is first assigned to each pixel, then the pixels within each category region are grouped into instances using graphical models, such as watershed transforms from inferred energy maps \citep{WTN17} or superpixel-based proposals \citep{Li2017,KirillovLASR17,mcg17}. Indeed, in scenes with few similar or many heterogeneous instances, category masks effectively reduce the search space and partially reveal instance boundaries, as category boundaries are also instance boundaries. However, in scenes full of many instances of the same class (Figure \ref{fig:piles}), such a categorization is of little use. Defining instead instance-sensitive categories also fails, due to the built-in translation invariance of FCNs (Figure \ref{fig:showcase}).

\paragraph{Recurrent segmentation} 

Instance segmentation has also been formulated as a recurrent process \citep{KongF18a,RenZ17,Romera-ParedesT16}. Specifically, a recurrent FCN is trained to iteratively update a mean-shift clustering \citep{KongF18a} or iteratively outline each instance \citep{RenZ17,Romera-ParedesT16}. Such memory-based pipelines are nevertheless harder to train than feedforward networks. \citep{RenZ17,Romera-ParedesT16} also assume a stationary scene, wheras in robotic applications, the scene is likely to change between two iterations due to physical interactions with the detected instances.

\paragraph{Proposal-based segmentation}

Alternatively, state-of-the-art strategies rely on two-step FCNs trained to first isolate each instance in a rectangle, then infer the corresponding mask after pooling the high-level features in the box proposal \citep{panet18,maskrcnn17,HayderHS17,Fan19s4net,DaiHS16}. Although these approaches are good at producing connected pixel clusters, the resulting mask boundaries suffer from the pooling quantization effect. Starting instead from binary rectangle masks on the box detector's last feature map \citep{Fan19s4net} or using a distance transform \citep{HayderHS17} to infer instance masks improves instance delineation, but still for instances that can fit a rectangle. As discussed in our introduction, these approaches also poorly address the problem of translation variance using FCNs, particularly in the case of multiple overlapping instances of the same object. Interestingly, mixing convolutional embeddings with hard-coded non-convolutional information, such as pixel locations, enables improvements in distinguishing adjacent instances \citep{NovotnyALV18,Liu18coordconv}.

\subsection{Occlusion Detection}

\paragraph{Depth estimation}

Finding occlusion relations has most\-ly been studied jointly with depth estimation in multiview contexts \citep{ZitnickK00,GrammalidisS98,GeigerLY95} and motion sequences \citep{SunLP14,AyvaciRS12,Humayun2011,He2010,AyvaciRS10,Stein2006,WilliamsIM05}, as occlusions often translate into missing pixel correspondences in different points of view or consecutive frames. Recent works have more ambitiously focused on learning-based monocular 3D reconstruction using FCNs \citep{GanXSL18,FuGWBT18,LiuSL016,LiSDHH15,EigenPF14}, but the results are still less accurate than standard multi-view 3D reconstruction algorithms, and these techniques require sensor-specific ground-truth depth maps difficult to obtain. Although depth estimation brings relevant hints such as depth discontinuities, understanding occlusions is possible without putting effort into an explicit dense 3D reconstruction, as shown hereinafter.

\paragraph{Amodal/multiclass segmentation}

In keeping with box proposal-based instance segmentation \citep{panet18,maskrcnn17}, two-step FCNs have been adapted for inferring, in each box proposal, either the mask including the visible and occluded instance parts \citep{Qi2019amodal,FollmannKHKB19,Zhu17amodal} or a multiclass segmentation according to predefined affordance categories \citep{Do18affordancenet}. However, in addition to the cons of box proposal-based segmentation, inferring masks including occluded instance parts, referred to as \emph{amodal segmentation}, is ambiguous because some pixels are attached to something invisible, whereas these pixels visually belong to another instance. Without explicit object models, the learning process is then conditioned on a guess only from global pixel relations, while fine-grained inferences require local pixel relations as well. Amodal annotations are also difficult to obtain unless synthesizing training images, leading to a domain shift. Defining instead affordance categories seems more reasonable, but in \citep{Do18affordancenet}, affordances are implicitly mapped to object part categories. For example, wrapping grasp affordances are cylinder-like objects such as bottles, bowls, knife handles. In a scene full of overlapping instances of the same affordance category, this strategy is prone to fail.

\paragraph{Oriented boundary detection}

FCNs prove more suitable for learning oriented contours, as this pixel labeling task does not require translation variance. Specifically, state-of-the-art approaches employ encoder-decoder networks including two task-specific decoders for recovering instance boundaries and occlusion-based orientations respectively \citep{Wang18doobnet,doc16}. However, these approaches have two drawbacks. First, occlusions are modelled as pixel-specific raw orientations specifying the occlusion relations, without guarantee of continuity. As a consequence, a post-inference step is needed to adjust the noisy inferred orientations using the local tangent vectors of the inferred boundaries. Most importantly, the inferred boundaries are not guaranteed to be closed. As a consequence, instance masks cannot be easily extrapolated, \eg{} by considering the dual connected components. An iterative refinement procedure has been proposed \citep{Batra19}, but does not really solve the issue.

\subsection{Pixel Multi-labeling}

\paragraph{Encoder-decoder networks}

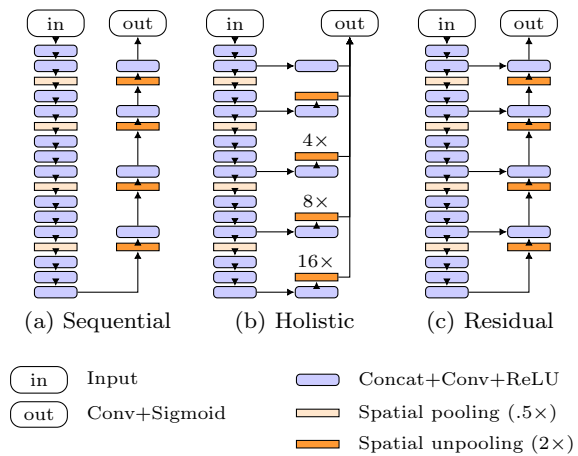
\begin{figure}[t]
\centering
\fontsize{7}{8}\selectfont
\setlength{\secondcolumn}{.25cm}
\setlength{\thirdcolumn}{.9cm}
\begin{tabular}{ccc}
\vcenteredinclude{\subfloat[Sequential]{\label{fig:arch-ed-sequential}
\tikzstyle{data} = [rectangle, draw, fill=white!20,%
    text width=2em, text centered, rounded corners, minimum height=.5mm]%
\tikzstyle{conv} = [rectangle, draw, fill=blue!20,%
    text width=2em, text centered, rounded corners=0.5mm, text height=1.5mm, inner sep=0]%
\tikzstyle{sum} = [circle, draw, fill=blue!20, text height=1mm, inner sep=0]%
\tikzstyle{relu} = [rectangle, draw, fill=red!20,minimum width=3em]%
\tikzstyle{pool} = [rectangle, draw, fill=orange!20,text width=2em,text height=1mm, inner sep=0]%
\tikzstyle{unpool} = [rectangle, draw, fill=orange!80,text width=2em,text height=1mm, inner sep=0]%
\begin{tikzpicture}[node distance = 2mm, auto]%
\node [data] (image) {in};%
\node [conv, below=1mm of image] (conv11) {};%
\node [conv, below of=conv11] (conv12) {};%
\node [pool, below of=conv12] (pool1) {};%
\node [conv, below of=pool1] (conv21) {};%
\node [conv, below of=conv21] (conv22) {};%
\node [pool, below of=conv22] (pool2) {};%
\node [conv, below of=pool2] (conv31) {};%
\node [conv, below of=conv31] (conv32) {};%
\node [conv, below of=conv32] (conv33) {};%
\node [pool, below of=conv33] (pool3) {};%
\node [conv, below of=pool3] (conv41) {};%
\node [conv, below of=conv41] (conv42) {};%
\node [conv, below of=conv42] (conv43) {};%
\node [pool, below of=conv43] (pool4) {};%
\node [conv, below of=pool4] (conv51) {};%
\node [conv, below of=conv51] (conv52) {};%
\node [conv, below of=conv52] (conv53) {};%
\node [unpool, right=1.2\thirdcolumn of pool1.center, anchor=center] (unpool1) {};%
\node [conv, right=1.2\thirdcolumn of conv12.center, anchor=center] (deconv1) {};%
\node [unpool, right=1.2\thirdcolumn of pool2.center, anchor=center] (unpool2) {};%
\node [conv, right=1.2\thirdcolumn of conv22.center, anchor=center] (deconv2) {};%
\node [unpool, right=1.2\thirdcolumn of pool3.center, anchor=center] (unpool3) {};%
\node [conv, right=1.2\thirdcolumn of conv33.center, anchor=center] (deconv3) {};%
\node [unpool, right=1.2\thirdcolumn of pool4.center, anchor=center] (unpool4) {};%
\node [conv, right=1.2\thirdcolumn of conv43.center, anchor=center] (deconv4) {};%
\node [data, right=1.2\thirdcolumn of image.center, anchor=center] (contours) {out};%
\draw[-{Latex[length=1mm,width=1mm]}] (image) -- (conv11);%
\draw[-{Latex[length=1mm,width=1mm]}] (conv11) -- (conv12);%
\draw[-{Latex[length=1mm,width=1mm]}] (conv12) -- (pool1);%
\draw[-{Latex[length=1mm,width=1mm]}] (pool1) -- (conv21);%
\draw[-{Latex[length=1mm,width=1mm]}] (conv21) -- (conv22);%
\draw[-{Latex[length=1mm,width=1mm]}] (conv22) -- (pool2);%
\draw[-{Latex[length=1mm,width=1mm]}] (pool2) -- (conv31);%
\draw[-{Latex[length=1mm,width=1mm]}] (conv31) -- (conv32);%
\draw[-{Latex[length=1mm,width=1mm]}] (conv32) -- (conv33);%
\draw[-{Latex[length=1mm,width=1mm]}] (conv33) -- (pool3);%
\draw[-{Latex[length=1mm,width=1mm]}] (pool3) -- (conv41);%
\draw[-{Latex[length=1mm,width=1mm]}] (conv41) -- (conv42);%
\draw[-{Latex[length=1mm,width=1mm]}] (conv42) -- (conv43);%
\draw[-{Latex[length=1mm,width=1mm]}] (conv43) -- (pool4);%
\draw[-{Latex[length=1mm,width=1mm]}] (pool4) -- (conv51);%
\draw[-{Latex[length=1mm,width=1mm]}] (conv51) -- (conv52);%
\draw[-{Latex[length=1mm,width=1mm]}] (conv52) -- (conv53);%
\draw[-{Latex[length=1mm,width=1mm]}] (conv53) -| (unpool4);%
\draw[-{Latex[length=1mm,width=1mm]}] (unpool4) -- (deconv4);%
\draw[-{Latex[length=1mm,width=1mm]}] (deconv4) -- (unpool3);%
\draw[-{Latex[length=1mm,width=1mm]}] (unpool3) -- (deconv3);%
\draw[-{Latex[length=1mm,width=1mm]}] (deconv3) -- (unpool2);%
\draw[-{Latex[length=1mm,width=1mm]}] (unpool2) -- (deconv2);%
\draw[-{Latex[length=1mm,width=1mm]}] (deconv2) -- (unpool1);%
\draw[-{Latex[length=1mm,width=1mm]}] (unpool1) -- (deconv1);%
\draw[-{Latex[length=1mm,width=1mm]}] (deconv1) -- (contours);%
\end{tikzpicture}
}}&
\vcenteredinclude{\subfloat[Holistic]{\setlength{\thirdcolumn}{.5cm}\label{fig:arch-ed-holistic}
\tikzstyle{data} = [rectangle, draw, fill=white!20,%
    text width=2em, text centered, rounded corners, minimum height=.5mm]%
\tikzstyle{convtext} = [draw=none, text width=2em, text centered, minimum height=.5mm]%
\tikzstyle{conv} = [rectangle, draw, fill=blue!20,%
    text width=2em, text centered, rounded corners=0.5mm, text height=1.5mm, inner sep=0]%
\tikzstyle{sum} = [circle, draw, fill=blue!20, inner sep=0]%
\tikzstyle{pool} = [rectangle, draw, fill=orange!20,text width=2em,text height=1mm, inner sep=0]%
\tikzstyle{unpool} = [rectangle, draw, fill=orange!80,text width=2em,text height=1mm, inner sep=0]%
\begin{tikzpicture}[node distance = 2mm, auto]
\node [data] (image) {in};%
\node [conv, below=1mm of image] (conv11) {};%
\node [conv, below of=conv11] (conv12) {};%
\node [pool, below of=conv12] (pool1) {};%
\node [conv, below of=pool1] (conv21) {};%
\node [conv, below of=conv21] (conv22) {};%
\node [pool, below of=conv22] (pool2) {};%
\node [conv, below of=pool2] (conv31) {};%
\node [conv, below of=conv31] (conv32) {};%
\node [conv, below of=conv32] (conv33) {};%
\node [pool, below of=conv33] (pool3) {};%
\node [conv, below of=pool3] (conv41) {};%
\node [conv, below of=conv41] (conv42) {};%
\node [conv, below of=conv42] (conv43) {};%
\node [pool, below of=conv43] (pool4) {};%
\node [conv, below of=pool4] (conv51) {};%
\node [conv, below of=conv51] (conv52) {};%
\node [conv, below of=conv52] (conv53) {};%
\node [conv, right=\thirdcolumn of conv12] (deconv1) {};%
\node [conv, right=\thirdcolumn of conv22] (deconv2) {};%
\node [conv, right=\thirdcolumn of conv33] (deconv3) {};%
\node [conv, right=\thirdcolumn of conv43] (deconv4) {};%
\node [conv, right=\thirdcolumn of conv53] (deconv5) {};%
\node [unpool, above of=deconv2] (unpool2) {};%
\node [unpool, above of=deconv3] (unpool3) {};%
\node [unpool, above of=deconv4] (unpool4) {};%
\node [unpool, above of=deconv5] (unpool5) {};%
\node [convtext, above of=unpool3] (unpool3text) {$4\times$};%
\node [convtext, above of=unpool4] (unpool4text) {$8\times$};%
\node [convtext, above of=unpool5] (unpool5text) {$16\times$};%
\node [data, right=1.5\thirdcolumn of image] (contours) {out};%
\draw[-{Latex[length=1mm,width=1mm]}] (image) -- (conv11);%
\draw[-{Latex[length=1mm,width=1mm]}] (conv11) -- (conv12);%
\draw[-{Latex[length=1mm,width=1mm]}] (conv12) -- (pool1);%
\draw[-{Latex[length=1mm,width=1mm]}] (pool1) -- (conv21);%
\draw[-{Latex[length=1mm,width=1mm]}] (conv21) -- (conv22);%
\draw[-{Latex[length=1mm,width=1mm]}] (conv22) -- (pool2);%
\draw[-{Latex[length=1mm,width=1mm]}] (pool2) -- (conv31);%
\draw[-{Latex[length=1mm,width=1mm]}] (conv31) -- (conv32);%
\draw[-{Latex[length=1mm,width=1mm]}] (conv32) -- (conv33);%
\draw[-{Latex[length=1mm,width=1mm]}] (conv33) -- (pool3);%
\draw[-{Latex[length=1mm,width=1mm]}] (pool3) -- (conv41);%
\draw[-{Latex[length=1mm,width=1mm]}] (conv41) -- (conv42);%
\draw[-{Latex[length=1mm,width=1mm]}] (conv42) -- (conv43);%
\draw[-{Latex[length=1mm,width=1mm]}] (conv43) -- (pool4);%
\draw[-{Latex[length=1mm,width=1mm]}] (pool4) -- (conv51);%
\draw[-{Latex[length=1mm,width=1mm]}] (conv51) -- (conv52);%
\draw[-{Latex[length=1mm,width=1mm]}] (conv52) -- (conv53);%
\draw[-{Latex[length=1mm,width=1mm]}] (deconv1) -| (contours);%
\draw[-{Latex[length=1mm,width=1mm]}] (unpool2) -| (contours);%
\draw[-{Latex[length=1mm,width=1mm]}] (unpool3) -| (contours);%
\draw[-{Latex[length=1mm,width=1mm]}] (unpool4) -| (contours);%
\draw[-{Latex[length=1mm,width=1mm]}] (unpool5) -| (contours);%
\draw[-{Latex[length=1mm,width=1mm]}] (conv12) -- (deconv1);%
\draw[-{Latex[length=1mm,width=1mm]}] (conv22) -- (deconv2);%
\draw[-{Latex[length=1mm,width=1mm]}] (conv33) -- (deconv3);%
\draw[-{Latex[length=1mm,width=1mm]}] (conv43) -- (deconv4);%
\draw[-{Latex[length=1mm,width=1mm]}] (conv53) -- (deconv5);%
\draw[-{Latex[length=1mm,width=1mm]}] (deconv2) -- (unpool2);%
\draw[-{Latex[length=1mm,width=1mm]}] (deconv3) -- (unpool3);%
\draw[-{Latex[length=1mm,width=1mm]}] (deconv4) -- (unpool4);%
\draw[-{Latex[length=1mm,width=1mm]}] (deconv5) -- (unpool5);%
\end{tikzpicture}
}}&
\vcenteredinclude{\subfloat[Residual]{\label{fig:arch-ed-residual}
\tikzstyle{data} = [rectangle, draw, fill=white!20,%
    text width=2em, text centered, rounded corners, minimum height=.5mm]%
\tikzstyle{conv} = [rectangle, draw, fill=blue!20,%
    text width=2em, text centered, rounded corners=0.5mm, text height=1.5mm, inner sep=0]%
\tikzstyle{sum} = [circle, draw, fill=blue!20, text height=1mm, inner sep=0]%
\tikzstyle{relu} = [rectangle, draw, fill=red!20,minimum width=3em]%
\tikzstyle{pool} = [rectangle, draw, fill=orange!20,text width=2em,text height=1mm, inner sep=0]%
\tikzstyle{unpool} = [rectangle, draw, fill=orange!80,text width=2em,text height=1mm, inner sep=0]%
\begin{tikzpicture}[node distance = 2mm, auto]
\node [data] (image) {in};
\node [conv, below=1mm of image] (conv11) {};%
    \node [conv, below of=conv11] (conv12) {};%
    \node [pool, below of=conv12] (pool1) {};%
    \node [conv, below of=pool1] (conv21) {};%
    \node [conv, below of=conv21] (conv22) {};%
    \node [pool, below of=conv22] (pool2) {};
    \node [conv, below of=pool2] (conv31) {};%
    \node [conv, below of=conv31] (conv32) {};%
    \node [conv, below of=conv32] (conv33) {};%
    \node [pool, below of=conv33] (pool3) {};%
    \node [conv, below of=pool3] (conv41) {};%
    \node [conv, below of=conv41] (conv42) {};%
    \node [conv, below of=conv42] (conv43) {};%
    \node [pool, below of=conv43] (pool4) {};%
    \node [conv, below of=pool4] (conv51) {};%
    \node [conv, below of=conv51] (conv52) {};%
    \node [conv, below of=conv52] (conv53) {};%
\node [unpool, right=1.2\thirdcolumn of pool1.center, anchor=center] (unpool1) {};%
\node [conv, right=1.2\thirdcolumn of conv12.center, anchor=center] (deconv1) {};%
\node [unpool, right=1.2\thirdcolumn of pool2.center, anchor=center] (unpool2) {};%
\node [conv, right=1.2\thirdcolumn of conv22.center, anchor=center] (deconv2) {};%
\node [unpool, right=1.2\thirdcolumn of pool3.center, anchor=center] (unpool3) {};%
\node [conv, right=1.2\thirdcolumn of conv33.center, anchor=center] (deconv3) {};%
\node [unpool, right=1.2\thirdcolumn of pool4.center, anchor=center] (unpool4) {};%
\node [conv, right=1.2\thirdcolumn of conv43.center, anchor=center] (deconv4) {};%
\node [data, right=1.2\thirdcolumn of image.center, anchor=center] (contours) {out};%
\draw[-{Latex[length=1mm,width=1mm]}] (image) -- (conv11);%
\draw[-{Latex[length=1mm,width=1mm]}] (conv11) -- (conv12);%
\draw[-{Latex[length=1mm,width=1mm]}] (conv12) -- (pool1);%
\draw[-{Latex[length=1mm,width=1mm]}] (pool1) -- (conv21);%
\draw[-{Latex[length=1mm,width=1mm]}] (conv21) -- (conv22);%
\draw[-{Latex[length=1mm,width=1mm]}] (conv22) -- (pool2);%
\draw[-{Latex[length=1mm,width=1mm]}] (pool2) -- (conv31);%
\draw[-{Latex[length=1mm,width=1mm]}] (conv31) -- (conv32);%
\draw[-{Latex[length=1mm,width=1mm]}] (conv32) -- (conv33);%
\draw[-{Latex[length=1mm,width=1mm]}] (conv33) -- (pool3);%
\draw[-{Latex[length=1mm,width=1mm]}] (pool3) -- (conv41);%
\draw[-{Latex[length=1mm,width=1mm]}] (conv41) -- (conv42);%
\draw[-{Latex[length=1mm,width=1mm]}] (conv42) -- (conv43);%
\draw[-{Latex[length=1mm,width=1mm]}] (conv43) -- (pool4);%
\draw[-{Latex[length=1mm,width=1mm]}] (pool4) -- (conv51);%
\draw[-{Latex[length=1mm,width=1mm]}] (conv51) -- (conv52);%
\draw[-{Latex[length=1mm,width=1mm]}] (conv52) -- (conv53);%
\draw[-{Latex[length=1mm,width=1mm]}] (conv53) -| (unpool4);%
\draw[-{Latex[length=1mm,width=1mm]}] (unpool4) -- (deconv4);%
\draw[-{Latex[length=1mm,width=1mm]}] (deconv4) -- (unpool3);%
\draw[-{Latex[length=1mm,width=1mm]}] (unpool3) -- (deconv3);%
\draw[-{Latex[length=1mm,width=1mm]}] (deconv3) -- (unpool2);%
\draw[-{Latex[length=1mm,width=1mm]}] (unpool2) -- (deconv2);%
\draw[-{Latex[length=1mm,width=1mm]}] (deconv2) -- (unpool1);%
\draw[-{Latex[length=1mm,width=1mm]}] (unpool1) -- (deconv1);%
\draw[-{Latex[length=1mm,width=1mm]}] (deconv1) -- (contours);%
\draw[-{Latex[length=1mm,width=1mm]}] (conv12) -- (deconv1);%
\draw[-{Latex[length=1mm,width=1mm]}] (conv22) -- (deconv2);%
\draw[-{Latex[length=1mm,width=1mm]}] (conv33) -- (deconv3);%
\draw[-{Latex[length=1mm,width=1mm]}] (conv43) -- (deconv4);%
\end{tikzpicture}
}}\\
\end{tabular}%

\subfloat{\setlength\mywidth{.15cm}
\tikzstyle{data} = [rectangle, draw, fill=white!20,%
    text width=2em, text centered, rounded corners, minimum height=.5mm]%
\tikzstyle{conv} = [rectangle, draw, fill=blue!20,%
    text width=2em, text centered, rounded corners=0.5mm, text height=1.5mm, inner sep=0]%
\tikzstyle{sum} = [circle, draw, fill=blue!20, text height=1mm, inner sep=0]%
\tikzstyle{relu} = [rectangle, draw, fill=red!20,minimum width=3em]%
\tikzstyle{pool} = [rectangle, draw, fill=orange!20,text width=2em,text height=1mm, inner sep=0]%
\tikzstyle{unpool} = [rectangle, draw, fill=orange!80,text width=2em,text height=1mm, inner sep=0]%
\begin{tikzpicture}
\node [data] (image) {in};%
\node [data, below=1mm of image] (out1) {out};%
\node [conv, right=3cm of image] (conv1) {};%
\node [pool, below=3mm of conv1] (pool1) {};%
\node [unpool, below=3mm of pool1] (unpool1) {};%
\node[draw=none,fill=none, align=left, right=\mywidth of image] {Input};%
\node[draw=none,fill=none, align=left, right=\mywidth of out1] {Conv+Sigmoid};%
\node[draw=none,fill=none, align=left, right=\mywidth of conv1] (convname) {Concat+Conv+ReLU};%
\node[draw=none,fill=none, align=left, right=\mywidth of pool1] (poolname) {Spatial pooling ($.5\times$)};%
\node[draw=none,fill=none, align=left, right=\mywidth of unpool1] (unpoolname) {Spatial unpooling ($2\times$)};%
\end{tikzpicture}%
}%
\caption{State-of-the-art decoding strategies for boundary detection, using a VGG16-based \citep{vgg15} encoder. Best viewed in color.}
\label{fig:ed-architectures}
\end{figure}

First introduced for single-task setups, such as semantic segmentation \citep{segnet17} and instance boundary detection \citep{cedn16}, encoder-decoder networks are designed to infer pixel labels despite the spatial resolution loss when encoding object-level semantics. Specifically, the encoder produces deep hierarchical features, then the decoder gradually outputs a probability map using symmetric unpooling stages (\cf{} Figure \ref{fig:arch-ed-sequential}). However, in a sequential encoder-decoder, the pixel labels are inferred only from the last encoder feature maps, where the information is the most spatially compressed. Instead, a multiscale view can be given to the decoder through holistically-nested connections (Figure \ref{fig:arch-ed-holistic}) \citep{RCF17, cob16, hed15}. Nevertheless, such a late fusion requires to upsample all the latent representations to the image resolution. A progressive multiscale decoding through scale-specific skip connections between the encoder and decoder (\cf{} Figure \ref{fig:arch-ed-residual}) has consequently proved superior \citep{DengSLWL18,CED17,unet15}. Indeed, at each decoding stage, the lower-resolution but higher-level semantics are merged with the higher-resolution information lost after pooling the encoder features of the current scale. Note that in application contexts requiring high resolutions, residual encoder-decoder networks may suffer from checkerboard artifacts, also referred to as the gridding effect \citep{LiuSHL18,Guan18,Shi16shuffle}. Interestingly, coupling residual encoder-decoder networks via cross-network skip connections helps to refine the localization of visual landmarks \citep{TangPGWZM18}.

\paragraph{Multi-task learning}

Sharing representations in learning multiple tasks generally enables to capture more generalizable invariants. In the context of semantic segmentation, \citep{LuoWLW17} proposed to merge local and global semantics through a dual-task training, by jointly decoding pixel labels and inferring image labels. Image-level classification is however unfeasible in a category-agnostic problem, although detecting instance boundaries and inter-instance occlusions require global cues as well. For pixel multi-labeling, various strategies of knowledge sharing have been explored, such as progressive layer splitting \citep{MisraSGH16stitch}, dynamic task loss weighing \citep{KendallGC18}, skip connection-like attention masks between a shared network and task-specific ones \citep{Liu19mtan}. These works are however focused on best learning task-shared and task-specific features to excel in every task. In this work, we are rather interested in exploiting an ordinal task decomposition to enforce a learning path, but not to excel in every subtask.

\subsection{Datasets}

\paragraph{Oriented boundary detection}

Monocular occlusion-aware boundary detection raised interest with the BSDS Border Ownership dataset (BSDS-BOW) \citep{Ren2006}, which contains 200 real images from the BSDS500 dataset \citep{MartinFTM01}, manually annotated with object part-level oriented contours. As state-of-the-art FCNs require more training data, \citep{doc16} presented the PASCAL Instance Occlusion Dataset (PIOD), consisting of 10,100 manually annotated real images from the PASCAL VOC Segmentation dataset \citep{pascalvoc12}. Despite their challenging intra-class variability, the images contain few instances and inter-instance occlusions (\cf{} Figure \ref{fig:datasets}).

\paragraph{Amodal segmentation}

\citep{Qi2019amodal,FollmannKHKB19, Zhu17amodal} also released datasets of real images, respectively the KITTI INStance dataset (KINS), the Densely Segmented Supermarket Amodal dataset (D2SA) and the COCO Amodal dataset (COCOA), that are subsets of larger datasets for box proposal-based instance segmentation, respectively KITTI \citep{GeigerLSU13}, COCO \citep{mscoco14} and D2S \citep{D2S18}, manually augmented with ground-truth amodal annotations. However, overcrowded scenes are also not represented in these datasets. Moreover, the ground-truth amodal annotations result from guesses, thereby introducing human biases in the learning process.

\paragraph{Synthetic images}

Synthetic datasets have emerged in various contexts as they offer rich multimodal annotations from fully controlled environments \citep{SceneNet17,Synthia16,GaidonWCV16virtualkitti,dol18,bregier2017iccv}. Yet, in these datasets, dense homogeneous layouts have received little attention. Proposed for evaluating pose detection and estimation, the Sil\'eane dataset \citep{bregier2017iccv} consists of top-view depth images of identical rigid instances in piles. Similarly, \citep{dol18} suggested synthetic depth maps of scanned objects instantiated in bulk. These synthetic datasets are however generated only for depth-based perception and elude the learning from a single RGB image.

\section{Proposed Model}
\label{sec:proposedmodel}

In this section, we first describe the proposed multicameral structuring for occlusion-aware instance-wise attention. Second, we detail the associated loss function.

\subsection{Problem Statement}

We aim to approximate a mapping between RGB images and instance-sensitive segmentations. As a showcase scenario, we look for sets of non-overlapping connected pixel clusters that represent unoccluded instances (see Figure \ref{fig:showcase}). Formally, let $\mathcal{X}$ be our set of $|\mathcal{X}|\in\mathbb{N}^\star$ RGB images, and $\mathcal{P}$ the set of pixel locations. For an image of width $W\in\mathbb{N}^\star$ and height $H\in\mathbb{N}^\star$, we write $P=W\times H$, and $\mathcal{P}=\{1,...,W\}\times\{1,...,H\}$. We aim at approximating a function $f$ defined as follows:
\begin{equation}\label{eq:ftoapprox}
f \colon \mathcal{X} \to \{0,1\}^P,\ X \mapsto Y.
\end{equation}
Given an image $X^n\in\mathcal{X}$, a pixel $\mathbf{p}\in\mathcal{P}$ is fired, \ie{} $Y_\mathbf{p}^n=1$ if it belongs to an unoccluded instance.

\subsection{Proposed Architecture}
\label{sec:proposedarch}

Generally, a residual encoder-decoder (RED) network is a sequence of scale-specific encoding feature transforms $E_{s}$, and residual decoding feature transforms $D_s$ such that:
\begin{equation}
\mathbf{x}_{s} = E_{s}(\mathbf{x}_{s-1}),
\end{equation}
\begin{equation}
\mathbf{y}_{s}= D_{s}(\mathbf{y}_{s+1}, \mathbf{x}_{s}),
\end{equation}
where $\mathbf{x}_{s}$ and $\mathbf{y}_{s}$ are the latent image representations at the resolution level $s$ in the encoder and decoder respectively. For example, $\mathbf{x}_1 = E_1(X)$. If we note $E=\{E_s\}_{s\in\{1,...,S\}}$ and $D=\{D_s\}_{s\in\{1,...,S\}}$ then a RED network is a sequence $[E,D]$. In a RED network, the decoder aims to gradually upsample the deep representations of the encoder. This is however unsufficient to discriminate between instances of the same object.

\begin{figure}[t]
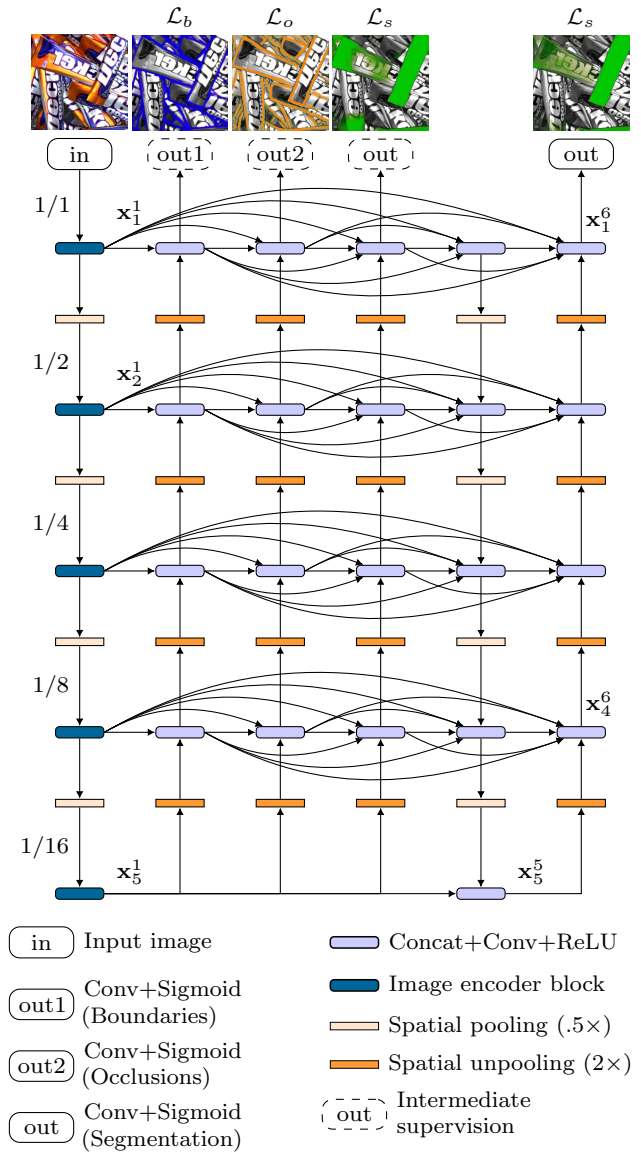

\centering

\caption{Proposed multicameral structuring with ordinal intermediate supervisions (MC6$\dagger$) for monocular attention to unoccluded instances. Best viewed in color.}
\label{fig:multicameralarch}
\end{figure}

By contrast, a multicameral (MC) network is a sequence of $T$ residual decoder and encoder-decoder units, densely connected through resolution-wise skip connections, to approximate a more complex decoding function (see Figure \ref{fig:multicameralarch}). If we define encoders and decoders as multiscale feature transforms, then a multicameral structuring is a matrix-like layout of latent representations at $S$ different resolutions. Each row thereby conveys high-level semantics at a fixed resolution. As the starting point is an image, the first element is a deep encoder based on a common backbone, for example a VGG16 encoder \citep{vgg15}. The first three decoders in cascade gradually recover the instance boundaries, the occluding boundary sides, and the segmentation outlining the unoccluded instances respectively. These ordinal units aim to structure the decoding process. It also encourages subtask-specific feature reuse: an occluding boundary side is expected to be near an instance boundary, and a pixel in an unoccluded instance is expected to be isotropically surrounded by occluding boundary sides. After these decoders, an encoder-decoder unit refines the segmentation.

Formally, let $\mathbf{x}_{s}^{t}$ be the latent representation at the row $s\in\{1,...,S\}$ and column $t\in\{1,...,T\}$. Then an encoding transform $E_{s}^{t}$ and a decoding transform $D_{s}^{t}$ at this position are defined respectively as:
\begin{equation}
\mathbf{x}_{s}^{t} = E_{s}^{t}(\mathbf{x}_{s-1}^{t}, \mathbf{x}_{s}^{t-1}, ..., \mathbf{x}_{s}^{1}),
\end{equation}
\begin{equation}
\mathbf{x}_{s}^{t} = D_{s}^{t}(\mathbf{x}_{s+1}^{t}, \mathbf{x}_{s}^{t-1}, ..., \mathbf{x}_{s}^{1}).
\end{equation}
If we note $E^t=\{E_s^t\}_{s\in\{1,...,S\}}$ and $D^t=\allowbreak\{D_s^t\}_{s\in\{1,...,S\}}$, then a multicameral design is the sequence $[E^1,D^2,D^3,\allowbreak D^4,E^5,D^6]$. In the following, we refer to a multicameral structure of $T$ columns as MC$T$. For examples, MC4 $= [E^1,D^2,D^3,D^4]$, MC3 $= [E^1,D^2,D^3]$, and RED $=$ MC2 $= [E^1,D^1]$.

\paragraph{Feature transforms}

\begin{figure}[t]
\centering
\fontsize{8}{9}\selectfont
\setlength\tabcolsep{1pt}%
\setlength\mywidth{.1cm}
\setlength\secondcolumn{.45cm}
\setlength\thirdcolumn{.3cm}
\begin{tabular}{c}
\vcenteredinclude{
\tikzstyle{img} = [draw=none,inner sep=0pt]%
\tikzstyle{conv} = [rectangle, draw, fill=blue!20,%
    text width=2em, text centered, rounded corners=0.5mm, text height=1.5mm, inner sep=0]%
\tikzstyle{hideconv} = [draw=white,fill=white,minimum width=1cm]
\tikzstyle{hideconv2} = [conv,draw=white,fill=white]
\tikzstyle{conv2} = [conv,fill=blue!60]%
\tikzstyle{conv3} = [conv,fill=blue!90!black]%
\tikzstyle{data} = [conv, rounded corners=2mm,fill=white!20,text width=2em,text centered, text height=.8em, text depth=.1em,inner sep=2pt]%
\tikzstyle{convtext} = [draw=none, text width=2em,text centered, text height=.8em, inner sep=1pt]
\tikzstyle{convdim} = [draw=none, text width=3em,text centered, text height=.8em, inner sep=1pt]
\tikzstyle{pool} = [rectangle, draw, fill=orange!20,minimum width=2em,text height=.2em, inner sep=0pt, line width=.03em,anchor=center]%
\tikzstyle{hidepool} = [rectangle, draw=white, fill=white,minimum width=2em,text height=.2em, inner sep=0pt, line width=.03em,anchor=center]%
\tikzstyle{unpool} = [pool,fill=orange!80]%
\tikzstyle{convarrow} = [-{Latex[length=2mm,width=1mm]},line width=.2mm]%
\tikzstyle{convline} = [-,line width=.4mm]%
\tikzstyle{poolarrow} = [convarrow,dashed]%
\tikzstyle{scaleann}=[draw=none,text width=1.4em,align=left,midway,left=1pt,font={\fontsize{10}{9}\selectfont}]%
\def\outIB{30}
\def\outIC{\outIB}%
\def\outID{\outIB}%
\def\outIE{\outIB}%
\def\inIB{150}
\def\inIC{\inIB}%
\def\inID{\inIB}%
\def\inIE{\inIB}%
\begin{tikzpicture}
\node [hideconv] (begin) {$N\times C_nHW$};%
\node [hideconv2, right=\thirdcolumn of begin] (conv1) {};
\node [hideconv2, right=\secondcolumn of conv1] (conv2) {};%
\node [conv, right=\secondcolumn of conv2] (conv3) {};%
\node [hideconv2, right=\secondcolumn of conv3] (conv4) {};%
\node [hideconv2, right=\secondcolumn of conv4] (conv5) {};
\node [hideconv, right=\thirdcolumn of conv5] (end) {$CHW$};
\draw[convarrow] (begin) -- (conv3);%
\draw[convarrow] (conv3) -- (end);%
\node [convtext, below=\mywidth of conv3] (convtext3) {$5\times5$\\d=1};
\end{tikzpicture}
}\\
\vspace{-.2cm}\\
(a) Convolution.\\
\\
\vcenteredinclude{%
\setlength\mywidthbis{.45cm}%
\tikzstyle{img} = [draw=none,inner sep=0pt]%
\tikzstyle{conv} = [rectangle, draw, fill=blue!20,%
    text width=2em, text centered, rounded corners=0.5mm, text height=1.5mm, inner sep=0]%
\tikzstyle{hideconv} = [draw=white,fill=white,minimum width=1cm]
\tikzstyle{hideconv2} = [conv,draw=white,fill=white]
\tikzstyle{conv2} = [conv,fill=blue!60]%
\tikzstyle{conv3} = [conv,fill=blue!90!black]%
\tikzstyle{data} = [conv, rounded corners=2mm,fill=white!20,text width=2em,text centered, text height=.8em, text depth=.1em,inner sep=2pt]%
\tikzstyle{convtext} = [draw=none, text width=2em,text centered, text height=.8em, inner sep=1pt]
\tikzstyle{convdim} = [draw=none, text width=3em,text centered, text height=.8em, inner sep=1pt]
\tikzstyle{pool} = [rectangle, draw, fill=orange!20,minimum width=2em,text height=.2em, inner sep=0pt, line width=.03em,anchor=center]%
\tikzstyle{hidepool} = [rectangle, draw=white, fill=white,minimum width=2em,text height=.2em, inner sep=0pt, line width=.03em,anchor=center]%
\tikzstyle{unpool} = [pool,fill=orange!80]%
\tikzstyle{convarrow} = [-{Latex[length=2mm,width=1mm]},line width=.2mm]%
\tikzstyle{convline} = [-,line width=.4mm]%
\tikzstyle{poolarrow} = [convarrow,dashed]%
\tikzstyle{scaleann}=[draw=none,text width=1.4em,align=left,midway,left=1pt,font={\fontsize{10}{9}\selectfont}]%
\begin{tikzpicture}
\node [hideconv] (begin) {$N\times C_nHW$};%
\node [hideconv, above=\mywidth of begin.north east, anchor=south east] (coords) {$2HW$};%
\node [hideconv2, right=\thirdcolumn of begin] (conv1) {};
\node [hideconv2, right=\secondcolumn of conv1] (conv2) {};%
\node [conv, right=\secondcolumn of conv2] (conv3) {};%
\node [hideconv2, right=\secondcolumn of conv3] (conv4) {};%
\node [hideconv2, right=\secondcolumn of conv4] (conv5) {};
\node [hideconv, right=\thirdcolumn of conv5] (end) {$CHW$};
\draw[convarrow] (begin) -- (conv3);%
\draw[convarrow] (coords) to [out=0,in=-180] (conv3);%
\draw[convarrow] (conv3) -- (end);%
\node [convtext, draw=none,text width=8em,above right=\mywidthbis and -7em of conv3.north] (coordstext1) {\textit{Pixel coordinates}};
\node [convtext, below=\mywidth of conv3] (convtext3) {$5\times5$\\d=1};
\end{tikzpicture}
}\\
\vspace{-.2cm}\\
(b) Coordinate-aware convolution (Coords).\\
\\
\vcenteredinclude{%
\setlength\mywidthbis{.35cm}%
\tikzstyle{img} = [draw=none,inner sep=0pt]
\tikzstyle{conv} = [rectangle, draw, fill=blue!20,%
    text width=2em, text centered, rounded corners=0.5mm, text height=1.5mm, inner sep=0]%
\tikzstyle{hideconv} = [draw=white,fill=white,minimum width=1cm]
\tikzstyle{conv2} = [conv,fill=blue!60]%
\tikzstyle{conv3} = [conv,fill=blue!90!black]%
\tikzstyle{data} = [conv, rounded corners=2mm,fill=white!20,text width=2em,text centered, text height=.8em, text depth=.1em,inner sep=2pt]%
\tikzstyle{convtext} = [draw=none, text width=2em,text centered, text height=.8em, inner sep=1pt]
\tikzstyle{convdim} = [draw=none, text width=3em,text centered, text height=.8em, inner sep=1pt]
\tikzstyle{pool} = [rectangle, draw, fill=orange!20,minimum width=2em,text height=.2em, inner sep=0pt, line width=.03em,anchor=center]%
\tikzstyle{hidepool} = [rectangle, draw=white, fill=white,minimum width=2em,text height=.2em, inner sep=0pt, line width=.03em,anchor=center]%
\tikzstyle{unpool} = [pool,fill=orange!80]%
\tikzstyle{convarrow} = [-{Latex[length=2mm,width=1mm]},line width=.2mm]%
\tikzstyle{convline} = [-,line width=.4mm]%
\tikzstyle{poolarrow} = [convarrow,dashed]%
\tikzstyle{scaleann}=[draw=none,text width=1.4em,align=left,midway,left=1pt,font={\fontsize{10}{9}\selectfont}]%
\def\outIB{40}
\def\inIB{150}
\begin{tikzpicture}
\node [hideconv] (begin) {$N\times C_nHW$};%
\node [conv, right=\thirdcolumn of begin] (conv1) {};
\node [conv, right=\secondcolumn of conv1] (conv2) {};%
\node [conv, right=\secondcolumn of conv2] (conv3) {};%
\node [conv, right=\secondcolumn of conv3] (conv4) {};%
\node [conv, right=\secondcolumn of conv4] (conv5) {};
\node [hideconv, right=\thirdcolumn of conv5] (end) {$CHW$};
\draw[convarrow] (begin) -- (conv1);%
\draw[convarrow] (conv1.east) to [out=\outIB,in=\inIB] (conv2.north);%
\draw[convarrow] (conv1.east) to [out=\outIB,in=\inIB] (conv3.north);%
\draw[convarrow] (conv1.east) to [out=\outIB,in=\inIB] (conv4.north);%
\draw[convarrow] (conv2.east) to [out=\outIB,in=\inIB] (conv5.north);%
\draw[convarrow] (conv3.east) to [out=\outIB,in=\inIB] (conv5.north);%
\draw[convarrow] (conv4.east) to [out=\outIB,in=\inIB] (conv5.north);%
\draw[convarrow] (conv5) -- (end);%
\node [convtext, below=\mywidth of conv1] (convtext1) {$1\times1$};
\node [convtext, below=\mywidth of conv2] (convtext2) {$5\times5$\\d=1};
\node [convtext, below=\mywidth of conv3] (convtext3) {$5\times5$\\d=3};
\node [convtext, below=\mywidth of conv4] (convtext4) {$5\times5$\\d=6};
\node [convtext, below=\mywidth of conv5] (convtext5) {$1\times1$};
\node [convdim, above=\mywidthbis of conv1] (convdim1) {$CHW$};
\node [convdim, text width=5em, above right=\mywidthbis and -2em of conv5] (convdim1) {$3\times CHW$};
\end{tikzpicture}
}\\
\vspace{-.2cm}\\
(c) Atrous spatial pyramid (Atrous).\\
\\
\vcenteredinclude{%
\setlength\mywidthbis{.35cm}%
\setlength\fourthcolumn{.7cm}%
\tikzstyle{img} = [draw=none,inner sep=0pt]
\tikzstyle{conv} = [rectangle, draw, fill=blue!20,%
    text width=2em, text centered, rounded corners=0.5mm, text height=1.5mm, inner sep=0]%
\tikzstyle{hideconv} = [draw=white,fill=white,minimum width=1cm]
\tikzstyle{hideconv2} = [conv,draw=white,fill=white]
\tikzstyle{conv2} = [conv,fill=blue!50!red]%
\tikzstyle{conv3} = [conv,fill=blue!90!black]%
\tikzstyle{data} = [conv, rounded corners=2mm,fill=white!20,text width=2em,text centered, text height=.8em, text depth=.1em,inner sep=2pt]%
\tikzstyle{convtext} = [draw=none, text width=2em,text centered, text height=.8em, inner sep=1pt]
\tikzstyle{convdim} = [draw=none, text width=3em,text centered, text height=.8em, inner sep=1pt]
\tikzstyle{pool} = [rectangle, draw, fill=orange!20,minimum width=2em,text height=.2em, inner sep=0pt, line width=.03em,anchor=center]%
\tikzstyle{hidepool} = [rectangle, draw=white, fill=white,minimum width=2em,text height=.2em, inner sep=0pt, line width=.03em,anchor=center]%
\tikzstyle{unpool} = [pool,fill=orange!80]%
\tikzstyle{convarrow} = [-{Latex[length=2mm,width=1mm]},line width=.2mm]%
\tikzstyle{convline} = [-,line width=.4mm]%
\tikzstyle{poolarrow} = [convarrow,dashed]%
\tikzstyle{scaleann}=[draw=none,text width=1.4em,align=left,midway,left=1pt,font={\fontsize{10}{9}\selectfont}]%
\def\outIB{30}
\def\inIB{150}
\begin{tikzpicture}
\node [hideconv] (begin) {$N\times C_nHW$};%
\node [conv, right=\thirdcolumn of begin] (conv1) {};
\node [conv, right=\secondcolumn of conv1] (conv2) {};%
\node [conv, minimum height=1.1em, text height=.65em, fill=white, right=\secondcolumn of conv2] (conv3) {$\odot$};%
\node [conv, right=\secondcolumn of conv3] (conv4) {};%
\node [hideconv2, right=\secondcolumn of conv4] (conv5) {};
\node [conv2, above=\fourthcolumn of conv2] (convup2) {};%
\node [conv, minimum height=1.1em, text height=.5em, fill=white, right=\secondcolumn of convup2] (convup3) {$\sigma$};%
\node [hideconv, right=\thirdcolumn of conv5] (end) {$CHW$};
\draw[convarrow] (begin) -- (conv1);%
\draw[convarrow] (conv1) -- (conv2);%
\draw[convarrow] (conv2) -- (conv3);%
\draw[convarrow] (conv1) to [out=0,in=-180] (convup2);
\draw[convarrow] (convup2) -- (convup3);%
\draw[convarrow] (convup3) -- (conv3);%
\draw[convarrow] (conv3) -- (conv4);%
\draw[convarrow] (conv4) -- (end);%
\node [convtext, below=\mywidth of conv1] (convtext1) {$1\times1$};
\node [convtext, below=\mywidth of convup2] (convup2text1) {$1\times1$};
\node [convtext, below=\mywidth of conv2] (convtext2) {$5\times5$\\d=1};
\node [convtext, below=\mywidth of conv4] (convtext4) {$5\times5$\\d=1};
\node [convdim, above=\mywidthbis of conv1] (convdim1) {$CHW$};
\node [convdim, above right=\mywidthbis and 0cm of conv3] {$CHW$};
\end{tikzpicture}%
}\\
\vspace{-.2cm}\\
(d) Attention branch (MTAN).\\
\\
\vcenteredinclude{%
\setlength\secondcolumn{.1cm}%
\setlength\thirdcolumn{4cm}%
\tikzstyle{img} = [draw=none,inner sep=0pt]%
\tikzstyle{conv} = [rectangle, draw, fill=blue!20,%
    text width=2em, text centered, rounded corners=0.5mm, text height=1.5mm, inner sep=0]%
\tikzstyle{hideconv} = [draw=white,fill=white,minimum width=1cm]
\tikzstyle{hideconv2} = [conv,draw=white,fill=white]
\tikzstyle{conv2} = [conv,fill=blue!50!red]%
\tikzstyle{conv3} = [conv,fill=blue!90!black]%
\tikzstyle{data} = [conv, rounded corners=2mm,fill=white!20,text width=2em,text centered, text height=.8em, text depth=.1em,inner sep=2pt]%
\tikzstyle{convtext} = [draw=none, text width=6em,text centered, text height=.8em, inner sep=1pt]
\tikzstyle{convdim} = [draw=none, text width=10em,text centered, text height=.8em, inner sep=1pt]
\begin{tikzpicture}
\node [conv] (conv1) {};
\node [convtext,below=\mywidth of conv1] (convtext1) {kernel size\\dilation rate};
\node [convdim,align=left,right=\secondcolumn of conv1] (convtext12) {Concat+Conv+ReLU};
\node [conv2,right=\thirdcolumn of conv1] (conv2) {};
\node [convdim,text width=3em,align=left,right=\secondcolumn of conv2] (convtext2) {Conv};
\node [convtext,below=\mywidth of conv2] (convtext22) {kernel size};
\end{tikzpicture}%
}\\
\end{tabular}
\caption{State-of-the-art node-level mechanisms for learning a contextual representation of size $CHW$ from $N$ latent representations of size $C_nHW$ respectively, where $n\in\{1,...,N\}$.(a) Soft feature sampling using gradient-based weights. (b) Features are attached to global pixel coordinates before sampling \citep{Liu18coordconv,NovotnyALV18}. (c) Longer-range sampling using aggregated dilated convolutions \citep{deeplabv3+18,WangCYLHHC18,YuK16}. (d) Soft feature sampling using inferred masks \citep{Liu19mtan}.}
\label{fig:nodelevelarchs}
\end{figure}
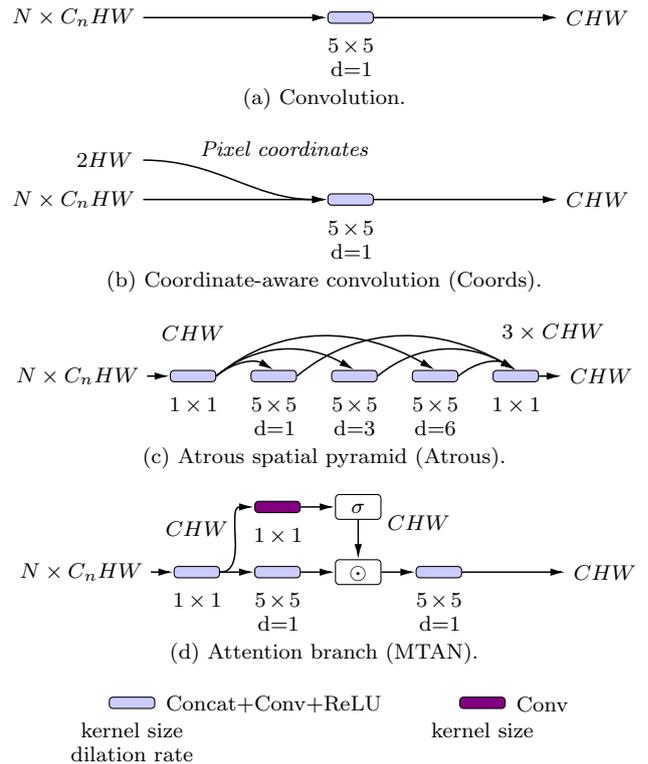

In the decoder and encoder-decoder units except the first encoder, the default encoding and decoding feature transforms consist of three operations: (1) concatenate the inputs along the channel axis (Concat); (2) apply a pixel-wise affine transformation (Conv); (3) apply a non-linear activation (ReLU). Only the transforms $E_s^1$ in the first encoder consists of more operations, such as sequential convolutions, to match common encoder backbones, such as a VGG16-based encoder \citep{vgg15}. The encoder and decoder transforms $E_s^{t>1}$ and $D_s^{t>1}$ of a row $s$ have the same number of filters. In practice, we set this number to be half the number of layers of the encoder representation (see details in our experimental setup in Section \ref{sec:experimentalsetup}). In our experiments, we also consider the sparse use of alternative feature transforms for capturing position-dependent representations (\cf{} Figure \ref{fig:nodelevelarchs} for an overview of these transforms).

\paragraph{Skip connections}

We use skip connections by concatenation. Concatenation is favored over element-wise max or sum operators because such operators are special cases of concatenation. Formally, let $K\in\mathbb{N}^\star$ be the depth of two layers to merge, and $e, d, f\in\mathbb{R}^K$ feature vectors respectively for the encoder, the decoder, and the resulting fusion. Let $w, w'\in\mathbb{R}^{K\times K}$ be trainable parameters. Using element-wise max operators: $\forall k\in \{1,...,K\},\allowbreak f_k = \sum_{i=1}^K w_{ik}\max(e_{ik}, d_{ik})$. Using element-wise sum operators: $\forall k\in \{1,...,K\}, f_k = \sum_{i=1}^K w_{ik}(e_{ik} + d_{ik})$. Using concatenation, $\forall k\in \{1,...,K\},\allowbreak f_k = \sum_{i=1}^N (w_{ik} e_{ik} + w'_{ik} d_{ik})$. If needed, an element-wise sum operator can then be modelled by setting $w=w'$. Similarly, an element-wise max operator can be obtained by setting $w_{ik}=0$ or $w'_{ik}=0$ depending on which of the $i$th encoder or decoder channel has greater importance.

\paragraph{Pooling types}

We use max operators in our spatial pooling layers, except in in the encoder ($E^5$) for refinement. In $E^5$, we use instead average pooling to gradually average the pixel embeddings within each instance. As a consequence, if the decoder $D^4$ infers an instance part instead of the whole instance, the representation of this instance will be altered. However, if an entire instance is correctly classified, then its average pixel embedding will remain unchanged. This behavior would not be possible with max pooling because max operators highlight salient pixel embeddings. A wrongly classified instance part could then represent the whole instance.

\subsection{Proposed Training}

A multicameral structure is an acyclic graph, trainable end-to-end. As detecting instance boundaries, detecting occluding boundary sides, and outlining unoccluded instances can be formulated as binary classification tasks, we use balanced cross-entropy loss functions, with instance boundary-aware penalties to synchronize the different supervisions. We are aware of alternative loss functions that address the imbalance between positive and negative examples \citep{DengSLWL18,SEAL18,focalloss17}. As it is not our main focus in this work, we leave the reader to adapt the following loss functions if needed.

\paragraph{Loss functions} 

Formally, let $\mathbf{p}\in\mathcal{P}$ be a pixel location -- typically $\mathcal{P}=\{1, .., W\}\times\{1, .., H\}$ for an image of width $W\in\mathbb{N}^*$ and height $H\in\mathbb{N}^*$. We note $\mathcal{N}=\{1,..,N\}$ where $N\in\mathbb{N}^*$ is the number of training images, and $M_\mathbf{p}\in\mathcal{V}$ the value at location $\mathbf{p}\in \mathcal{P}$ in a matrix $M \in \mathcal{V}^\mathcal{P}$. Let $B^n, O^n, Y^n\in\{0,1\}^P$ be the ground-truth binary images for instance boundaries, occluding boundary sides, and segmentation respectively. Let $\hat{B}^n$, $\hat{O}^n$, $\hat{Y}^n\in[0,1]^P$ be the corresponding network inferences.
\begin{itemize}
\item For instance boundary detection, the decoder $D^2$ minimizes the loss function $\mathcal{L}_b(\theta)$ defined as follows:
\begin{multline}\label{boundary-loss-term}
\mathcal{L}_b(\theta) = - \frac{1}{|\mathcal{N}||\mathcal{P}|}\sum_{n\in\mathcal{N}}\sum_{\mathbf{p}\in\mathcal{P}} \alpha B_{\mathbf{p}}^{n}\log (\hat{B}_\mathbf{p}^n)\\
+\ (1-B_\mathbf{p}^n)\log (1-\hat{B}_\mathbf{p}^n),
\end{multline}
where $\alpha\in\mathbb{R}$ is a penalty to counterbalance the low number of boundary pixels against non-boundary pixels. In our experiments, we set $\alpha=10$.

\item For occluding boundary side detection, the decoder $D^3$ minimizes the loss function $\mathcal{L}_b(\theta)$ defined as follows:
\begin{multline}\label{occlusion-loss-term}
\mathcal{L}_o(\theta) = - \frac{1}{|\mathcal{N}||\mathcal{P}|}\sum_{n\in\mathcal{N}}\sum_{\mathbf{p}\in\mathcal{P}} \alpha O_{\mathbf{p}}^{n}\log (\hat{O}_{\mathbf{p}}^{n})\\
+\ \beta(1-O_{\mathbf{p}}^{n})\log (1-\hat{O}_{\mathbf{p}}^{n}),
\end{multline}
where $\beta=\alpha\ if\ B_\mathbf{p}^n=1\ else\ 1$.

\item For segmentation, the decoders $D^4$ and $D^6$ both minimize the loss function $\mathcal{L}_s(\theta)$ defined as follows:
\begin{multline}\label{seg-loss-term}
\mathcal{L}_s(\theta) = - \frac{1}{|\mathcal{N}||\mathcal{P}|}\sum_{n\in\mathcal{N}}\sum_{\mathbf{p}\in\mathcal{P}} \alpha Y_{\mathbf{p}}^{n}\log (\hat{Y}_{\mathbf{p}}^{n})\ +\ \\
\beta(1-Y_{\mathbf{p}}^{n})\log (1-\hat{Y}_{\mathbf{p}}^n))).
\end{multline}%
\end{itemize}

In the following, if a multicameral structure MC$T$ is trained with these ordinal intermediate supervisions, we write MC$T\dagger$. For example, MC3$\dagger$ is a bicameral structure trained for occlusion-aware boundary detection. RED=MC2=MC2$\dagger$ is a residual encoder-decoder network trained for segmentation.

\paragraph{Ground truth generation}

For each training and test images, we assume that we have the corresponding instance segmentation and the corresponding depth or instance-wise order (in that case, we consider it as a pseudo-depth). The depth (or pseudo-depth) is only used to create the ground truth, but never as input modality.
\begin{itemize}
\item The ground-truth boundaries are trivially derived from the instance segmentation. 
\item For generating the ground-truth occluding boundary sides, we sweep all the ground-truth instance boundaries and at each boundary pixel, we binarize the centered local region by computing the mean Z-offset in each segment of the region (see Figure \ref{fig:mikadoSupp} in appendix). In the end, the ground truth for occlusions is a binary image in which the positive pixels are the instance boundaries slightly translated to one side or another, according to the relative depth difference of the boundary sides. Note that local patches that contain more than two segments are fully set to 0 as they cannot be binarized. This proves to be a reasonable limitation as in practice an overwhelming majority of boundary pixels are between only two instances or between an instance and the background (\eg, 97.1\% of the boundary pixels in Mikado, and 99.4\% in PIOD). We leave for future work the study of the minority of pixels at the junction of more than two instances. 
\item For generating the ground-truth segmentation outlining the unoccluded instances, we compute the number of occluding boundary pixels within each instance. If this ratio is very close to the instance perimeter, then the instance is considered as unoccluded.
\end{itemize}

\section{Proposed Dataset}
\label{sec:proposeddataset}

In this section, we describe the proposed pipeline for generating synthetic homogeneous instance layouts, referred to as Mikado.

\begin{figure}[t]
\centering
\fontsize{7}{8}\selectfont
\setlength{\secondcolumn}{.2cm}
\tikzstyle{data} = [rectangle, draw, fill=white!20,%
    text width=.45\textwidth, text centered, rounded corners, minimum height=.5mm]%
\tikzstyle{conv} = [rectangle, draw, fill=blue!20,%
    text width=2em, text centered, rounded corners=0.5mm, text height=1.5mm, inner sep=0]%
\tikzstyle{sum} = [circle, draw, fill=blue!20, text height=1mm, inner sep=0]%
\tikzstyle{relu} = [rectangle, draw, fill=red!20,minimum width=3em]%
\tikzstyle{pool} = [rectangle, draw, fill=orange!20,text width=2em,text height=1mm, inner sep=0]%
\tikzstyle{unpool} = [rectangle, draw, fill=orange!80,text width=2em,text height=1mm, inner sep=0]%
\setlength{\mywidth}{.11\textwidth}%
\begin{tikzpicture}[node distance = 2mm, auto]%
\node [draw=none,fill=none] (sachet) {\includegraphics[height=\mywidth]{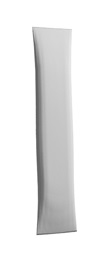}};%
\node [draw=none,fill=none,right=.1\secondcolumn of sachet] (plus) {+};%
\node [draw=none,fill=none,right=1.8\secondcolumn of plus] (texture1) {\includegraphics[height=.9\mywidth]{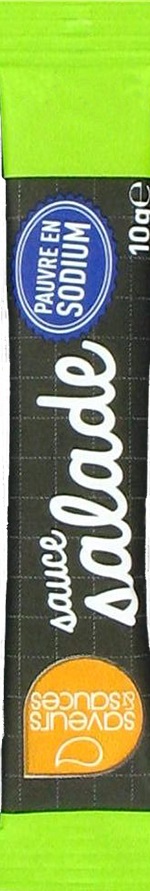}};%
\node [draw=none,fill=none,right=.6\secondcolumn of texture1] (texture2) {\includegraphics[height=.9\mywidth]{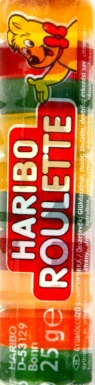}};%
\node [draw=none,fill=none,right=.6\secondcolumn of texture2] (texture3) {\includegraphics[height=.9\mywidth]{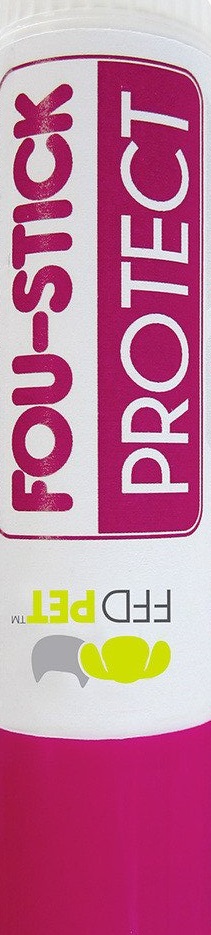}};%
\node [draw=none,fill=none,align=left, left=2.5\secondcolumn of sachet] (sachetmodel) {Sachet model};%
\node [draw=none,fill=none,align=left, above=0.05cm of sachetmodel] (inputs) {\textbf{Inputs:}};%
\node [draw=none,fill=none,align=left, below=0.05cm of sachetmodel] (texturedataset) {Texture images};%
\node [data,below=0.1cm of sachet] (simulation) {\textbf{Physics simulation of piles of sachets}};%
\node [draw=none,fill=none,below=0.01cm of simulation] (simulation2) {\includegraphics[height=\mywidth]{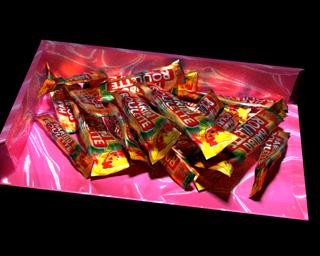}};%
\node [draw=none,fill=none,left=.6\secondcolumn of simulation2] (simulation1) {\includegraphics[height=\mywidth]{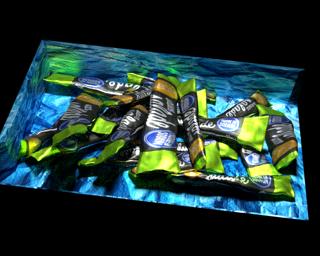}};%
\node [draw=none,fill=none,right=.6\secondcolumn of simulation2] (simulation3) {\includegraphics[height=\mywidth]{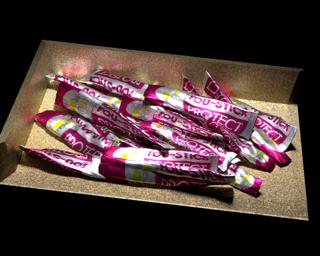}};%
\node [data, below=0.1cm of simulation2] (rendering) {\textbf{Top-view camera (RGB and depth) rendering}};%
\node [draw=none,fill=none,below=0.01cm of rendering] (rendering2) {\includegraphics[height=\mywidth]{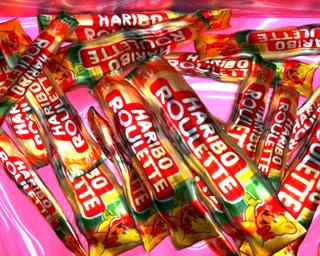}};%
\node [draw=none,fill=none,left=.6\secondcolumn of rendering2] (rendering1) {\includegraphics[height=\mywidth]{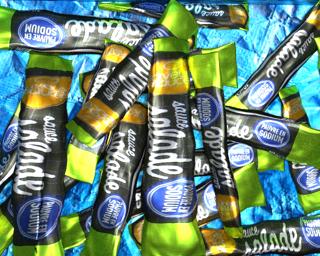}};%
\node [draw=none,fill=none,right=.6\secondcolumn of rendering2] (rendering3) {\includegraphics[height=\mywidth]{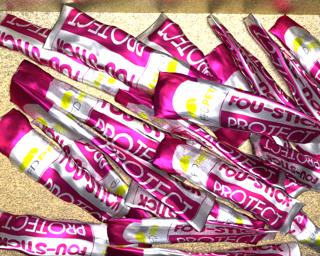}};%
\node [data, below=0.1cm of rendering2] (preparation) {\textbf{Training and test data preparation}};%
\node [draw=none,fill=none,below=0.01cm of preparation] (trainingdata002) {\includegraphics[height=.9\mywidth]{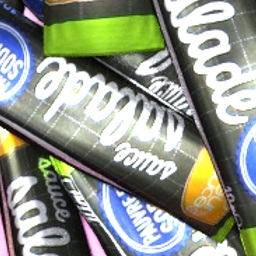}\hspace{0.05cm}\includegraphics[height=.9\mywidth]{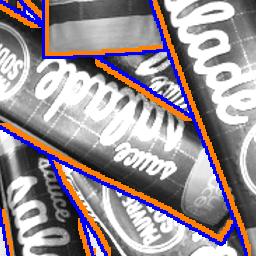}\quad\includegraphics[height=.9\mywidth]{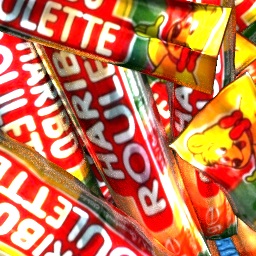}\hspace{0.05cm}\includegraphics[height=.9\mywidth]{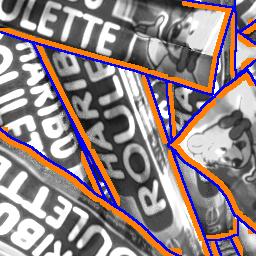}};%
\draw[-{Latex[length=1mm,width=1mm]}] (sachet) -- (simulation);%
\draw[-{Latex[length=1mm,width=1mm]}] (simulation2) -- (rendering);%
\draw[-{Latex[length=1mm,width=1mm]}] (rendering2) -- (preparation);%
\end{tikzpicture}

\caption{Overview of the Mikado pipeline (best viewed in color). Given a mesh template and texture images, piles of deformed instances are generated using a physics engine. A top-view camera is then rendered to capture RGB and depth. The synthetic images and their annotations (ground-truth boundaries are in blue, unoccluded side in orange) are finally prepared to be fed-forward through the network.}
\label{fig:making-mikado}
\end{figure}

\subsection{Data Generation}

In the same vein of \citep{bregier2017iccv,dol18}, we generate synthetic data using custom code on top of Blender \citep{blender}  by simulating scenes of objects piled up in bulk and rendering the corresponding top views, as depicted in Figure \ref{fig:making-mikado}. More precisely, after modelling a static open box and, on top, a perspective camera, a variable number of object instances, in random initial pose, are successively dropped above the box using Blender's physics engine (a video showing the generation of a scene is provided in supplementary material). We then render the camera view, and the corresponding depth image, using Cycles render engine. In this configuration, we ensure a large pose variability and a lot occlusions between instances. The ground-truth unoccluded instances and occluding instance boundary sides can be trivially derived from depth (\cf{} Figure \ref{fig:mikadoSupp}).

However, differently from \citep{bregier2017iccv,dol18}, we consider here piles of many instances with intra-class variations and using only RGB as input modality. We generate RGB images of sachets piled up in bulk by randomly applying global and local deformations to one mesh template of sachet that we texture successively with one out of 120 texture images of sachets retrieved using the Google Images search engine\footnote{https://images.google.com/} and manually cropped to remove any background. Each scene is composed of many instances using the same texture image so as to make the occlusions between instances more challenging to detect. Besides, to prevent the network from simply substracting the background, we apply to the box a texture randomly chosen among 40 background images, retrieved using the Google Images search engine as well. A comprehensive overview of the textures and background images used for generating the Mikado dataset is provided in Figure \ref{fig:mikadoSupp}. Between each image generation, we also randomly jitter the cameras and light locations to prevent the network from learning a fixed source of light, and so fixed reflections and shadows. The proposed dataset finally comprises on average 20.1 instances per image, hence 8 times more instances and 40 times more inter-instance occlusions per image than PIOD. Figure \ref{fig:datasets} provides samples and sums up the Mikado characteristics compared to the state-of-the-art datasets for oriented boundary detection \citep{doc16, FuWTB16} and amodal instance segmentation \citep{Qi2019amodal,FollmannKHKB19,Zhu17amodal}.

\begin{figure}[t]
\centering
\setlength{\mywidth}{.11\textwidth}
\setlength{\mywidthbis}{1pt}
\subfloat[Offline augmentation.]{\label{tab:mikadodatasets}\begin{tabular}{l|l|l}
 & \textbf{Mikado} & \textbf{Mikado+}\\
Mesh templates&
1 \vcenteredinclude{\includegraphics[width=.1\linewidth]{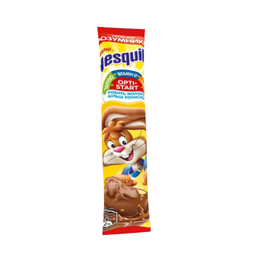}}&
4 \vcenteredinclude{\includegraphics[width=.1\linewidth]{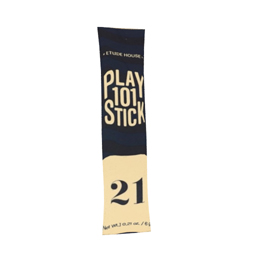}\includegraphics[width=.1\linewidth]{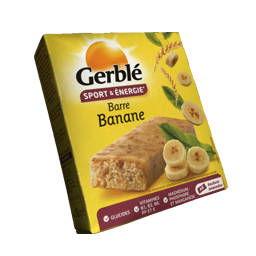}\includegraphics[width=.1\linewidth]{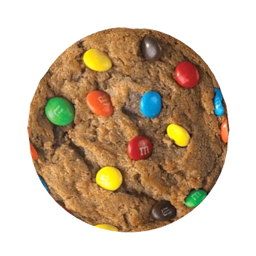}\includegraphics[width=.1\linewidth]{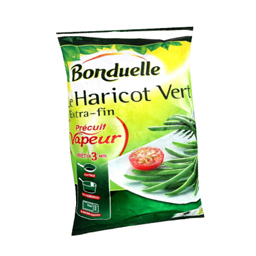}}\\
Backgrounds & 40  & 600\\
Textures & 120 & 2,400 \\
Images & 2,400 & 14,560 \\
\end{tabular}}

\subfloat[Online augmentation.]{\label{fig:onlinemikadoaugmentation}%
\setlength\mywidth{.22\linewidth}%
\setlength\mywidthbis{1em}%
\setlength\secondcolumn{0em}%
\setlength\thirdcolumn{2em}%
\tikzstyle{image} = [draw=none,text width=\mywidth,inner sep=0pt]
\tikzstyle{ann} = [draw=none,text width=5em,text centered,font=\itshape]%
\tikzstyle{link} = [-{Latex[length=2mm,width=1mm]},color=black,line width=.4mm]%
\begin{tikzpicture}
\node [image] (input) {\includegraphics[width=\mywidth]{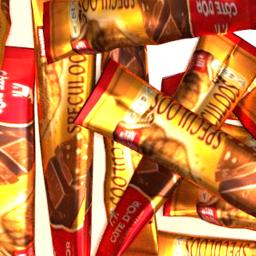}};%
\node [image,right=\mywidthbis of input] (cc1) {\includegraphics[width=\mywidth]{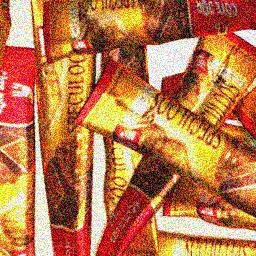}};%
\node [image,right=.25\mywidthbis of cc1] (cc2) {\includegraphics[width=\mywidth]{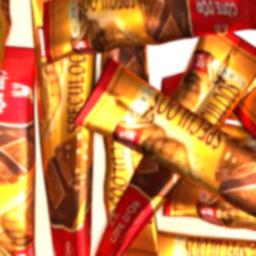}};%
\node [image,right=\mywidthbis of cc2] (cc3) {\includegraphics[width=\mywidth]{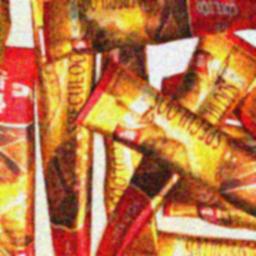}};%
\node [ann,below=\secondcolumn of input] {Raw};%
\node [ann,below=\secondcolumn of cc1] {Jittered};%
\node [ann,below=\secondcolumn of cc2] {Blurred};%
\node [ann,below=\secondcolumn of cc3] {Final};%
\node [image, below=\thirdcolumn of input] (input2) {\includegraphics[width=\mywidth]{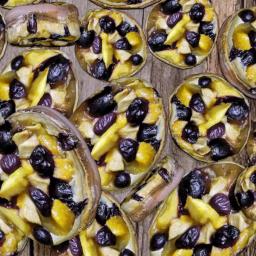}};%
\node [image,right=\mywidthbis of input2] (cc12) {\includegraphics[width=\mywidth]{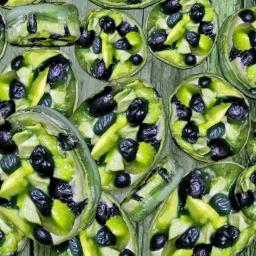}};%
\node [image,right=.25\mywidthbis of cc12] (cc22) {\includegraphics[width=\mywidth]{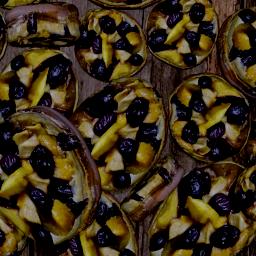}};%
\node [image,right=\mywidthbis of cc22] (cc32) {\includegraphics[width=\mywidth]{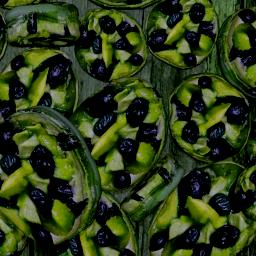}};%
\node [ann,below=\secondcolumn of input2] {Raw};%
\node [ann,below=\secondcolumn of cc12] {Recolored};%
\node [ann,below=\secondcolumn of cc22] {Darkened};%
\node [ann,below=\secondcolumn of cc32] {Final};%
\draw[link]  (input) -- (cc1);%
\draw[link]  (cc2) -- (cc3);%
\draw[link]  (input2) -- (cc12);%
\draw[link]  (cc22) -- (cc32);%
\end{tikzpicture}%
}
\caption{Our synthetic data augmentation for Mikado and its extension Mikado+.}
\label{fig:mikado-augmentation}
\end{figure}

Furthermore, to study the benefits of a richer synthetic data distribution, we make an extension of Mikado, namely Mikado+, following the same proposed generation pipeline but using more mesh templates (sachet, square sachet, box, cylinder-like shape), and more texture and background images. Figure \ref{tab:mikadodatasets} sums up the differences between Mikado and Mikado+.

\subsection{Data Augmentation}

As our RGB images are generated using heuristic rendering models, the training and evaluation may be biased by a lack of realism in the sense that, unlike physical sensors and despite the variations of textures, deformations, and simulated specular reflections, a noise-free pixel information is provided to the network. To remedy this issue, we dynamically filter one image out of two with a gaussian blur and jitter independently the RGB values, as shown in Figure \ref{fig:onlinemikadoaugmentation}, randomly at both training and testing times. The parameters for gaussian filtering and value jittering are randomly chosen within empirically predefined intervals. This prevents the network from overfitting the too perfect synthetic color variations. In addition to dynamic blurring and RGB jittering, the Mikado+ images are also augmented with random permutation of the RGB channels and random under or over-exposition, as also illustrated in Figure \ref{fig:onlinemikadoaugmentation}. Thus, Mikado+ depicts more color and lighting variations than Mikado.

We are aware of optimization-based data augmentation techniques out of the scope of this paper, such as the use of generative models \citep{AntoniouSE18} or automatic search to find the best augmentation policies \citep{CubukZMVL19}. Nevertheless, our augmentation strategy is in line with the work of \citep{CubukZMVL19}, for their search space consists of basic operations, such as rotation and color jittering, just as the ones that we manually apply on our synthetic images.

\begin{figure*}
\centering
\fontsize{8}{9}\selectfont
\setlength\tabcolsep{5pt}
\setlength{\mywidth}{.19\linewidth}%
\setlength{\thirdcolumn}{0.9cm}
\subfloat[Our multicameral structure compared with alternative design patterns for learning contextual representations.]{

}

\bigskip

\fontsize{9}{10}\selectfont
\subfloat[Comparative training and validation errors on Mikado.]{%
\setlength\fwidth{.42\linewidth}
\setlength\fheight{4cm}
%
}%

\bigskip

\subfloat[Comparative performances on Mikado.]{%
\label{fig:segscores}%
\begin{tabular}{l|c|c|c|c}
\multirow{2}*{\textbf{Architecture}} & \textbf{Number of} & \multicolumn{3}{c}{\textbf{Segmentation}} \\
&\textbf{parameters}&ODS&AP&AP$_{60}$\\
\hline
\orangeline RED-Atrous & 1,957,137 & .631 & .619 & .506 \\ 
\greenline RED-Coords & 1,471,105 & .703 & .747 & .599 \\ 
\yellowline RED-Dense/E & 1,202,217 &  .724 &  .774 & .593 \\ 
\blueline \textbf{MC6$\dagger$ (Ours)} & 5,411,916 & \textbf{.767} & \textbf{.825} & \textbf{.691} \\
\multicolumn{5}{c}{}\\
MC2(=RED)$^{4}$ & 1,465,105 & .696 & .732 & .587 \\
MC3$^{4}$ & 2,145,225 & .705 & .750 & .598 \\
MC4$^{4}$ & 2,961,345 & .709 & .762 & .609 \\
MC4$\dagger$$^{4}$ & 2,961,747 & .752 & .802 & .666 \\
\hline
MC2-Coords/D$^{4}$ &  1,490,113 & .713 & .754 & .607 \\
\textbf{MC6$\dagger$-Coords/D4}$^{4}$ & 5,417,916 & \textbf{.766} & \textbf{.824} & \textbf{.696} \\
\hline
MC2-Atrous/D$^{4}$ & 1,367,665 & .591 & .604 & .454 \\
MC4$\star\dagger$-Atrous/D2$^{4}$ & 3,273,834 & .609 & .626 & .476 \\
\textbf{MC6$\dagger$-Atrous/D4}$^{4}$ & 5,053,356 & \textbf{.784} & \textbf{.837} & \textbf{.706} \\
\end{tabular}%
}%
\subfloat[Comparative precision-recall curves on Mikado.]{%
\setlength\fwidth{.35\linewidth}
\setlength\fheight{4.8cm}
%
}%

\begin{flushleft}
\small
$^{4}$ See Figure \ref{fig:mcAblationStudy} for an overview of these architectures. \\
\end{flushleft}

\caption{Comparative results for occlusion-aware instance-sensitive segmentation on Mikado. In these experiments, a pruned VGG16 (or a pruned DenseNet121 for RED-Dense/E) is used as encoder backbone. Best viewed in color.}
\label{fig:segarchs}
\end{figure*}

\begin{figure*}
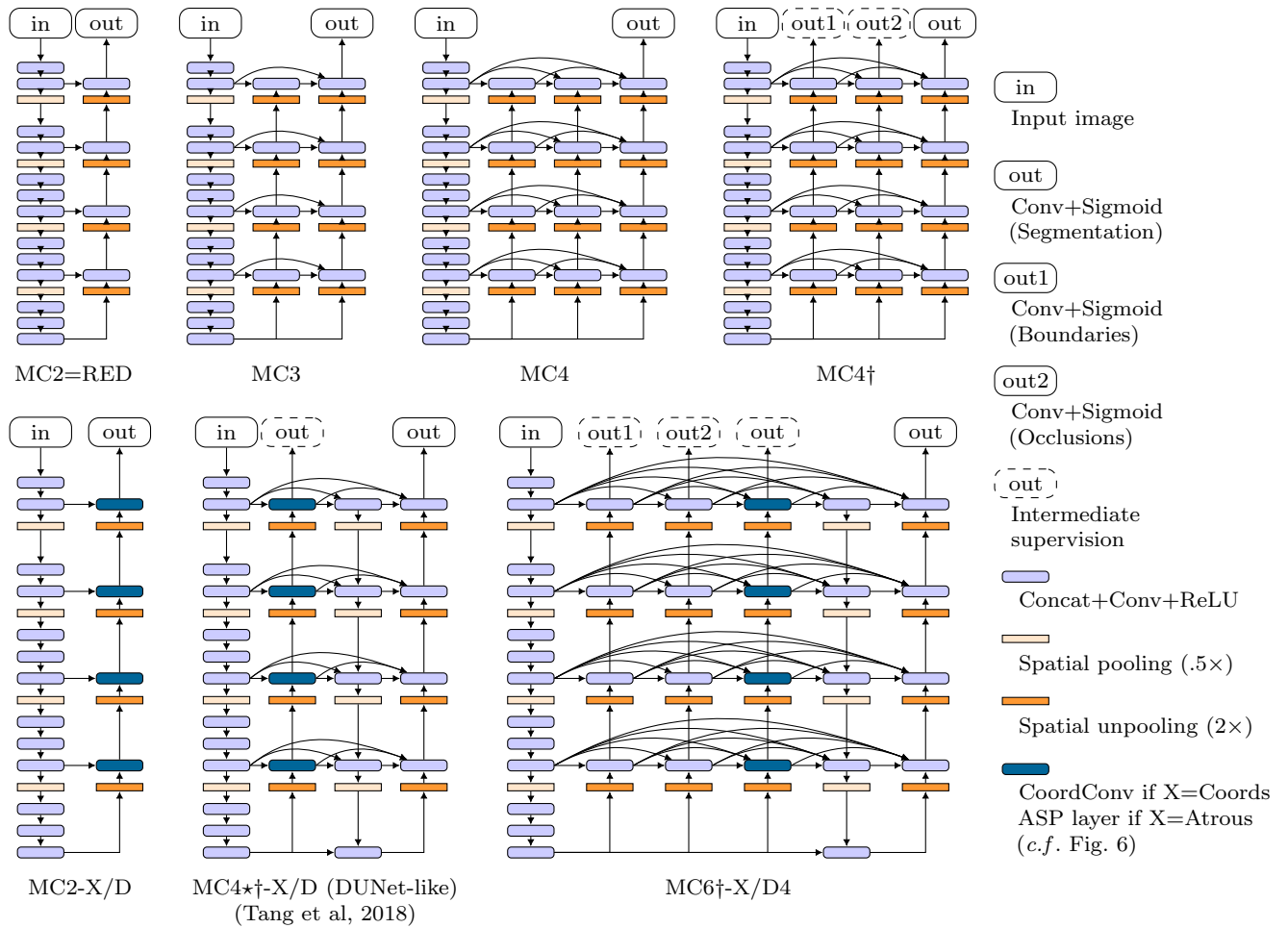

\centering%
\setlength{\mywidth}{.22\linewidth}%
\begin{minipage}{.75\linewidth}
\centering%
\setlength\tabcolsep{8pt}%
\setlength{\thirdcolumn}{0.9cm}%
%
\end{minipage}%
\hspace{.6cm}
\begin{minipage}{.15\linewidth}
\setlength{\secondcolumn}{0.2cm}%
\setlength{\thirdcolumn}{.35\linewidth}%
\tikzstyle{data} = [rectangle, draw, fill=white!20,%
    text width=2em, text centered, rounded corners, minimum height=.5mm]%
\tikzstyle{conv} = [rectangle, draw, fill=blue!20,%
    text width=2em, text centered, rounded corners=0.5mm, text height=1.5mm, inner sep=0]%
\tikzstyle{convtan} = [conv,fill=blue!60!green]%
\tikzstyle{sum} = [circle, draw, fill=blue!20, text height=1mm, inner sep=0]%
\tikzstyle{relu} = [rectangle, draw, fill=red!20,minimum width=3em]%
\tikzstyle{pool} = [rectangle, draw, fill=orange!20,text width=2em,text height=1mm, inner sep=0]%
\tikzstyle{unpool} = [rectangle, draw, fill=orange!80,text width=2em,text height=1mm, inner sep=0]%

%
\end{minipage}
\caption{Multicameral structures with different numbers of encoder and decoder units, and different node types for segmentation inference. Best viewed in color.}
\label{fig:mcAblationStudy}
\end{figure*}

\begin{figure*}
\centering
\setlength\mywidth{.16\linewidth}
\setlength\tabcolsep{1pt}
\begin{tabular}{cccccc}
\vcenteredinclude{\includegraphics[width=\mywidth]{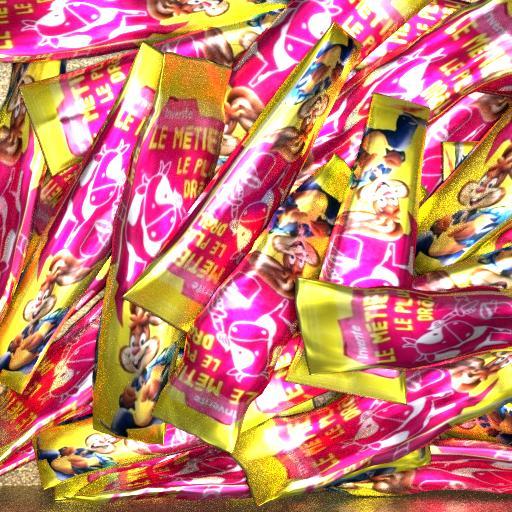}}&
\vcenteredinclude{\includegraphics[width=\mywidth]{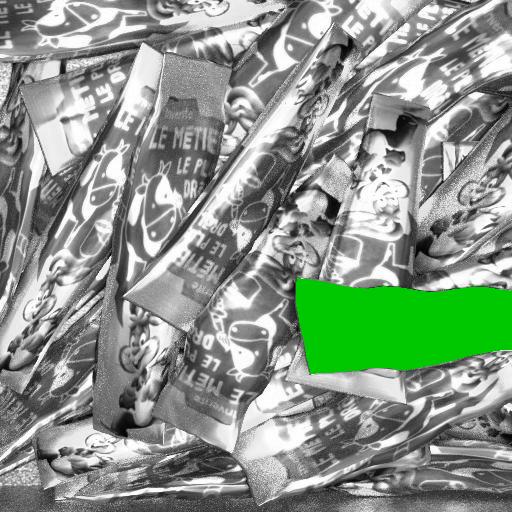}}&
\vcenteredinclude{\includegraphics[width=\mywidth]{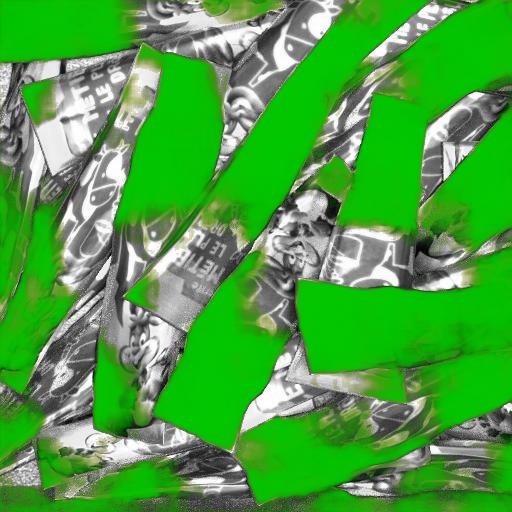}}&
\vcenteredinclude{\includegraphics[width=\mywidth]{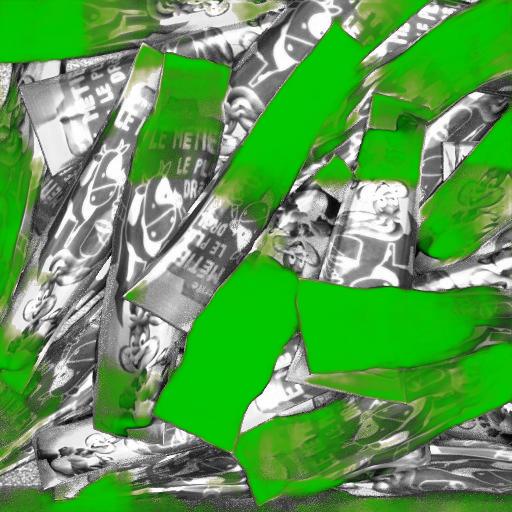}}&
\vcenteredinclude{\includegraphics[width=\mywidth]{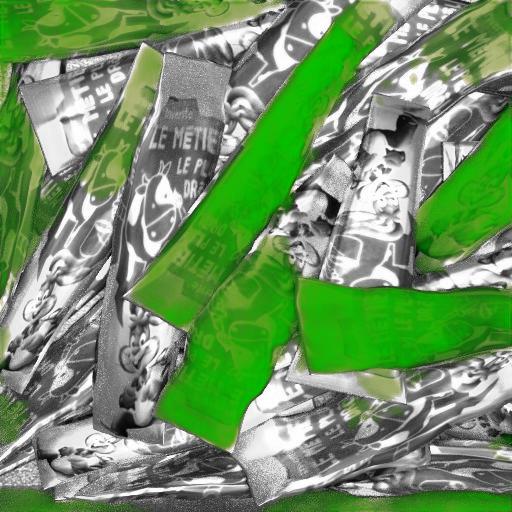}}&
\vcenteredinclude{\includegraphics[width=\mywidth]{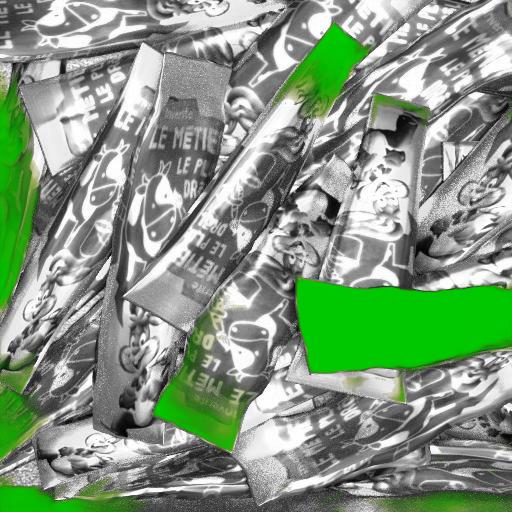}}\\
\vspace{-.3cm}\\
\vcenteredinclude{\includegraphics[width=\mywidth]{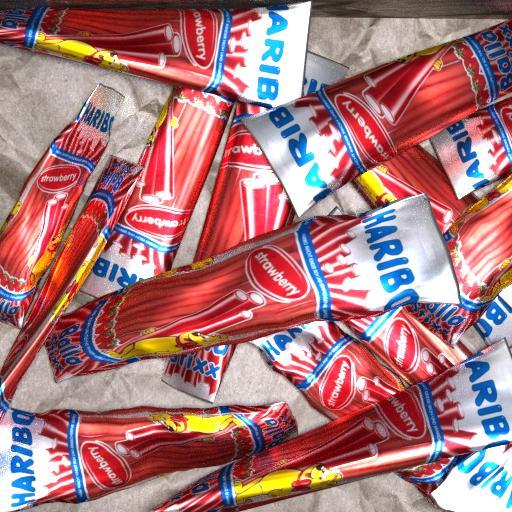}}&
\vcenteredinclude{\includegraphics[width=\mywidth]{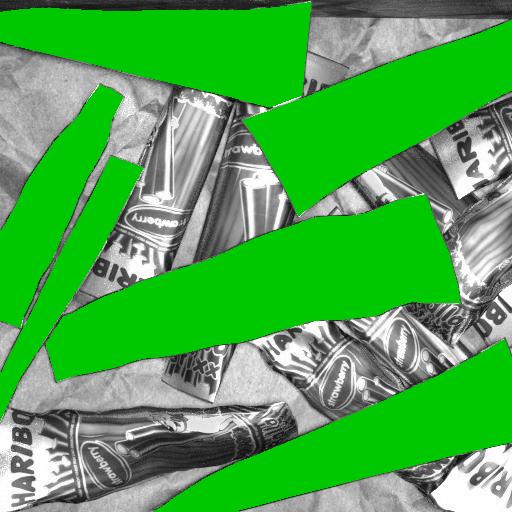}}&
\vcenteredinclude{\includegraphics[width=\mywidth]{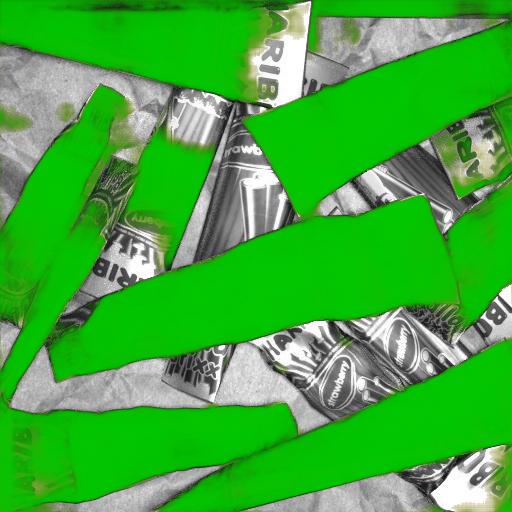}}&
\vcenteredinclude{\includegraphics[width=\mywidth]{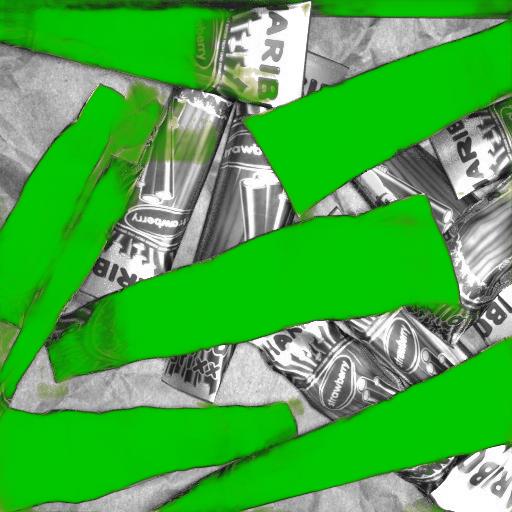}}&
\vcenteredinclude{\includegraphics[width=\mywidth]{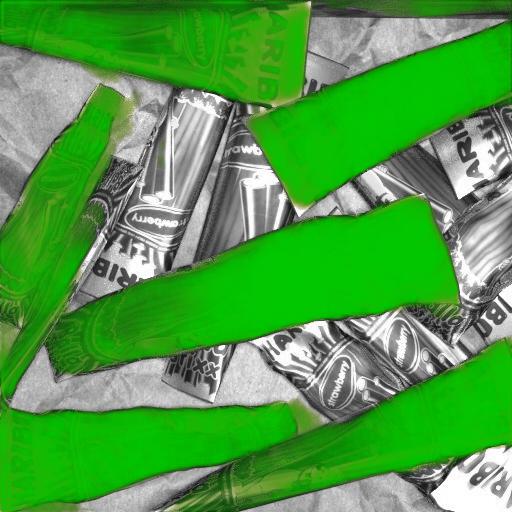}}&
\vcenteredinclude{\includegraphics[width=\mywidth]{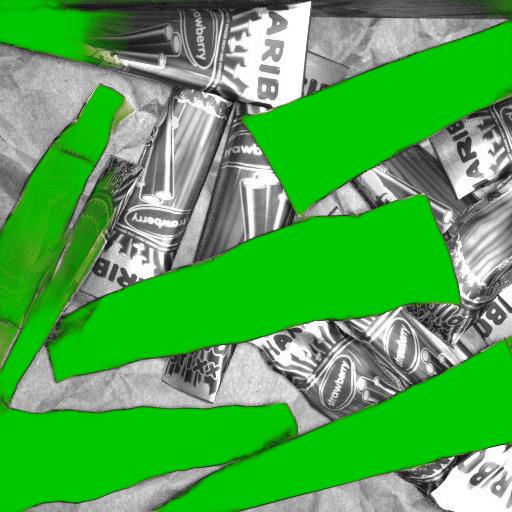}}\\
\vspace{-.3cm}\\
\vcenteredinclude{\includegraphics[width=\mywidth]{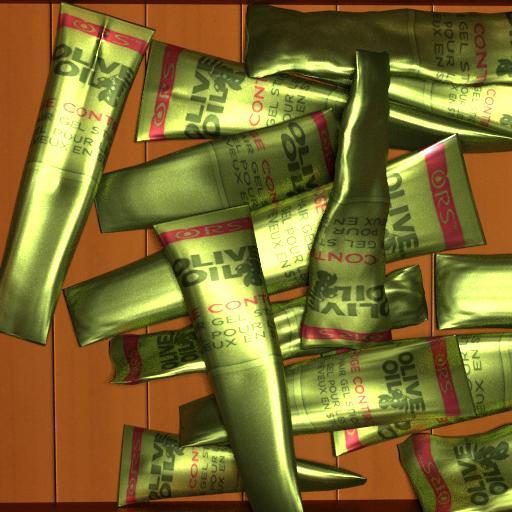}}&
\vcenteredinclude{\includegraphics[width=\mywidth]{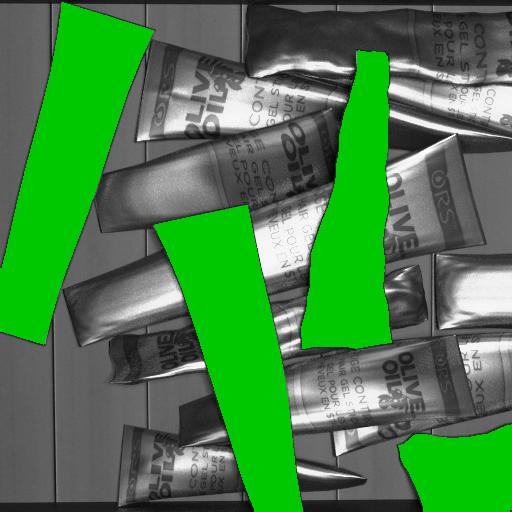}}&
\vcenteredinclude{\includegraphics[width=\mywidth]{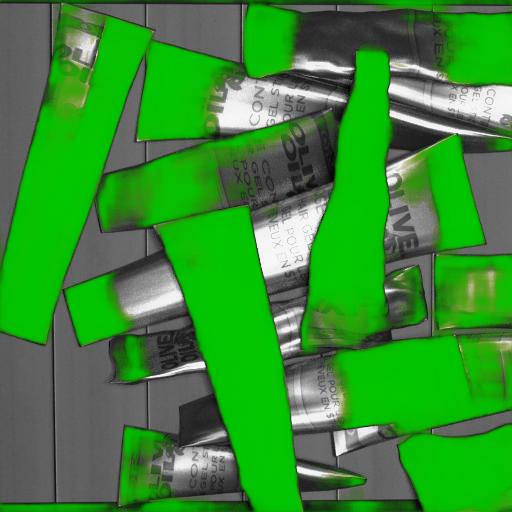}}&
\vcenteredinclude{\includegraphics[width=\mywidth]{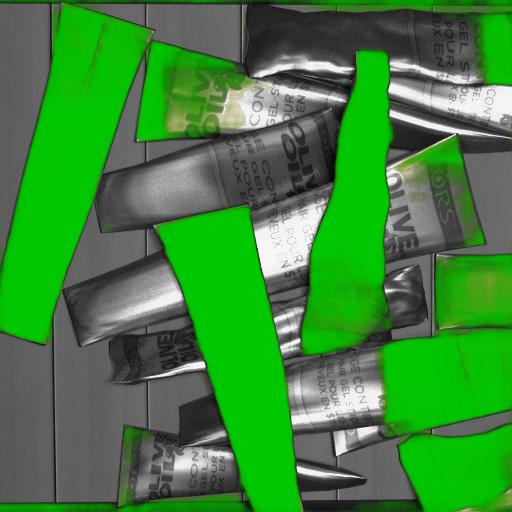}}&
\vcenteredinclude{\includegraphics[width=\mywidth]{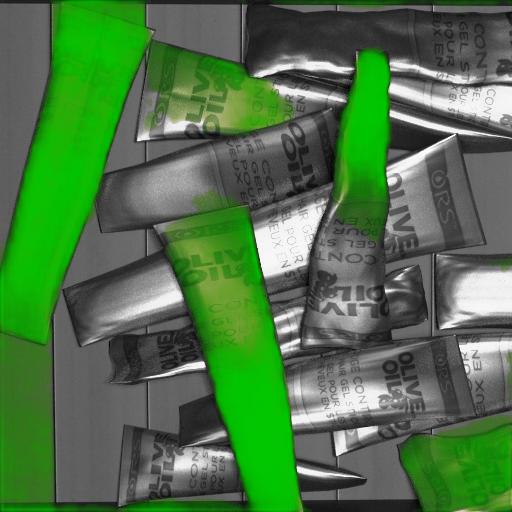}}&
\vcenteredinclude{\includegraphics[width=\mywidth]{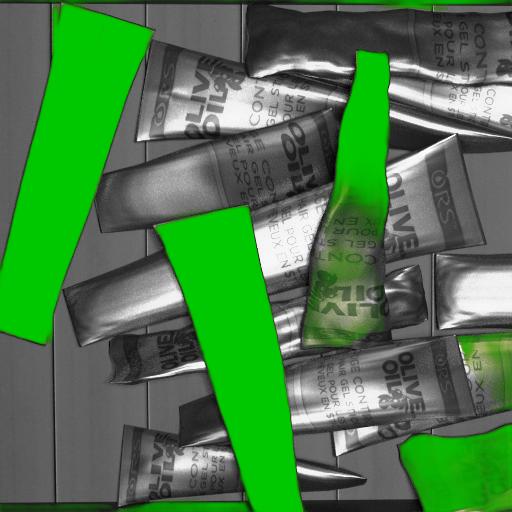}}\\
\vspace{-.3cm}\\
\end{tabular}
\caption{Comparative results on Mikado using different encoder-decoder designs. Best viewed in color.}
\label{fig:results-mask}
\end{figure*}

\begin{figure*}
\centering
\captionsetup[subfigure]{labelformat=empty}
\fontsize{8}{9}\selectfont
\setlength\tabcolsep{6pt}
\setlength{\mywidth}{2.2cm}
\setlength{\thirdcolumn}{0.9cm}
\subfloat{%
%
%
}

\subfloat[Comparative performances on Mikado and PIOD.]{%
\setlength\tabcolsep{5pt}%
\begin{tabular}{l|c|c|c|c|c|c|c|c|c|c|c|c|c}
\multicolumn{2}{c|}{Dataset:}&\multicolumn{6}{c|}{\textbf{Mikado}}& \multicolumn{6}{c}{\textbf{PIOD}}\\
\hline
\multirow{2}*{\textbf{Architecture}} & \textbf{Number of} & \multicolumn{3}{c|}{\textbf{Boundaries}} & \multicolumn{3}{c|}{\textbf{Occlusions}} & \multicolumn{3}{c|}{\textbf{Boundaries}} & \multicolumn{3}{c}{\textbf{Occlusions}} \\
&\textbf{parameters} &ODS&AP&AP$_{60}$&ODS&AP&AP$_{60}$&ODS&AP&AP$_{60}$&ODS&AP&AP$_{60}$\\
\hline
DOOBNet-like & 1,497,330 & .703 & .764 & .583 & .729 & .806 & .633 & .639 & .674 & .446 & .629 & .669 & .451 \\
MTAN-like & 2,075,546 &.700 & .762 & .579 & .727 & .809 & .628 & .646 & .683 & .437 & .632 & .676 & .444 \\
\textbf{MC3$\dagger$ (Ours)} & 2,145,426 & .701 & .762 & .581 & \textbf{.737} & \textbf{.815} & \textbf{.645} & .642 & .673 & .450 & \textbf{.633} & \textbf{.683} & \textbf{.454} \\
\end{tabular}%
}%

\caption{A bicameral structure (MC3$\dagger$) compared with state-of-the-art design patterns adapted for occlusion-aware boundary detection. (a) Encoder and low-resolution half-decoder shared by two independent high-resolution half-decoders.  (b) Task-specific decoders with attention mechanisms to select shared features. (c) Encoder shared by two cascaded decoders. In these experiments, a pruned VGG16 is used as encoder backbone. Best viewed in color.}
\label{fig:obdsarchs}
\end{figure*}

\begin{figure*}
\centering
\captionsetup[subfigure]{labelformat=empty}
\fontsize{7}{8}\selectfont
\setlength{\tabcolsep}{6pt}
\setlength{\mywidth}{.14\textwidth}
\setlength{\secondcolumn}{.2cm}
\setlength{\thirdcolumn}{1cm}


\bigskip

\setlength{\tabcolsep}{8pt}
\fontsize{9}{10}\selectfont
\begin{tabular}{l|c|c|c|c|c|c|c|c|c}
\multicolumn{2}{c|}{Dataset:}&\multicolumn{4}{c|}{\textbf{Mikado}}& \multicolumn{4}{c}{\textbf{PIOD}}\\
\hline
\multirow{2}*{\textbf{Architecture}} & \textbf{Number of} & \multicolumn{2}{c|}{\textbf{Boundaries}} & \multicolumn{2}{c|}{\textbf{Occlusions}} & \multicolumn{2}{c|}{\textbf{Boundaries}} & \multicolumn{2}{c}{\textbf{Occlusions}} \\
&\textbf{parameters}&ODS&AP&ODS&AP&ODS&AP&ODS&AP\\
 \hline
Two streams & 46,839,938 ($\times1.0$) & .755 & .832 & .788 & .872 & .673 & .708 & .681 & .733 \\
Shared encoder & 32,125,250 ($\times.69$) & .769 & .847 & .792 & .876 & .692 & .732 & .686 & .738 \\
Cascaded decoders & 29,949,250 ($\times.64$) & .766 & .844 & .795 & .880 & .694 & .735 & .689 & .748  \\
Multi-task decoder & 23,420,770 ($\times.50$) &.767 & .845 & .795 & .880 & .691 & .731 & .679 & .731 \\
Bicameral (=MC3$\dagger$) & 34,301,250 ($\times.73$) & \textbf{.769} & \textbf{.847} & \textbf{.801} & \textbf{.884} &  \textbf{.697} & \textbf{.738} & \textbf{.692} & \textbf{.747}   \\
\end{tabular}%
\caption{Ablation study on a bicameral structure for occlusion-aware boundary detection. In these experiments, a full VGG16 is used as encoder backbone. The best overall performances are obtained by sharing a single encoder and cascaded decoders, altogether linked via resolution-wise skip connections. Best viewed in color.}
\label{fig:bicameralAblation}
\end{figure*}

\begin{figure*}
\centering

\subfloat[Comparative results for instance boundary (blue) and occluding boundary side (orange) detection on D2SA. From top to bottom: input (i), ground truth (ii), prediction using the proposed network trained on D2SA (iii), using the proposed network pretrained on Mikado then finetuned on D2SA with the first three encoder blocks frozen (iv). Pretraining the proposed network on Mikado before finetuning on D2SA leads to significant improvements.]{%
\label{fig:resultsd2sim}%
\setlength{\mywidth}{.115\textwidth}
\setlength{\mywidthbis}{-0.25cm} 
\setlength{\tabcolsep}{1pt}
\begin{tabular}{rcccccccc}
(i) &
\vcenteredinclude{\includegraphics[width=\mywidth]{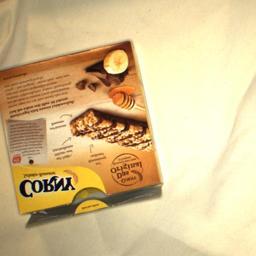}}&
\vcenteredinclude{\includegraphics[width=\mywidth]{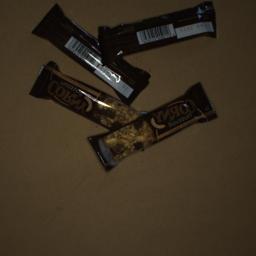}}&
\vcenteredinclude{\includegraphics[width=\mywidth]{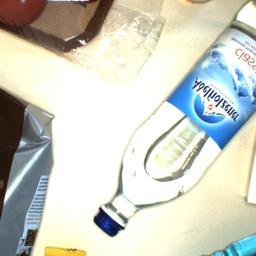}}&
\vcenteredinclude{\includegraphics[width=\mywidth]{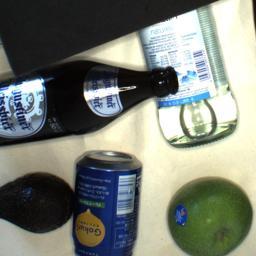}}&
\vcenteredinclude{\includegraphics[width=\mywidth]{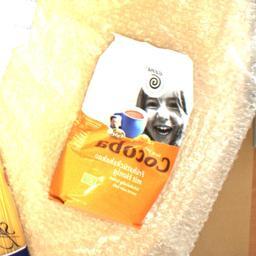}}&
\vcenteredinclude{\includegraphics[width=\mywidth]{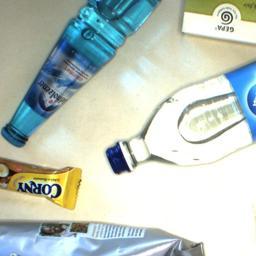}}&
\vcenteredinclude{\includegraphics[width=\mywidth]{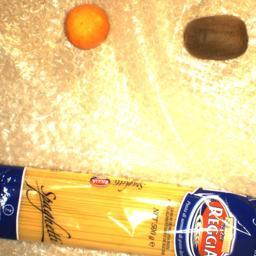}}&
\vcenteredinclude{\includegraphics[width=\mywidth]{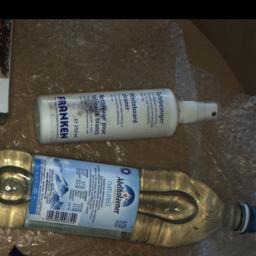}}\\
\vspace{\mywidthbis}\\
(ii) & 
\vcenteredinclude{\includegraphics[width=\mywidth]{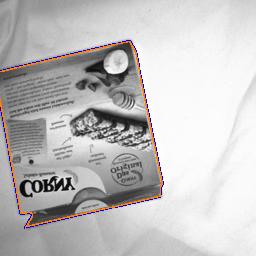}}&
\vcenteredinclude{\includegraphics[width=\mywidth]{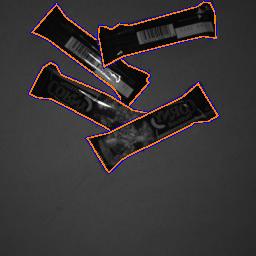}}&
\vcenteredinclude{\includegraphics[width=\mywidth]{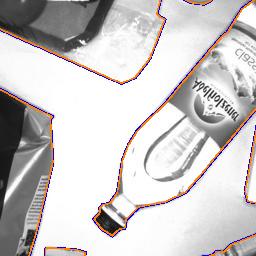}}&
\vcenteredinclude{\includegraphics[width=\mywidth]{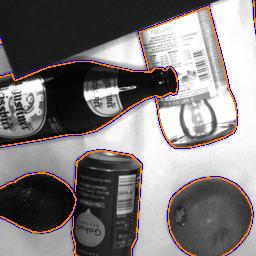}}&
\vcenteredinclude{\includegraphics[width=\mywidth]{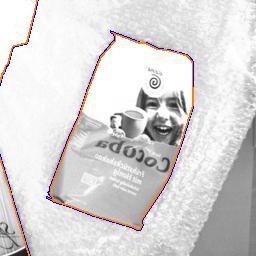}}&
\vcenteredinclude{\includegraphics[width=\mywidth]{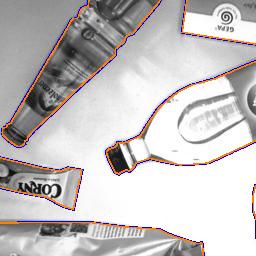}}&
\vcenteredinclude{\includegraphics[width=\mywidth]{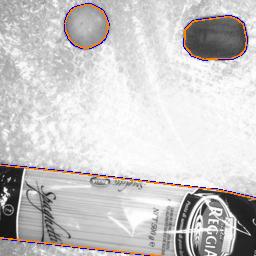}}&
\vcenteredinclude{\includegraphics[width=\mywidth]{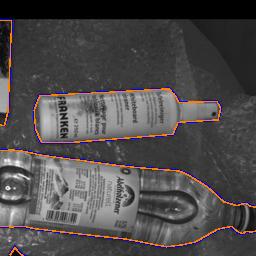}}\\
\vspace{\mywidthbis}\\
(iii) &
\vcenteredinclude{\includegraphics[width=\mywidth]{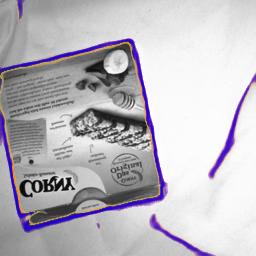}}&
\vcenteredinclude{\includegraphics[width=\mywidth]{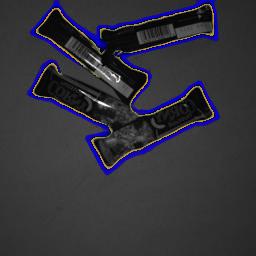}}&
\vcenteredinclude{\includegraphics[width=\mywidth]{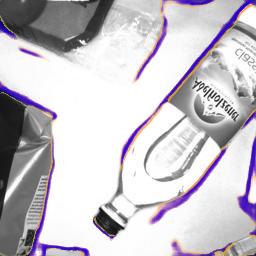}}&
\vcenteredinclude{\includegraphics[width=\mywidth]{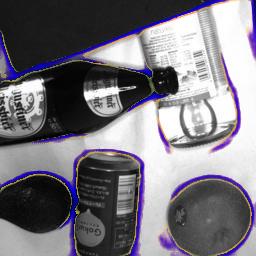}}&
\vcenteredinclude{\includegraphics[width=\mywidth]{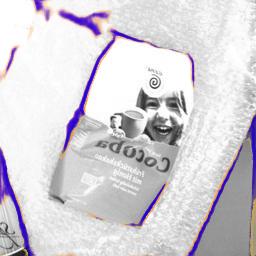}}&
\vcenteredinclude{\includegraphics[width=\mywidth]{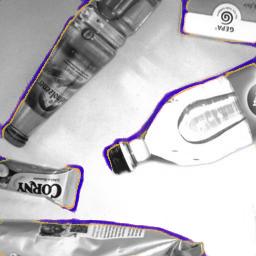}}&
\vcenteredinclude{\includegraphics[width=\mywidth]{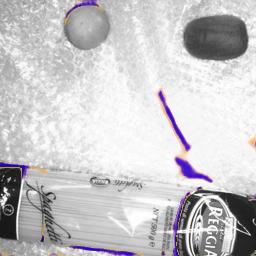}}&
\vcenteredinclude{\includegraphics[width=\mywidth]{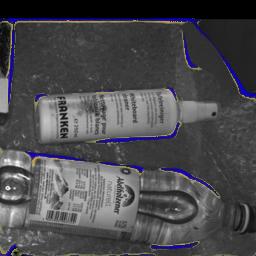}}\\
\vspace{\mywidthbis}\\
\textbf{(iv)} &
\vcenteredinclude{\includegraphics[width=\mywidth]{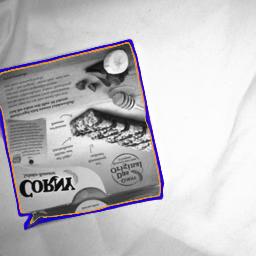}}&
\vcenteredinclude{\includegraphics[width=\mywidth]{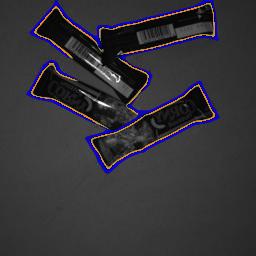}}&
\vcenteredinclude{\includegraphics[width=\mywidth]{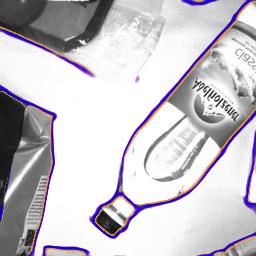}}&
\vcenteredinclude{\includegraphics[width=\mywidth]{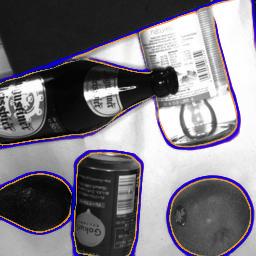}}&
\vcenteredinclude{\includegraphics[width=\mywidth]{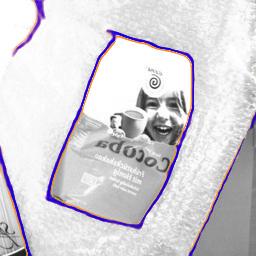}}&
\vcenteredinclude{\includegraphics[width=\mywidth]{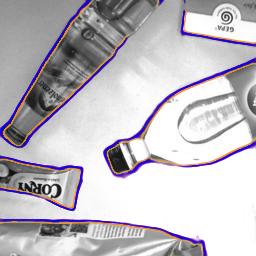}}&
\vcenteredinclude{\includegraphics[width=\mywidth]{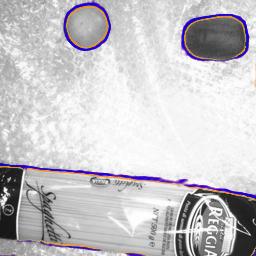}}&
\vcenteredinclude{\includegraphics[width=\mywidth]{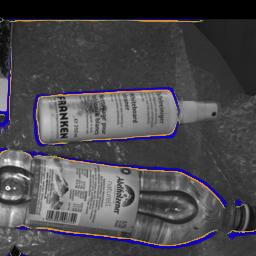}}\\
\vspace{\mywidthbis}\\
\end{tabular}%
}

\subfloat[Performances of a bicameral network pretrained on Mikado/Mikado+ then finetuned on D2SA with the encoder blocks 1, 2, 3 frozen (see also Figure \ref{fig:arch-bicameral-frozen} in appendix). The performances are shown \wrt{} the percentage of real images retained for finetuning. Exploring a wider range of configurations in simulation (Mikado+) enables to learn more abstract local representations of the boundaries and occlusions, thus achieving state-of-the-art performances while drastically reducing the number of real images for finetuning.]{%
\label{fig:resultsd2shist}
\fontsize{8}{7}\selectfont
\setlength\fwidth{.5\linewidth}
\setlength\fheight{5cm}
\setlength\mywidth{.035\fwidth}
\setlength{\tabcolsep}{1pt}
\begin{tabular}{cc}
\vcenteredinclude{
\definecolor{mycolor1}{rgb}{0.85000,0.32500,0.09800}
\definecolor{mycolor3}{rgb}{0.00000,0.44700,0.74100}
\definecolor{mycolor7}{rgb}{0.30100,0.74500,0.93300}%
\definecolor{mycolor8}{rgb}{0.32500,0.85000,0.8}
\definecolor{mycolor9}{rgb}{0.98800,0.55,0.349}%
\begin{tikzpicture}
  \centering
  \begin{axis}[
        ybar=.02cm, axis on top,
        height=\fheight, width=\fwidth,
        bar width=\mywidth,
        minor ytick={.700,.783},        
        grid=minor,
        yticklabels={,},  
	x tick label style={xshift=-(\tick<100)*(\mywidth+.01cm)},        
        minor grid style={draw=mycolor1,line width=1},
        enlarge y limits={value=.1,upper},
        ymin=0.56, ymax=.9,
        axis x line*=bottom,
        y axis line style={opacity=0},
        tickwidth=0pt,
        enlarge x limits=.15, 
        legend style={
            at={(0.5,-0.2)},
            anchor=north,
            legend columns=-1,
            /tikz/every even column/.append style={column sep=.5\mywidth}
        },
        xlabel={Percentage of the initial D2SA finetuning set},
        ylabel={Boundary ODS},
        symbolic x coords={
           0,13,25,38,50,75,100},
       xtick=data,
       nodes near coords={
        \rotatebox{70}{\pgfmathprintnumber[skip 0.,precision=3,zerofill]{\pgfplotspointmeta}}
       }
    ]
    \addplot [draw=none, fill=mycolor3] coordinates {
      (0,.652)
      (13,.764)
      (25, .788) 
      (38,.795)
      (50,.797) 
      (75,.794) 
      (100,.793) };
   \addplot [draw=none,fill=mycolor7] coordinates {
      (0,.581)   
      (13,.790)
      (25, .804) 
      (38,.806)
      (50,.804) 
      (75, .809) 
      (100, .809) };    
   \addplot [draw=none,fill=mycolor1] coordinates { 
      (100, .700) };          
   \addplot [draw=none,fill=mycolor9] coordinates { 
      (100, .783) };        
      
  \end{axis}
  \end{tikzpicture}
} & \vcenteredinclude{
\definecolor{mycolor1}{rgb}{0.85000,0.32500,0.09800}
\definecolor{mycolor3}{rgb}{0.00000,0.44700,0.74100}
\definecolor{mycolor7}{rgb}{0.30100,0.74500,0.93300}%
\definecolor{mycolor8}{rgb}{0.32500,0.85000,0.8}
\definecolor{mycolor9}{rgb}{0.98800,0.55,0.349}%
\begin{tikzpicture}
  \centering
  \begin{axis}[
        ybar=.02cm, axis on top,
        height=\fheight, width=\fwidth,
        bar width=\mywidth,
        minor ytick={.715,.792},        
        grid=minor,
        yticklabels={,},  
	x tick label style={xshift=-(\tick<100)*(\mywidth+.01cm)},                
        minor grid style={draw=mycolor1,line width=1},
        enlarge y limits={value=.1,upper},
        ymin=0.56, ymax=.9,
        axis x line*=bottom,
        y axis line style={opacity=0},
        tickwidth=0pt,
        enlarge x limits=0.15,
        legend style={
            at={(0.5,-0.2)},
            anchor=north,
            legend columns=-1,
            /tikz/every even column/.append style={column sep=.5\mywidth}
        },
        xlabel={Percentage of the initial D2SA finetuning set},
        ylabel={Boundary AP},
        symbolic x coords={
           0,13,25,38,50,75,100},
       xtick=data,
       nodes near coords={
        \rotatebox{70}{\pgfmathprintnumber[skip 0.,precision=3,zerofill]{\pgfplotspointmeta}}
       }
    ]
    \addplot [draw=none, fill=mycolor3] coordinates {
      (0,.649)
      (13,.782)
      (25, .809) 
      (38,.815)
      (50,.812) 
      (75,.819) 
      (100,.819) };
   \addplot [draw=none,fill=mycolor7] coordinates {
      (0,.580)   
      (13,.816)
      (25, .828) 
      (38,.830)
      (50,.829) 
      (75, .835) 
      (100, .836) };   
   \addplot [draw=none,fill=mycolor1] coordinates { 
      (100, .715) };          
   \addplot [draw=none,fill=mycolor9] coordinates { 
      (100, .792) };        

  \end{axis}
  \end{tikzpicture}
} \\
\vcenteredinclude{
\definecolor{mycolor1}{rgb}{0.85000,0.32500,0.09800}
\definecolor{mycolor3}{rgb}{0.00000,0.44700,0.74100}
\definecolor{mycolor7}{rgb}{0.30100,0.74500,0.93300}%
\definecolor{mycolor8}{rgb}{0.32500,0.85000,0.8}
\definecolor{mycolor9}{rgb}{0.98800,0.55,0.349}%
\begin{tikzpicture}
  \centering
  \begin{axis}[
        ybar=.02cm, axis on top,
        height=\fheight, width=\fwidth,
        bar width=\mywidth,
        minor ytick={.725,.785},        
        grid=minor,
        yticklabels={,},        
	x tick label style={xshift=-(\tick<100)*(\mywidth+.01cm)},                
        minor grid style={draw=mycolor1,line width=1},
        enlarge y limits={value=.1,upper},
        ymin=0.35, ymax=.94,
        axis x line*=bottom,
        y axis line style={opacity=0},
        tickwidth=0pt,
        enlarge x limits=.15,
        legend style={
            at={(0.5,-0.2)},
            anchor=north,
            legend columns=-1,
            /tikz/every even column/.append style={column sep=.5\mywidth}
        },
        xlabel={Percentage of the initial D2SA finetuning set},
        ylabel={Occlusion ODS},
        symbolic x coords={
           0,13,25,38,50,75,100},
       xtick=data,
       nodes near coords={
        \rotatebox{70}{\pgfmathprintnumber[skip 0.,precision=3,zerofill]{\pgfplotspointmeta}}
       }
    ]
    \addplot [draw=none, fill=mycolor3] coordinates {
      (0,.458)
      (13,.768)
      (25, .796) 
      (38,.804)
      (50,.808) 
      (75,.808) 
      (100,.810) };
   \addplot [draw=none,fill=mycolor7] coordinates {
      (0,.357)   
      (13,.801)
      (25, .816) 
      (38,.819)
      (50,.817) 
      (75, .821) 
      (100, .823) };    
   \addplot [draw=none,fill=mycolor1] coordinates { 
      (100, .725) };          
   \addplot [draw=none,fill=mycolor9] coordinates { 
      (100, .785) };             

  \end{axis}
  \end{tikzpicture}
} & \vcenteredinclude{
\definecolor{mycolor1}{rgb}{0.85000,0.32500,0.09800}
\definecolor{mycolor3}{rgb}{0.00000,0.44700,0.74100}
\definecolor{mycolor7}{rgb}{0.30100,0.74500,0.93300}%
\definecolor{mycolor8}{rgb}{0.32500,0.85000,0.8}
\definecolor{mycolor9}{rgb}{0.98800,0.55,0.349}%
\begin{tikzpicture}
  \centering
  \begin{axis}[
        ybar=.02cm, axis on top,
        height=\fheight, width=\fwidth,
        bar width=\mywidth,
        minor ytick={.756,.795},        
        grid=minor,
        yticklabels={,},  
	x tick label style={xshift=-(\tick<100)*(\mywidth+.01cm)},         
        minor grid style={draw=mycolor1,line width=1},
        enlarge y limits={value=.1,upper},
        ymin=0.35, ymax=.94,
        axis x line*=bottom,
        y axis line style={opacity=0},
        tickwidth=0pt,
        enlarge x limits=.15,
        legend style={
            at={(0.5,-0.2)},
            anchor=north,
            legend columns=-1,
            /tikz/every even column/.append style={column sep=.5\mywidth}
        },
        xlabel={Percentage of the initial D2SA finetuning set},
        ylabel={Occlusion AP},
        symbolic x coords={
           0,13,25,38,50,75,100},
       xtick=data,
       nodes near coords={
        \rotatebox{70}{\pgfmathprintnumber[skip 0.,precision=3,zerofill]{\pgfplotspointmeta}}
       }
    ]
    \addplot [draw=none, fill=mycolor3] coordinates {
      (0,.400)
      (13,.793)
      (25, .832) 
      (38,.842)
      (50,.850) 
      (75,.849) 
      (100,.849) };
   \addplot [draw=none,fill=mycolor7] coordinates {
      (0,.365)   
      (13,.834)
      (25, .849) 
      (38,.851)
      (50,.850) 
      (75, .857) 
      (100, .859) };    
   \addplot [draw=none,fill=mycolor1] coordinates { 
      (100, .756) };          
   \addplot [draw=none,fill=mycolor9] coordinates { 
      (100, .795) };        
      
  \end{axis}
  \end{tikzpicture}
} \\
\\
\multicolumn{2}{c}{\orangeline/\lightorangeline Training only on D2SA/D2SA+ \hspace{\mywidth} \blueline Pretraining on Mikado \hspace{\mywidth} \lightblueline Pretraining on Mikado+}\\
\end{tabular}%
}

\caption{Comparative results on D2SA using a bicameral structure trained for occlusion-aware boundary detection, under different pretraining conditions. Best viewed in color.}

\label{fig:resultsd2s}
\end{figure*}

\section{Experimental Setup}
\label{sec:experimentalsetup}

In this section, we describe our experiments to evaluate the proposed model and check the plausibility of the jointly proposed synthetic data. Specifically, the proposed model is evaluated on two differents aspects: (i) learning to map an image or a region that contains multiple overlapping similar instances to an instance-sensitive segmentation; (ii) learning to detect occlusion-aware instance boundaries. Our experiments are divided into three parts: 

\begin{enumerate}
\item We compare variants of multicameral structures with alternative encoder-decoder designs, trained for oc\-clusion-aware instance-sensitive segmentation.
\item We compare the bicameral part of our model with alternative layer and connection structurings, trained for occlusion-aware boundary detection. 
\item We evaluate the plausibility of the proposed synthetic data on a real-world setup.
\end{enumerate}

\subsection{Evaluation Metrics}

We use the same metrics to evaluate oc\-clusion-aware segmentations and boundaries, as they all result from pixel-wise binary classification tasks. Specifically, we compute the precision and recall for different binarization thresholds, then typical derived metrics: the best F-score on dataset scale (ODS), the average precision (AP), and the average precision in high-recall regime (AP$_{60}$). 
\begin{itemize}
\item ODS is the best harmonic mean of precision and recall over the full recall interval. 
\item AP conveys the area under the precision-recall curve over the full recall interval.
\item AP$_{60}$ is the average precision on the recall interval $[.6,1]$, thus without taking into account high precisions due to empty inferences.
\end{itemize}
As matching tolerance, \ie the maximum $\ell_2$-distance to the closest ground-truth pixel for a positive or negative to be considered as true or false respectively, we set a hard value of 0 pixels for Mikado (which contains perfect ground-truth annotations) and a state-of-the-art value of $\tau = 0.0075 \sqrt{W^2+H^2} (\simeq 2.7$ pixels for 256$\times$256 images) for PIOD and D2SA that contain approximative hand-made annotations, where $W\in\mathbb{N}^\star$ and $H\in\mathbb{N}^\star$ are the image width and height respectively. Evaluation is performed without non-maximum suppression, which may artificially improve precision.

\subsection{Instance-Sensitive Segmentation}
\label{sec:segexpsetup}

In our first set of experiments, we evaluate and analyze the proposed design for instance-sensitive segmentation on Mikado.

\paragraph{Baselines}

We first compare our design with state-of-the-art variants of residual encoder-decoder (RED) networks for reducing the translation invariance of the latent representations (see Figures \ref{fig:nodelevelarchs} and \ref{fig:segarchs}).
\begin{itemize}
\item \textbf{Atrous spatial pyramid (Atrous)} Aggregating convolutions with different dilation rates on top of the encoder enables to capture longer-range pixel relations \citep{deeplabv3+18,WangCYLHHC18,YuK16}. Such relations are key cues to understand the notions of instance and occlusion. We compare with a RED network equipped with aggregated dilated convolutions on top of the encoder (RED-Atrous), similarly to \citep{deeplabv3+18}.

\item \textbf{Coordinate-aware convolutions (Coords)} Concatenating feature maps and hard-coded pixel coordinates, namely CoordConv, improves the learning of pixel classification tasks that require some translation variance \citep{Liu18coordconv}. We compare the proposed model with a RED network in which all the convolution layers are swapped to CoordConv ones (RED-Coords).

\item \textbf{Dense encoder blocks (Dense/E)} Deepening the encoder blocks using densely connected layers has proved efficient for capturing more discriminative representations \citep{densenet17}. Deeper hierarchical representations enable to encode more complex and longer-range pixel relations, as the receptive fields implicitly grow layer after layer. We include a RED network equipped with a DenseNet121-based encoder (RED-Dense/E) in our comparison.
\end{itemize}

\paragraph{Ablation study}

To further our evalution, we analyze three important aspects: the number of units in a multicameral sequence, the presence of intermediate supervisions, and the optional use of specific nodes in the decoding process. The resulting designs are illustrated in Figure \ref{fig:mcAblationStudy}.
\begin{itemize}
\item \textbf{Number of cascaded units} Adding decoder and encoder-decoder units in a multicameral sequence implies more parameters to train and more memory at inference time. We thus quantify the impact of many decoder units (MC2 vs.{} MC3 vs.{} MC4), and the presence of a refinement encoder-decoder unit (MC2 vs.{} MC4$\star\dagger$; MC4$\dagger$ vs.{} MC6$\dagger$). Note that MC4$\star\dagger$ is a periodic multicameral sequence of encoder-decoder units. This special case has been studied in \citep{TangPGWZM18}, as DUNet, for refining visual landmark detection. Comparing MC4$\star\dagger$ with MC6$\dagger$ therefore also shows the benefits of a more general coupling of units with ordinal intermediate supervisions.

\item \textbf{Intermediate supervision} Generally, intermediate supervisions improve the training of complex graphs. In this work, we show the impact of ordinal intermediate supervisions to enforce a learning path: (1) detect image cues; (2) infer instance boundaries; (3) infer occluding boundary sides; (4) infer unoccluded instances. In our experiments, the first three decoders are supervised to infer the instance boundaries, the occluding boundary sides and the unoccluded instances respectively, using the loss functions presented in Section \ref{sec:proposedmodel} (MC4$\dagger$ and MC6$\dagger$). Comparing MC4 with MC4$\dagger$ thus shows the impact of such supervisions.

\item \textbf{Optional specific nodes} Dilated and coordinate-aware convolutions locally reduce the translation invariance of convolutional embeddings. We try to combine these design patterns within our multicameral sequence. Specifically, we compare variants of MC2 and MC6$\dagger$ networks in which we use such nodes in the first decoder for outlining the unoccluded instances (D and D4 respectively). These variants are thus referred to as MC2-X/D and MC6$\dagger$-X/D4 respectively, with X$\ \in\{$Coords,Atrous$\}$. 
\end{itemize}

\paragraph{Implementation details}

Due to hardware limitations, we compare the networks using a pruned VGG16 (or a pruned DenseNet121 for the RED-Dense/E design) as first encoder backbone. Specifically, we keep the first quarter of filters at each layer in the original encoder. For the remaining layers, we set a kernel size of $5\times 5$ and the numbers of filters reported in Table \ref{tab:numfilters}.

\begin{table}[h]
\centering
\begin{tabular}{c|lll|ll}
& \multicolumn{3}{c|}{$E_s^1$} & \multicolumn{2}{c}{$\{E,D\}_s^{t>1}$}\\
Resolution & VGG16 & full & pruned& full & pruned\\
\hline
$s=1$ & conv1\_x & 64 & 16 & 32 & 8 \\ 
$s=2$ & conv2\_x & 128 & 32 & 64 & 16 \\  
$s=3$ & conv3\_x & 256 & 64 & 128 & 32 \\ 
$s=4$ & conv4\_x & 512 & 128 & 256 & 64 \\ 
$s=5$ & conv5\_x & 512 & 128 & 256 & 64 \\  
\end{tabular}%
\caption{Number of filters for each layer in our full or pruned network implementations, using a full or pruned VGG16 as first encoder backbone ($E_s^1$).}
\label{tab:numfilters}
\end{table}

\subsection{Occlusion-Aware Boundaries}

Our most performance-enhancing multicameral design (MC6$\dagger$) includes a bicameral structure (MC3$\dagger$) trained for occlusion-aware boundary detection. To further our analysis on the multicameral components, we evaluate this structure alone on Mikado and PIOD.

\paragraph{Baselines} We compare MC3$\dagger$ with related layer and connection structurings, released concurrently to our work (see Figure \ref{fig:obdsarchs}).

\begin{itemize}
\item \textbf{DOOBNet} \citep{Wang18doobnet} proposed an incremental improvement of \citep{doc16} for occlusion-aware boundary detection. \citep{doc16} employed two independent VGG16-based encoder-decoder networks for boundaries and occlusion orientations respectively. Instead, \citep{Wang18doobnet} used a single encoder and a single low-resolution half-decoder, both shared by two independent high-resolution decoders. They also proposed incremental improvements: a ResNet-based encoder, an ASP layer on top of it like in \citep{deeplabv3+18}, and a focal loss-like function to drive the training \citep{focalloss17}. We compare a bicameral structure with the core DOOBNet design, \ie{} without these incremental improvements.

\item \textbf{MTAN} In a more general context, \citep{Liu19mtan} have introduced attention masks at each resolution for pixel-wise multi-task learning. Such masks enable resolution-wise task-specific selections of shared features. As learning jointly boundaries and occlusions also requires shared and task-specific representations, we compare bicameral decoders with MTAN-like decoders for boundaries and occlusions respectively.
\end{itemize}

\paragraph{Ablation study} 

To further our above comparison, we isolate the impacts of sharing a single encoder and cascading decoders, and we study how bicameral decoders compare with partially shared decoders (\cf{} Figure \ref{fig:bicameralAblation}). In appendix, we also study the impact of bicameral skip connections (see Figures \ref{fig:resultsskip} and \ref{fig:resultsskiploss}).
\begin{itemize}
\item \textbf{Bicameral components} We compare a bicameral structure with three intermediate designs: two independent encoder-decoder streams (DOC-like \citep{doc16}); two independent decoders sharing a single encoder; two cascaded decoders sharing a single encoder.

\item \textbf{Partial decoder sharing} We compare a bicameral structure with four alternative levels of decoder sharing: bicameral decoders sharing their lowest-resolution layer; sharing their two lowest-resolution layers; their three lowest-resolution ones; all their layers, which is equivalent to multi-task decoding. 
\end{itemize}

\paragraph{Implementation details}

We use a pruned VGG16 as encoder backbone for our comparison with DOOBNet-like and MTAN-like architectures. In our ablation study, a full VGG16 is used as encoder backbone. Our pruning scheme and layer hyperparameters are the same as the ones in Section \ref{sec:segexpsetup}.

\subsection{Data Plausibility Check}

As Mikado is a computer-generated dataset, one may raise the question whether it is realistic. The answer is obviously no, but we claim that it is valuable for significative evaluations. To prove this point, we evaluate the transferability of features learned from Mikado to real data. In line with \citep{YosinskiCBL14}, features learned from a source domain are transferable if they can be repurposed and boost generalization on a target domain. As target domain, we use D2SA \citep{D2S18} (see samples in Figure \ref{fig:datasets}).

\paragraph{Synthetic feature transferability} 

As deep features transition from general to specific by the last layers, we train a bicameral network for occlusion-aware boundary detection on Mikado, then freeze some of the encoder blocks and retrain the remaining layers on D2SA. We conduct different finetunings, by reducing progressively the number of D2SA images used for finetuning.

\paragraph{Synthetic data distribution}

To highlight the benefits of synthetic data in contrast with hardly extensible real-world datasets, we additionally study how a richer synthetic data distribution, \ie{} Mikado+, impacts the domain adaptation. As the ranges of texture, shape, and pose variations are more widely represented in Mikado+, better transferable invariants are expected to be learned. In a limited manner, D2SA addresses this case by overlaying manually isolated instances into fake training images \citep{D2S18}. We thus compare with this augmentation strategy, referred to as D2SA+.

\paragraph{Implementation details}

To expose the most transferable features learned from Mikado, we first compare bicameral networks finetuned on D2SA with different encoder block at which the network is chopped and retrained (\cf{} Fig. \ref{fig:arch-bicameral-frozen} in appendix). We define a block as a set of convolutional layers between two pooling layers. A VGG16-based encoder is therefore composed of 5 blocks. A block is said ``frozen'' when the corresponding parameters remain unchanged during finetuning. Note that the choice of the layers to freeze is application-dependent because the levels of semantics to freeze depend on the differences between the source and target domains.

Note also that we consider D2SA instead of PIOD or COCOA for transfer learning from Mikado because the data distributions of PIOD and COCOA are very different from Mikado. Indeed, \citep{BenDavidLLP10, BenDavidBCKPV10} show that a low divergence between the source and target domain distributions is a necessary condition for the success of domain adaptation. Table \ref{tab:transferlearningpiod} in appendix empirically shows that this condition is not met for Mikado and PIOD. Unlike PIOD and COCOA, which contain natural images of indoor and urban scenes with people, cars and animals, D2SA and Mikado both contain top-view images of household objects in bulk.

\begin{table*}[t]
\centering
\setlength\mywidth{.09\linewidth}
\begin{tabular}{r|c|c|c|c|c|c}
\multicolumn{1}{c}{}&
\multicolumn{1}{c}{\vcenteredinclude{\includegraphics[width=\mywidth]{datasets-mikado-001}}}&
\multicolumn{1}{c}{\vcenteredinclude{\includegraphics[width=\mywidth]{datasets-piod-005}}}&
\multicolumn{1}{c}{\vcenteredinclude{\includegraphics[width=\mywidth]{datasets-d2s-001}}}&
\multicolumn{1}{c}{\vcenteredinclude{\includegraphics[width=\mywidth]{datasets-d2s-009}}}&
\multicolumn{1}{c}{\vcenteredinclude{\includegraphics[width=\mywidth]{datasets-mikadoplus-006}}}&
\multicolumn{1}{c}{\vcenteredinclude{\includegraphics[width=\mywidth]{datasets-cocoa-005}}}\\
\vspace{-.1cm}\\
Dataset: & \textbf{Mikado} & \textbf{PIOD} & \textbf{D2SA} & \textbf{D2SA+} & \textbf{Mikado+} & \textbf{COCOA}\\
\hline
Training images & 13,600 & 9,600 & 512 & 2,960 & 28,800 & 12,800\\
Validation images & 800 & 800 & 56 & 328 & 4,800 & 1,424 \\
Test images & 4,800 & 800 & 5,992 & 5,992 & --  & 1,323\\
\hline
Iterations per epoch & 1,700 & 1,200 & 64 & 370 & 3,600 & 1,600\\
\end{tabular}
\caption{Image folds for each dataset after offline augmentation.}
\label{tab:trainingsplits}
\end{table*}

\subsection{Training Settings} 

Each network is trained and tested in the same conditions (including fixed random seeds) using Caffe \citep{caffe}.

\paragraph{Data preparation}

The networks are not fed with the original images but 256$\times$256 sub-images randomly extracted from each original image, and augmented offline with random geometric transformations (flipping, scaling and rotation). The folds of Mikado and Mikado+ are defined such that a texture appears only in one of the three subsets. The folds of PIOD and D2SA are defined with respect to the initial split proposed by their authors. Specifically, the original training images are used for training or validation in our folds, and the original validation images for test. The original test images are never used as they are not publicly available.

\paragraph{Optimization}

We use the Adam solver \citep{adamsolver15} with $\beta_1=.9$, $\beta_2=.999$, $\epsilon=10^{-8}$, and an initial learning rate of $10^{-4}$. We add a $\ell_2$-regularization with a weight decay of $10^{-4}$. The batch size is set to 8, and the training images are randomly permuted at each epoch. Since we solve a non-convex optimization problem, without theoretical convergence guarantees, the number of training iterations is chosen for each dataset from an empiric analysis on training and validation subsets. As generally adopted, the optimization is stopped when the validation error stagnates or increases while the training error keeps decreasing.
\begin{itemize}
\item In our comparative experiments (Figures \ref{fig:segarchs} and \ref{fig:obdsarchs}), we stop each training after 60 epochs for both Mikado and PIOD. Due to hardware limitations, each score results from one data fold.
\item In our ablation study on bicameral structuring (Figure \ref{fig:bicameralAblation}), each optimization is stopped after 20 and 15 epochs for Mikado and PIOD respectively, and each score is averaged over three optimizations using different data folds.
\item In our transfer learning experiments (Figure \ref{fig:resultsd2s}), each finetuning on D2SA is stopped after 15 epochs, and each score is averaged over three optimizations using different data folds. Pretraining on Mikado+ is stopped after 30 epochs.
\end{itemize}
Details on the epochs and data folds for each dataset are provided in Table \ref{tab:trainingsplits}. Please note that although the chosen stopping criterion may not be optimal for reaching the best performances on each dataset, it is however sufficient for significative comparisons since each network is trained under the same conditions.

\paragraph{Initialization}

For all experiments, except finetuning from weights pretrained on Mikado or Mikado+ in our synthetic data plausiblity check, each network has its first encoder initialized with weights pretrained on ImageNet \citep{imagenet15}, and the remaining layers with the Xavier method \citep{xavierinit}. To avoid overfitting, each convolutional block is ended with a dropout layer (we set the dropout ratio to .5), except in the first encoder.

\section{Discussion}
\label{sec:results}

In this section, we argue in light of our experimental results that the proposed multicameral decoder is more effective for dense homogeneous layouts than alternative design patterns, and that the jointly proposed synthetic data is plausible with respect to real-world problems.

\subsection{On the Proposed Model}

\paragraph{Homogeneous layouts require a complex decoding process.}

When localizaling specific instances in dense homogeneous layouts, the decoding process has great importance because the pixel embeddings must discriminate between instances of the same object. Figure \ref{fig:segarchs} confirms that a multicameral design proves more effective on Mikado than state-of-the-art design patterns for capturing position-sensitive representations. Specifically, our MC6$\dagger$ design outperforms RED-Atrous, RED-Coords, and RED-Dense/E networks by 20.6, 7.8 and 5.1 points in AP respectively. We explain these differences as follows: RED-Atrous enlarges the receptive field at the lowest resolution, which may lead to overfitting the training object layouts or mistakenly capturing relations between similar patterns far away from each other; RED-Coords associates each latent representation with a global location, thereby reducing the generalizability of these representations; RED-Dense/E uses DenseNet121 encoder blocks to softly capture more complex image representations that can hardly be fully exploited within a simple decoding process. Using only a VGG16 encoder, our multicameral decoding process produces higher-quality segmentations and more contrasted pixel-wise decisions, as illustrated in Figure \ref{fig:results-mask}. Nevertheless, half-outlined instances still appear (see the third row of Figure \ref{fig:results-mask}), seemingly due to a lack of long-range pixel associations in the learned representations.

\paragraph{Structured decoding units improves the learning.}

The success of a multicameral design results from our design choices to structure the decoding process: cascading subtask-specific decoder and encoder-decoder units. As reported by Figure \ref{fig:segscores}, cascading simple decoders without intermediate supervisions gradually improves the performances. Starting from MC2, adding one decoder (MC3) increases AP by 1.8 points, adding another decoder (MC4) by 3 points. Furthermore, structuring the backpropagation signals with ordinal intermediate supervisions for instance boundary and occluding boundary side detections (MC4$\dagger$) enables an additional gain of 4 points. Finally appending an encoder-decoder unit for refining the segmentation (MC6$\dagger$) leads to an overall pixel-wise improvement of 9.3 points over MC2, a VGG16-based RED network without additional state-of-the-art components. All these experimental results confirm that encouraging subtask-specific feature through ordinal multiscale units is an effective design pattern for dense homogeneous layouts. 

\paragraph{Learning position-sensitive representations proves more effective late in the decoding process.}

A multicameral design can be enhanced by enlarging the receptive fields just before decoding the unoccluded instances (MC6$\dagger$-Atrous/D4). As reported by Figure \ref{fig:segscores}, MC6$\dagger$-Atrous/D4 outperforms MC6$\dagger$ by 1.2 points. Learning explicity position-sensitive representations late in the decoding process enhances the performances in alternative design upgrades. Specifically, Figure \ref{fig:segscores} reports various similar improvements. First, using coordinate-aware convolutions: between RED-Coords and MC2-Coords/D (note that MC2 and RED are equal); between MC2-Coords/D and MC6$\dagger$-Coords/D4. Second, using dilated convolutions: between MC2-Atrous/D and MC4$\star\dagger$-Atrous/D2; between MC4$\star\dagger$-Atrous/D2 and MC6$\dagger$-Atrous/D4. These observations strongly suggest that the use of position-sensitive transforms, which partially break the translation invariance property of convolutional layers, should be thought with respect to the convolutional and non-convolutional aspects of the learned task. We applied this principle in our MC6$\dagger$-Atrous/D4 design: instance-aware segmentation requires some translation variance, while occlusion-aware boundary detection does not.

\paragraph{Ordinal decoders are important for detecting occlusion-aware boundaries as well.}

Our discussion on the importance of structure decoding extends to the lower-level task of occlusion-aware instance boundary detection. As reported by Figure \ref{fig:obdsarchs}, a bicameral network trained for jointly detecting instance boundaries and occluding boundary sides (MC3$\dagger$) compares favorably with DOOBNet-like and MTAN-like designs. Specifically, our design increases AP in the high-recall regime for occlusions by 1.7 points and 1 point on Mikado and PIOD respectively. Indeed, a key difference between MC3$\dagger$ and these state-of-the-art structurings is the ordinal relation between our decoders to encourage subtask-specific feature reuse. A bicameral structure is particularly suited to occlusion-aware boundary detection because occluding boundary sides can be interpreted as instance boundaries translated in the direction of the occluding instance.

Our ablation study on bicameral structuring (Figure \ref{fig:bicameralAblation}) confirms this important aspect. Specifically, a bicameral structure, which combines a shared encoder and cascaded decoders, achieves the best overall performances on both Mikado and PIOD. A bicameral structure also compares favorably with bicameral decoders that partially share their layers.

\subsection{On the Proposed Synthetic Data}

\paragraph{Mikado enables a meaningful evaluation.}

We create Mika\-do for our evaluation because, to the best of our knowledge, dense homogeneous layouts are missing from the public datasets for occlusion-aware instance segmentation. Although Mikado is a synthetic dataset, it is valuable for a meaningful evaluation. Our experimental results in Figure \ref{fig:resultsd2s} show that Mikado enables transferable feature learning in line with \citep{YosinskiCBL14}. Specifically, we show that using synthetic representations learned from Mikado enables to better detect occlusion-aware instance boundaries on D2SA \citep{D2S18}. As reported by Figure \ref{fig:resultsd2shist}, a gain of more than 10 points in AP for boundaries and 9 points for occlusions is achieved when finetuning the proposed network on D2SA with the first three encoder blocks frozen after pretraining on Mikado, instead of training all the layers only on D2SA (see also Figure \ref{fig:arch-bicameral-frozen} in appendix). This gain is qualitatively corroborated by Figure \ref{fig:resultsd2sim}. It suggests that a network trained on Mikado, which contains more occlusion relations between instances than the D2SA images for finetuning, learns a more general notion of occlusion. Our simulation-based pretraining also proves more effective than D2SA+ \citep{D2S18}, \ie{} creating training images by overlaying manually isolated instances. Despite the domain shift between Mikado and D2SA, using simulation enables more physics-consistent rendering at boundaries and less redundancy in terms of poses, unlike brute-force overlaying of instance segments from real images. Furthermore, almost equivalent performances are achieved when reducing the number of human-labeled real images for finetuning. Figure \ref{fig:resultsd2shist} shows that a bicameral network finetuned on D2SA using only 25\% of the initial D2SA finetuning subset, with the first three encoder blocks frozen after pretraining on Mikado, still outperforms a bicameral network trained only on D2SA or D2SA+. All of these results confirm that the representations learned from Mikado are meaningful \wrt{} real-world setups.
 
\paragraph{Mikado+ leads to even better results.}

Unlike real-world datasets, a synthetic dataset is readily extensible. By enriching Mikado with 20 times more texture images, 15 times more background images and 4 mesh templates, namely Mikado+, the ranges of color, texture, shape, and pose variations are better represented. As shown by Figure \ref{fig:resultsd2shist}, this leads to more generalizable invariants. Specifically, pretraining on Mikado+ instead of training only on D2SA increases AP by 10.1 points for boundaries and 7.8 points for occlusions while using only 12.5\% of the initial D2SA finetuning set. By contrast, using Mikado in the same conditions leads to a gain of 3.4 points for boundaries and 4.1 points for occlusions. These results imply that Mikado+ enables to learn more abstract local representations than Mikado. However, when applied on D2SA without finetuning, a pretraining on Mikado+ proves less effective than on Mikado. Consistently with the results after finetuning on D2SA, this could be explained by an overgeneralization of the task-specific layers. The neurons indeed co-adapt to capture the most discriminative patterns that are not likely to be the colors nor the object and background textures in Mikado+. An over-randomization of the colors and textures may disconnect the learned representations from concrete examples. This has nevertheless the advantage of easing the finetuning on D2SA, as the real-world scenes then appear as one variation within the learned range of variations. All these observations are incentives to favor synthetic training data when pixel-wise annotations on real-world images are hardly collectable. Hand-made annotations may also hinder the training due to their inaccuracy and incompleteness. As illustrated by Figure \ref{fig:badresults} in appendix, a bicameral network trained on PIOD is able to fairly predict non-annotated boundaries, \eg{} internal boundaries of instances with holes, missing instances, or instances ambiguously considered as part of the background. Furthermore, objects with complex shape, such as houseplants, which are often coarsely annotated by humans, are finely delineated by the proposed network.

\section{Conclusion}
\label{sec:conclusion}

We aimed at outlining unoccluded instances in dense homogeneous layouts, using a deep residual encoder-decoder design. However, decoding translation-invariant representations becomes problematic for distinguishing identical instances.
Unlike the state-of-the-art solutions which strengthen the encoder while reducing the decoder to a mere upsampling branch, we increased the complexity in the decoder by coupling decoder and encoder-decoder units in cascade, using resolution-wise skip connections. We also introduced a synthetic data generation pipeline (Mikado) to produce images of dense homogeneous layouts, as this scenario is missing from the public datasets. Our experiments on Mikado and PIOD showed that: (i) a multicameral design gives better results than aggregated dilated or coordinate-aware convolutions; (ii) ordinal multiscale latent representations improve the attention to unoccluded instances; (iii) design patterns for reducing the translation invariance are more efficient later in the decoding process. Furthermore, our experiments on transfer learning from Mikado to D2SA showed that a pretraining on Mikado enables state-of-the-art performances, while reducing by more than 85\% the number of real images for finetuning.

The proposed synthetically pretrained multicameral FCN establishes a new baseline for parsing images of dense homogeneous layouts. Nevertheless, there are still open research directions. Due to the ``horizontal'' skip connections, the number of filters severely increases with the number of decoding units, which may be prohibitive in terms of computational cost and memory requirements. It would be worth investigating optimization-based strategies, such as network architecture search approaches \citep{proxylessnas19, slimnets19}, to determine the optimal grid node and subtask ordering with respect to the application. Executing the model on the image at a lower-resolution then using adaptive sparse representations to iteratively refine the inferred boundaries could be another path to explore, as suggested by \citep{Kirillov19}. Furthermore, the proposed model does not explicitly exploit the redundancy within the scene. Yet, instances of the same object provide many cues to build an implicit object representation. Explicitly capturing the correspondences between the instances of a pile could be achieved using graph convolutional modules, in the same vein as dual graph networks for heterogeneous scenes \citep{Zhang2019dual}. Finally, a pretraining on Mikado requires some domain adaptation to achieve expert-level performances on a specific application. Although the proposed pretraining drastically reduces the need of annotations, producing the segmentation of a dense layout manually is very tedious. Coupling the proposed learning with a generative adversarial network \citep{DongYO018} or using self-supervision \citep{LeeNK19} would enable ordinal decoder units to adapt to novel conditions from unlabeled images. 

\begin{acknowledgements}
We thank Romain Br{\'e}gier, Florian Sella and the anonymous reviewers for their insightful comments and suggestions that helped us to greatly improve this article.
\end{acknowledgements}

\textbf{Note:} This is a pre-print of an article published in International Journal of Computer Vision, Special Issue on Deep Learning for Robotic Vision. The final authenticated version is available online at: 

https://doi.org/10.1007/s11263-020-01323-0



\appendix

\begin{figure*}
\centering
\setlength{\mywidth}{.115\textwidth}
\setlength{\mywidthbis}{-0.25cm} 
\setlength{\tabcolsep}{1pt}
\begin{tabular}{ccccccccc}
(a) &
\vcenteredinclude{\includegraphics[width=\mywidth]{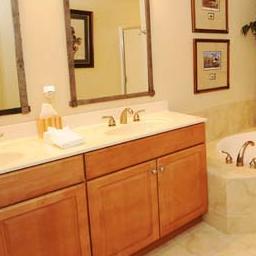}}&
\vcenteredinclude{\includegraphics[width=\mywidth]{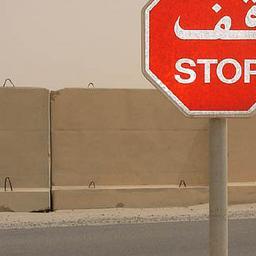}}&
\vcenteredinclude{\includegraphics[width=\mywidth]{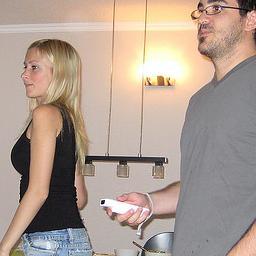}}&
\vcenteredinclude{\includegraphics[width=\mywidth]{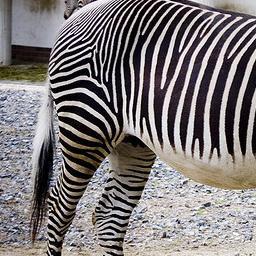}}&
\vcenteredinclude{\includegraphics[width=\mywidth]{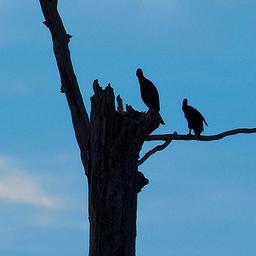}}&
\vcenteredinclude{\includegraphics[width=\mywidth]{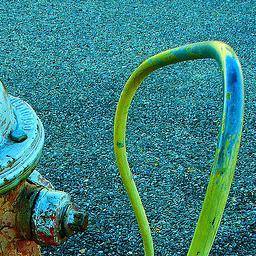}}&
\vcenteredinclude{\includegraphics[width=\mywidth]{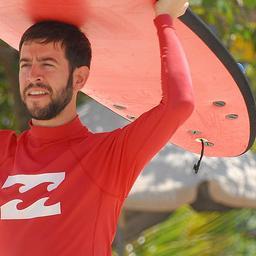}}&
\vcenteredinclude{\includegraphics[width=\mywidth]{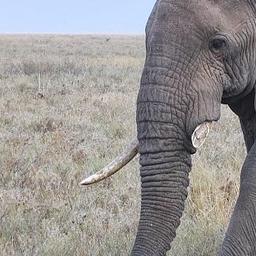}}\\
\vspace{\mywidthbis}\\
(b) & 
\vcenteredinclude{\includegraphics[width=\mywidth]{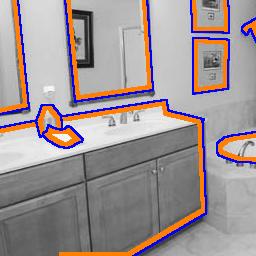}}&
\vcenteredinclude{\includegraphics[width=\mywidth]{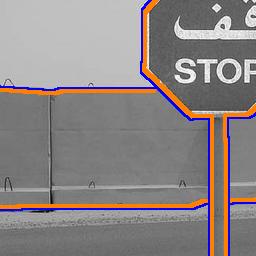}}&
\vcenteredinclude{\includegraphics[width=\mywidth]{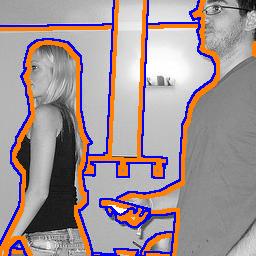}}&
\vcenteredinclude{\includegraphics[width=\mywidth]{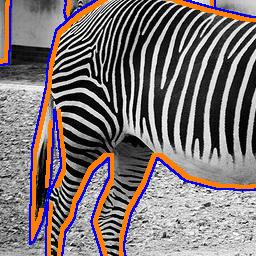}}&
\vcenteredinclude{\includegraphics[width=\mywidth]{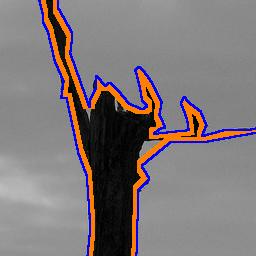}}&
\vcenteredinclude{\includegraphics[width=\mywidth]{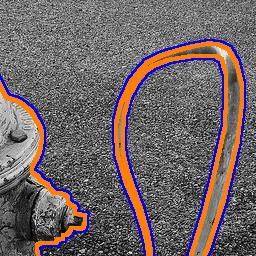}}&
\vcenteredinclude{\includegraphics[width=\mywidth]{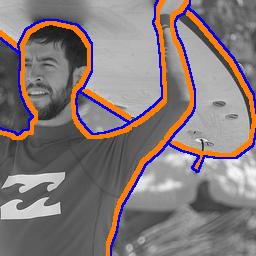}}&
\vcenteredinclude{\includegraphics[width=\mywidth]{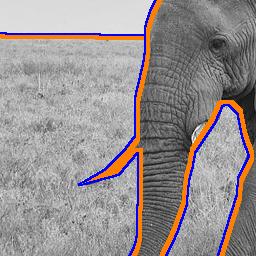}}\\
\vspace{\mywidthbis}\\
(c) &
\vcenteredinclude{\includegraphics[width=\mywidth]{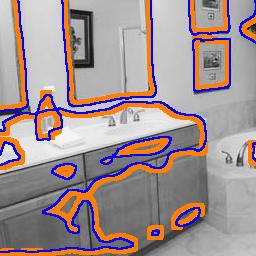}}&
\vcenteredinclude{\includegraphics[width=\mywidth]{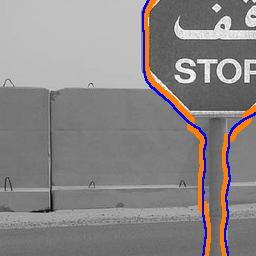}}&
\vcenteredinclude{\includegraphics[width=\mywidth]{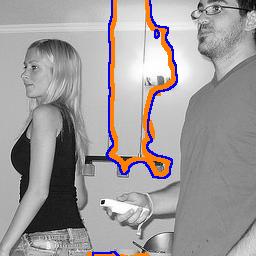}}&
\vcenteredinclude{\includegraphics[width=\mywidth]{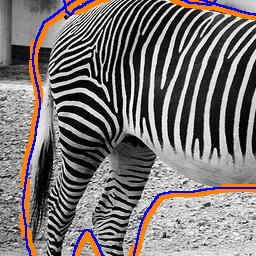}}&
\vcenteredinclude{\includegraphics[width=\mywidth]{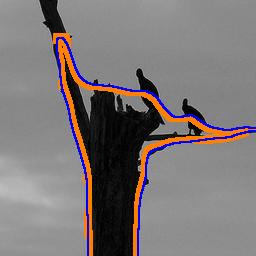}}&
\vcenteredinclude{\includegraphics[width=\mywidth]{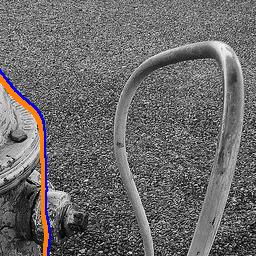}}&
\vcenteredinclude{\includegraphics[width=\mywidth]{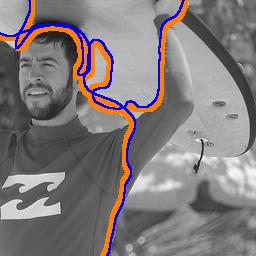}}&
\vcenteredinclude{\includegraphics[width=\mywidth]{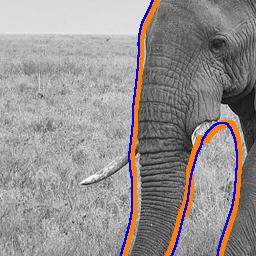}}\\
\vspace{\mywidthbis}\\
\textbf{(d)} &
\vcenteredinclude{\includegraphics[width=\mywidth]{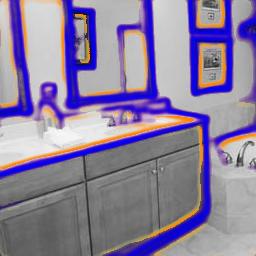}}&
\vcenteredinclude{\includegraphics[width=\mywidth]{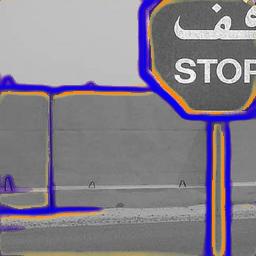}}&
\vcenteredinclude{\includegraphics[width=\mywidth]{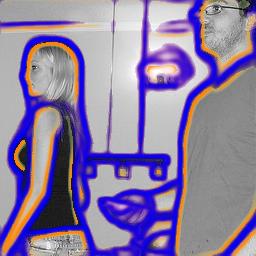}}&
\vcenteredinclude{\includegraphics[width=\mywidth]{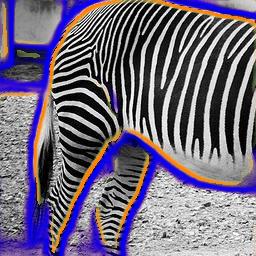}}&
\vcenteredinclude{\includegraphics[width=\mywidth]{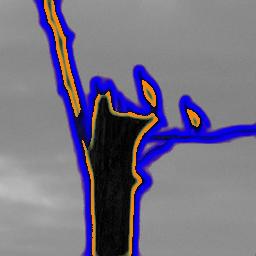}}&
\vcenteredinclude{\includegraphics[width=\mywidth]{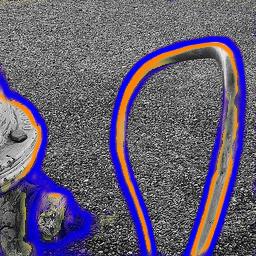}}&
\vcenteredinclude{\includegraphics[width=\mywidth]{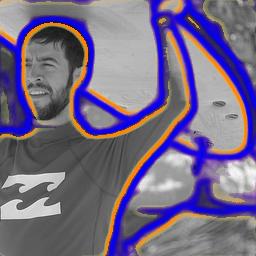}}&
\vcenteredinclude{\includegraphics[width=\mywidth]{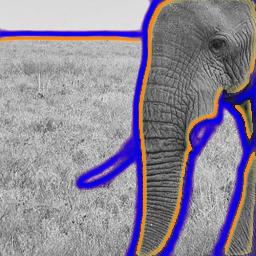}}\\
\vspace{\mywidthbis}\\
\end{tabular}

\bigskip

\setlength\tabcolsep{5pt}
\begin{tabular}{l|c|c|c|c|c|c|c|c|c|c|c|c}
\multirow{3}*{\textbf{Approach}} & \multicolumn{4}{c|}{\textit{All regions}} &  \multicolumn{4}{c|}{\textit{Things$^{1}$ only}} &  \multicolumn{4}{c}{\textit{Stuff$^{1}$ only}} \\
\cline{2-13}
& \multicolumn{2}{c|}{\textbf{Boundaries}} & \multicolumn{2}{c|}{\textbf{Occlusions}} & \multicolumn{2}{c|}{\textbf{Boundaries}} & \multicolumn{2}{c|}{\textbf{Occlusions}} & \multicolumn{2}{c|}{\textbf{Boundaries}} & \multicolumn{2}{c}{\textbf{Occlusions}} \\
& ODS&AP&ODS&AP & ODS&AP&ODS&AP & ODS&AP&ODS&AP \\
\hline
(c) Amodal segmentation $^{2}$ & .492 & -- & .529 & -- & .536 & -- & .608 & -- & .489 & -- & .397 & -- \\%
(d) \textbf{MC3$\dagger$ (Ours)} & \textbf{.666} & .694 & \textbf{.637} & .673 & \textbf{.666} & .690 & \textbf{.640} & .674 & \textbf{.687} & .727 & \textbf{.648} & .693 \\
\end{tabular}

\vspace{-.1cm}

\begin{flushleft}
\small
$^{1}$ Things are objects with well-defined shape, \eg{} car, person, and stuff instances amorphous regions, \eg{} grass, sky \citep{cocostuff18}.\\%

$^{2}$ The evaluation is performed on the binary segment proposals made available by the authors. We derive occlusion-aware boundaries from the ground truth and the precomputed results alike: after intersecting the modal and amodal masks of an instance, the amodal pixels that don't belong to the intersection are considered as occluded.
\end{flushleft}

\caption{Comparative results for instance boundary (blue) and unoccluded boundary side (orange) detection on COCOA. From top to bottom: input (a), ground truth (b), inference by amodal instance segmentation \citep{Zhu17amodal} (c), using a bicameral structure (d). Unlike the proposed approach, using a region proposal-based detection qualitatively leads to coarse segmentations and non-detected instances. Best viewed in color.}

\label{fig:resultscocoa}
\end{figure*}

\begin{figure*}
\centering

\setlength{\mywidth}{.11\textwidth}%
\setlength\mywidthbis{.45\textwidth}
\begin{minipage}{.5\linewidth}
\fontsize{7}{8}\selectfont%
\setlength{\secondcolumn}{.2cm}%
\tikzstyle{data} = [rectangle, draw, fill=white!20,%
    text width=\mywidthbis, text centered, rounded corners, minimum height=.5mm]%
\tikzstyle{conv} = [rectangle, draw, fill=blue!20,%
    text width=2em, text centered, rounded corners=0.5mm, text height=1.5mm, inner sep=0]%
\tikzstyle{sum} = [circle, draw, fill=blue!20, text height=1mm, inner sep=0]%
\tikzstyle{relu} = [rectangle, draw, fill=red!20,minimum width=3em]%
\tikzstyle{pool} = [rectangle, draw, fill=orange!20,text width=2em,text height=1mm, inner sep=0]%
\tikzstyle{unpool} = [rectangle, draw, fill=orange!80,text width=2em,text height=1mm, inner sep=0]%
\begin{tikzpicture}[node distance = 2mm, auto]%
\node [data] (rendering) {\textbf{Top-view camera (RGB and depth) rendering}};%
\node [draw=none,fill=none,text width=1.25\mywidth,align=center,below=0.02cm of rendering] (segmentation) {\includegraphics[height=\mywidth]{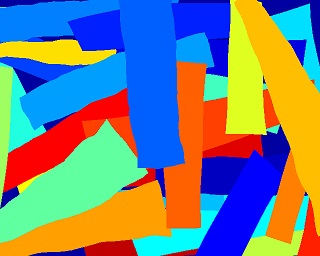}\\Segmentation};%
\node [draw=none,fill=none,text width=1.25\mywidth,align=center,left=.6\secondcolumn of segmentation] (rgb) {\includegraphics[height=\mywidth]{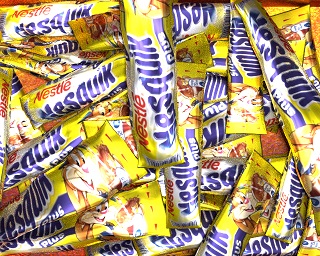}\\RGB};%
\node [draw=none,fill=none,text width=1.25\mywidth,align=center,right=.6\secondcolumn of segmentation] (depth) {\includegraphics[height=\mywidth]{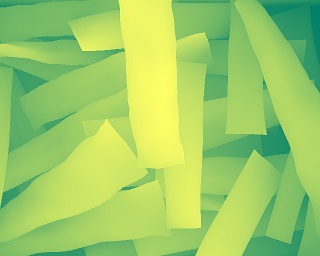}\\Depth};%
\node [data, below=0.1cm of segmentation] (generation) {\textbf{Generating ground-truth boundaries and occlusions}};%
\node [draw=none,fill=none,text width=1.25\mywidth,align=center,below=0.03cm of generation] (patches) {\includegraphics[height=\mywidth]{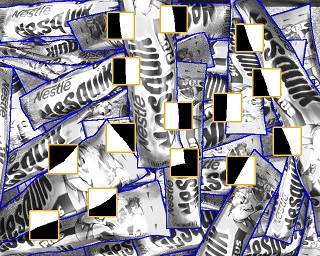}\\Local depth-based\\segmentations};%
\node [draw=none,fill=none,text width=1.25\mywidth,align=center,left=.6\secondcolumn of patches] (boundaries) {\includegraphics[height=\mywidth]{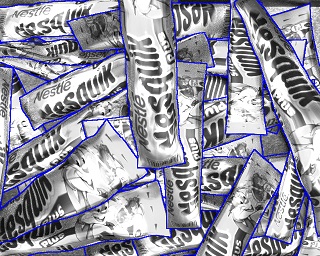}\\Instance\\boundaries};%
\node [draw=none,fill=none,text width=1.25\mywidth,align=center,right=.6\secondcolumn of patches] (occlusions) {\includegraphics[height=\mywidth]{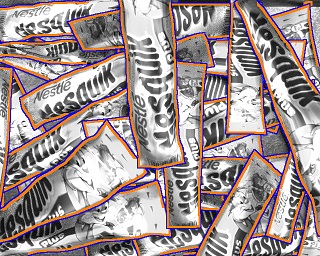}\\Boundaries and\\unoccluded side};%
\node [data, below=0.1cm of patches] (preparation) {\textbf{Training and test data preparation}};%
\draw[-{Latex[length=1mm,width=1mm]}] (segmentation) -- (generation);%
\draw[-{Latex[length=1mm,width=1mm]}] (patches) -- (preparation);%
\draw[-{Latex[length=1mm,width=1mm]}] (boundaries) -- (patches);%
\draw[-{Latex[length=1mm,width=1mm]}] (patches) -- (occlusions);%
\end{tikzpicture}
\end{minipage}\begin{minipage}{.3\linewidth}
\begin{flushleft}%
(a) Pipeline for generating the ground-truth occluding boundary side. At each boundary pixel, a depth-based binary segmentation of the neighborhood is performed to label each side, such that the higher side is set to 1 and the lower side to 0.%
\end{flushleft}
\end{minipage}

\setlength{\mywidth}{.85\linewidth}
\subfloat[Overview of the sachet textures used for generating Mikado.]{\includegraphics[width=\mywidth]{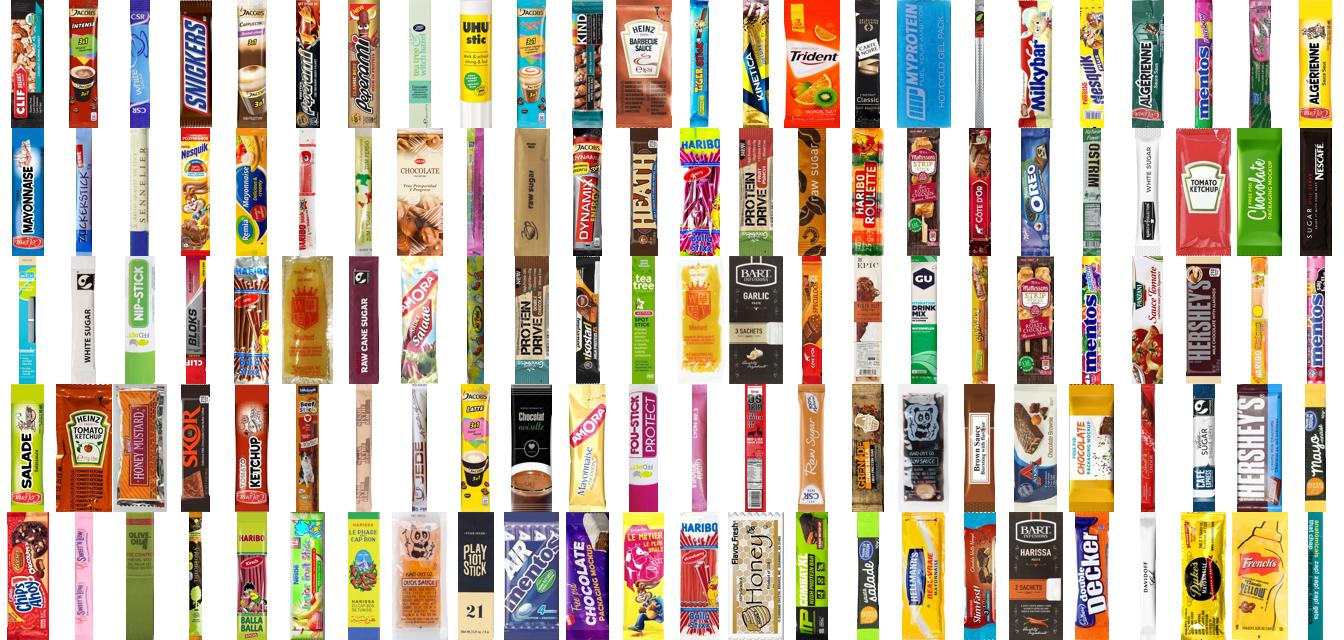}}

\subfloat[Overview of the background textures used for generating Mikado.]{\includegraphics[width=\mywidth]{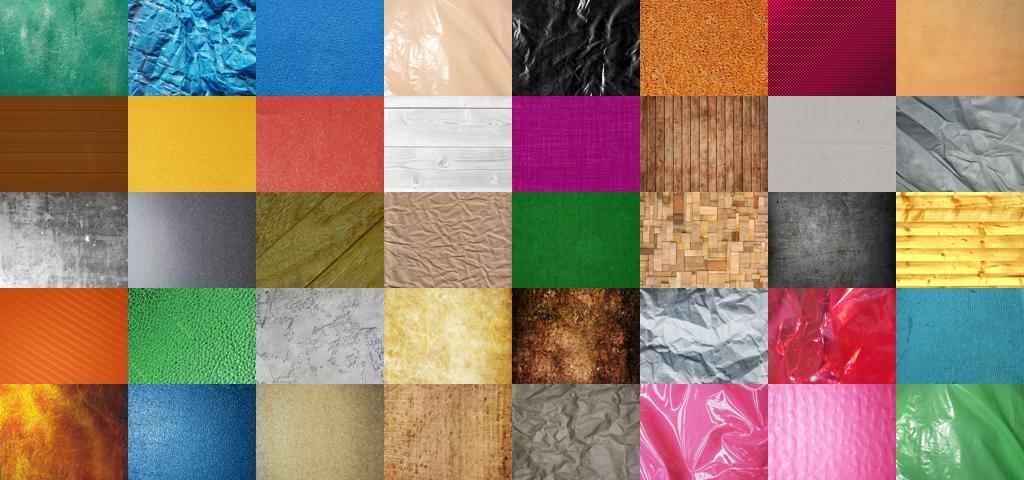}}
\caption{Supplementary material on the proposed synthetic data generation pipeline.}
\label{fig:mikadoSupp}
\end{figure*}

\begin{figure*}
\centering

\subfloat[From top to bottom: input (i), ground truth (ii), inference using two independent streams (iii), using a bicameral structure (iv). Instance boundaries are in blue, their unoccluded side in orange. Red rectangles highlight some false positive erased when using instead a single encoder shared by cascaded decoders.]{%
\setlength{\mywidth}{.115\textwidth}
\setlength{\mywidthbis}{-0.25cm} 
\setlength{\tabcolsep}{1pt}
\begin{tabular}{ccccccp{4pt}ccc}
& \multicolumn{5}{c}{\textbf{PIOD}} && \multicolumn{3}{c}{\textbf{Mikado}} \\
(i) &
\vcenteredinclude{\includegraphics[width=\mywidth]{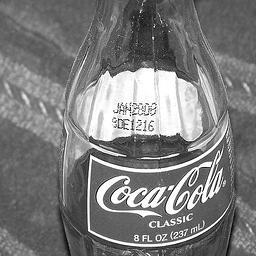}}&
\vcenteredinclude{\includegraphics[width=\mywidth]{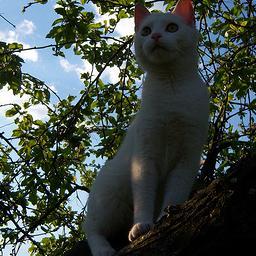}}&
\vcenteredinclude{\includegraphics[width=\mywidth]{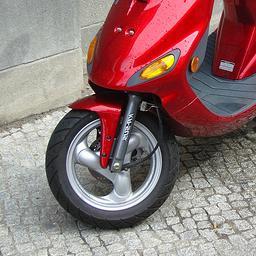}}&
\vcenteredinclude{\includegraphics[width=\mywidth]{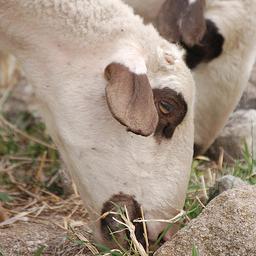}}&
\vcenteredinclude{\includegraphics[width=\mywidth]{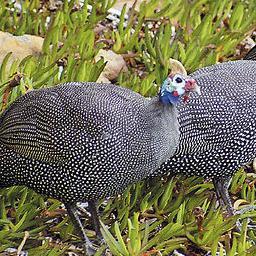}}&&
\vcenteredinclude{\includegraphics[width=\mywidth]{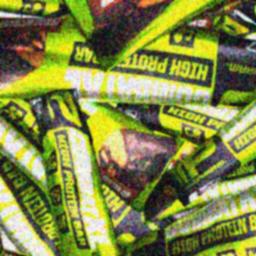}}&
\vcenteredinclude{\includegraphics[width=\mywidth]{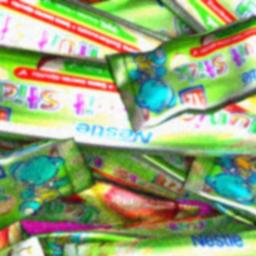}}&
\vcenteredinclude{\includegraphics[width=\mywidth]{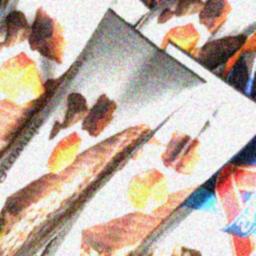}}\\
\vspace{\mywidthbis}\\
(ii) & 
\vcenteredinclude{\includegraphics[width=\mywidth]{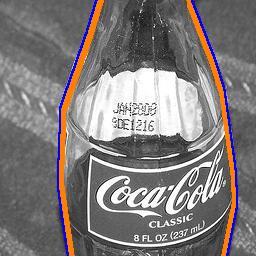}}&
\vcenteredinclude{\includegraphics[width=\mywidth]{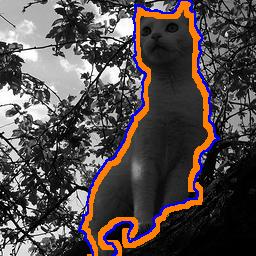}}&
\vcenteredinclude{\includegraphics[width=\mywidth]{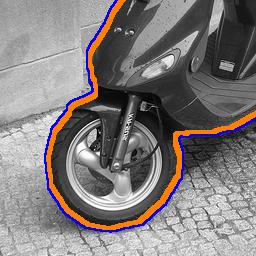}}&
\vcenteredinclude{\includegraphics[width=\mywidth]{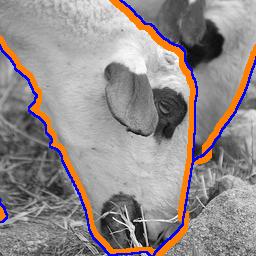}}&
\vcenteredinclude{\includegraphics[width=\mywidth]{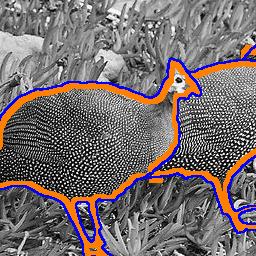}}&&
\vcenteredinclude{\includegraphics[width=\mywidth]{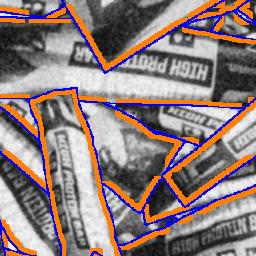}}&
\vcenteredinclude{\includegraphics[width=\mywidth]{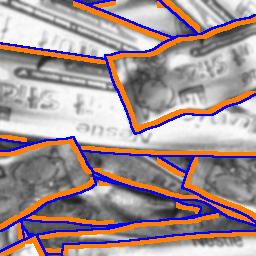}}&
\vcenteredinclude{\includegraphics[width=\mywidth]{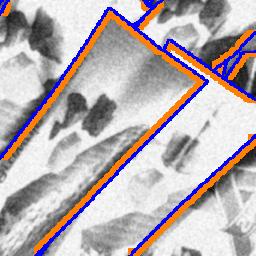}}\\
\vspace{\mywidthbis}\\
(iii) &
\vcenteredinclude{\includegraphics[width=\mywidth]{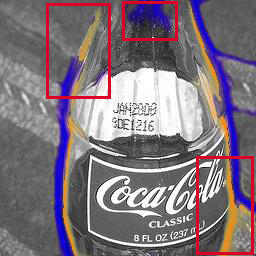}}&
\vcenteredinclude{\includegraphics[width=\mywidth]{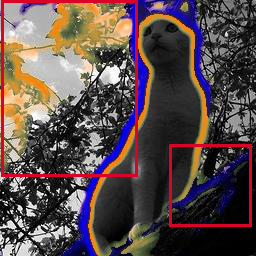}}&
\vcenteredinclude{\includegraphics[width=\mywidth]{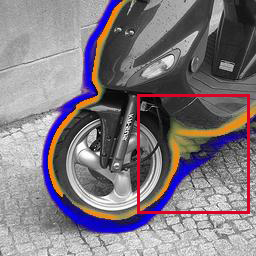}}&
\vcenteredinclude{\includegraphics[width=\mywidth]{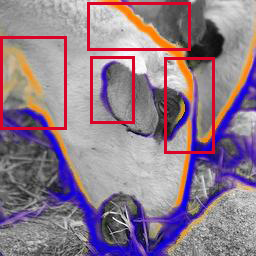}}&
\vcenteredinclude{\includegraphics[width=\mywidth]{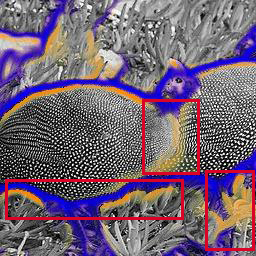}}&&
\vcenteredinclude{\includegraphics[width=\mywidth]{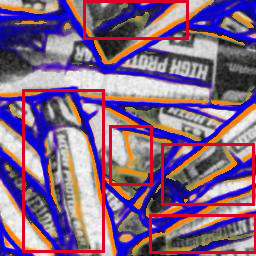}}&
\vcenteredinclude{\includegraphics[width=\mywidth]{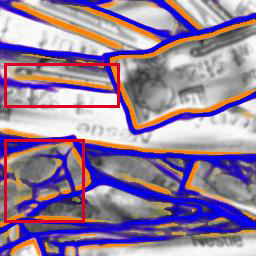}}&
\vcenteredinclude{\includegraphics[width=\mywidth]{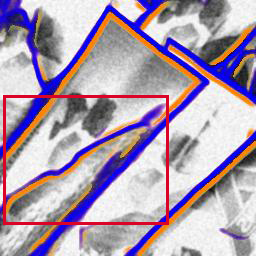}}\\
\vspace{\mywidthbis}\\
\textbf{(iv)} &
\vcenteredinclude{\includegraphics[width=\mywidth]{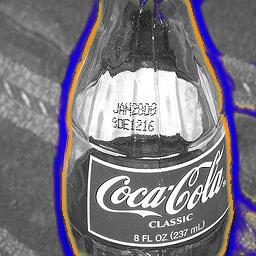}}&
\vcenteredinclude{\includegraphics[width=\mywidth]{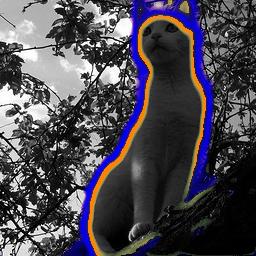}}&
\vcenteredinclude{\includegraphics[width=\mywidth]{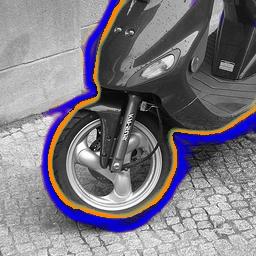}}&
\vcenteredinclude{\includegraphics[width=\mywidth]{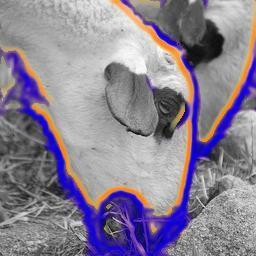}}&
\vcenteredinclude{\includegraphics[width=\mywidth]{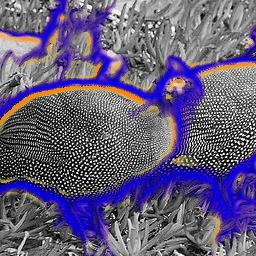}}&&
\vcenteredinclude{\includegraphics[width=\mywidth]{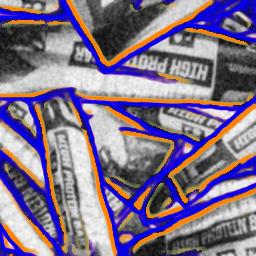}}&
\vcenteredinclude{\includegraphics[width=\mywidth]{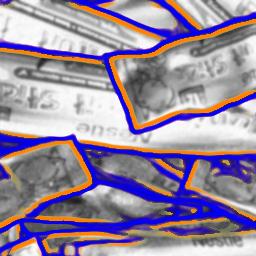}}&
\vcenteredinclude{\includegraphics[width=\mywidth]{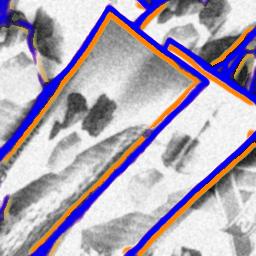}}\\
\vspace{\mywidthbis}\\
\end{tabular}%
}

\subfloat[From top to bottom: input (i), ground-truth (ii), inference using a bicameral structure (iv). The proposed network fairly infers non-annotated boundaries and delineates instances coarsely annotated by humans.]{%
\label{fig:badresults}%
\setlength{\mywidth}{.104\textwidth}
\setlength{\tabcolsep}{1pt}
\begin{tabular}{cccccccccc}
(i) &
\vcenteredinclude{\includegraphics[width=\mywidth]{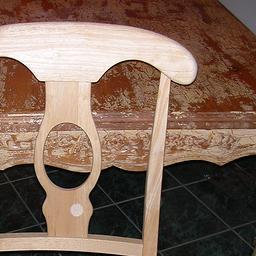}}&
\vcenteredinclude{\includegraphics[width=\mywidth]{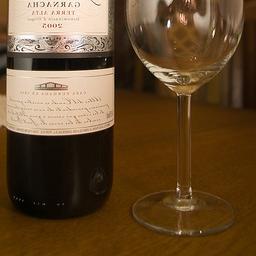}}&
\vcenteredinclude{\includegraphics[width=\mywidth]{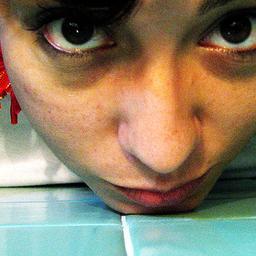}}&
\vcenteredinclude{\includegraphics[width=\mywidth]{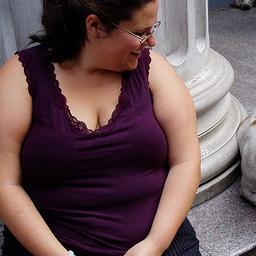}}&
\vcenteredinclude{\includegraphics[width=\mywidth]{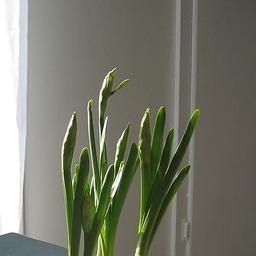}}&
\vcenteredinclude{\includegraphics[width=\mywidth]{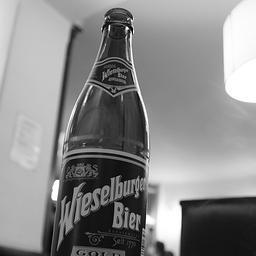}}&
\vcenteredinclude{\includegraphics[width=\mywidth]{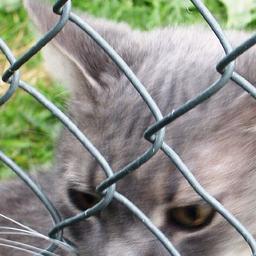}}&
\vcenteredinclude{\includegraphics[width=\mywidth]{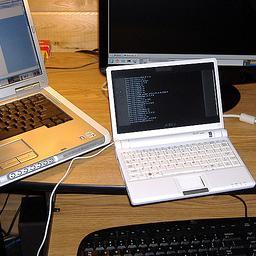}}&
\vcenteredinclude{\includegraphics[width=\mywidth]{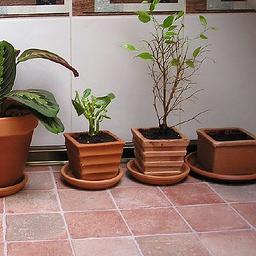}}\\
\vspace{-0.3cm}\\
(ii) & 
\vcenteredinclude{\includegraphics[width=\mywidth]{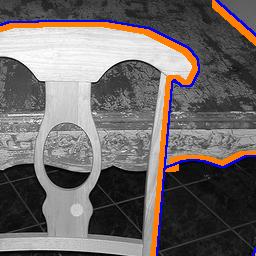}}&
\vcenteredinclude{\includegraphics[width=\mywidth]{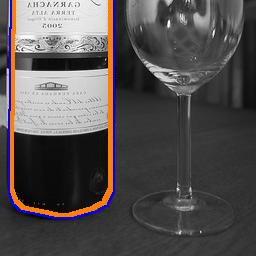}}&
\vcenteredinclude{\includegraphics[width=\mywidth]{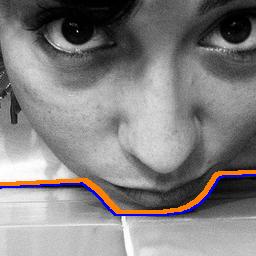}}&
\vcenteredinclude{\includegraphics[width=\mywidth]{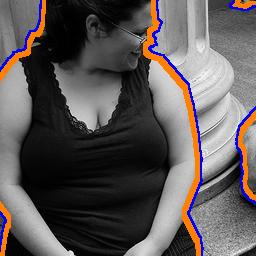}}&
\vcenteredinclude{\includegraphics[width=\mywidth]{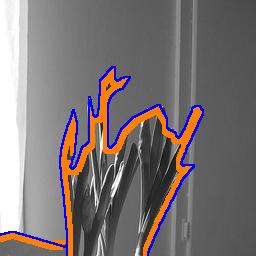}}&
\vcenteredinclude{\includegraphics[width=\mywidth]{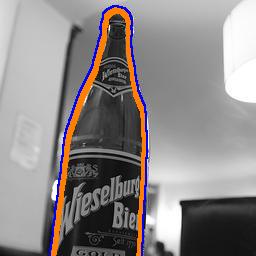}}&
\vcenteredinclude{\includegraphics[width=\mywidth]{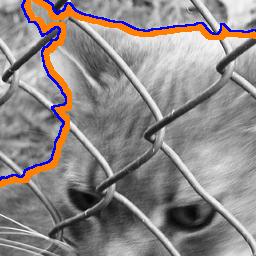}}&
\vcenteredinclude{\includegraphics[width=\mywidth]{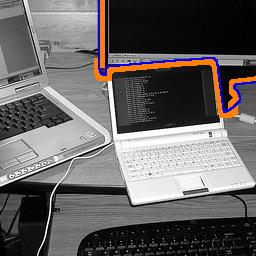}}&
\vcenteredinclude{\includegraphics[width=\mywidth]{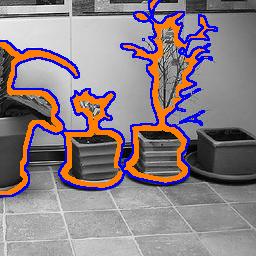}}\\
\vspace{-0.3cm}\\
\textbf{(iv)} &
\vcenteredinclude{\includegraphics[width=\mywidth]{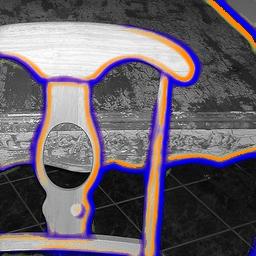}}&
\vcenteredinclude{\includegraphics[width=\mywidth]{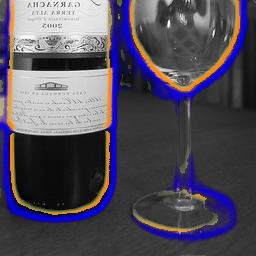}}&
\vcenteredinclude{\includegraphics[width=\mywidth]{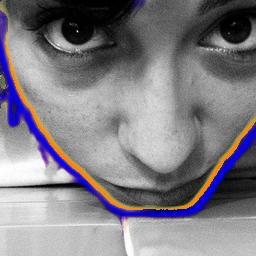}}&
\vcenteredinclude{\includegraphics[width=\mywidth]{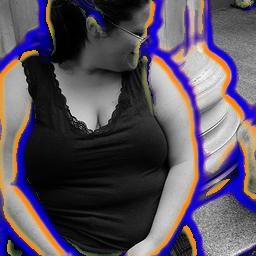}}&
\vcenteredinclude{\includegraphics[width=\mywidth]{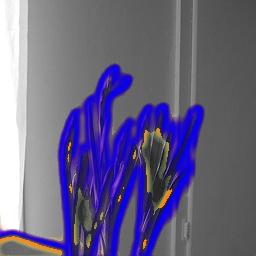}}&
\vcenteredinclude{\includegraphics[width=\mywidth]{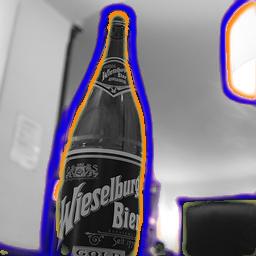}}&
\vcenteredinclude{\includegraphics[width=\mywidth]{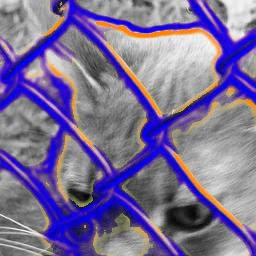}}&
\vcenteredinclude{\includegraphics[width=\mywidth]{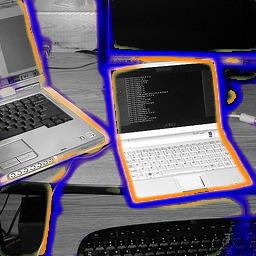}}&
\vcenteredinclude{\includegraphics[width=\mywidth]{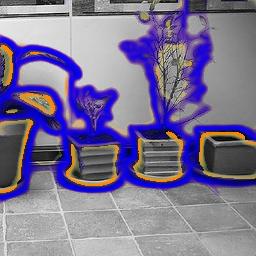}}\\
\end{tabular}%
}

\subfloat[Cross-dataset performances between Mikado and PIOD using a bicameral design. Both datasets perform poorly on each other because they follow very different texture, shape, and pose distributions.]{\label{tab:transferlearningpiod}%
\begin{tabular}{r|c|c|c|c}
&\multicolumn{4}{c}{Tests on \textbf{Mikado}}\\
\hline
\multirow{2}*{Trained on} & \multicolumn{2}{c|}{\textbf{Boundaries}} & \multicolumn{2}{c}{\textbf{Occlusions}} \\
&ODS&AP&ODS&AP\\
 \hline
\textbf{Mikado} & .769 & .847 & .801 & .884 \\
\textbf{\textbf{PIOD}} & .300 & .233 & .326 & .267\\
\end{tabular}%
\hspace{2cm}%
\begin{tabular}{r|c|c|c|c}
&\multicolumn{4}{c}{Tests on \textbf{PIOD}}\\
\hline
\multirow{2}*{Trained on} & \multicolumn{2}{c|}{\textbf{Boundaries}} & \multicolumn{2}{c}{\textbf{Occlusions}} \\
&ODS&AP&ODS&AP\\
 \hline
\textbf{\textbf{PIOD}} & .697 & .738 & .692 & .747 \\
\textbf{Mikado} & .405 & .350 & .400 & .349 \\
\end{tabular}
}
\caption{Comparative results for occlusion-aware boundary detection on PIOD and Mikado. Best viewed in color.}

\label{fig:results-full}
\end{figure*}

\begin{figure*}
\centering

\subfloat[Left: a bicameral structure with and without skip connections. Right: different skip connection types for merging two 2-channel feature vectors ($c_1,c_2$) and ($c_3,c_4$) into a new 2-channel one, using parameters $w_i$ and $w'_i$. From top to bottom: by element-wise max (i); by element-wise sum (ii); by concatenation (iii).]{%
\fontsize{7}{8}\selectfont
\setlength{\thirdcolumn}{.9cm}
\setlength\tabcolsep{8pt}
\begin{tabular}{cc}
\vcenteredinclude{
\tikzstyle{data} = [rectangle, draw, fill=white!20,%
    text width=2em, text centered, rounded corners, minimum height=.5mm]%
\tikzstyle{conv} = [rectangle, draw, fill=blue!20,%
    text width=2em, text centered, rounded corners=0.5mm, text height=1.5mm, inner sep=0]%
\tikzstyle{hideconv} = [conv,draw=white,fill=white]%
\tikzstyle{sum} = [circle, draw, fill=blue!20, text height=1mm, inner sep=0]%
\tikzstyle{relu} = [rectangle, draw, fill=red!20,minimum width=3em]%
\tikzstyle{pool} = [rectangle, draw, fill=orange!20,text width=2em,text height=1mm, inner sep=0]%
\tikzstyle{unpool} = [rectangle, draw, fill=orange!80,text width=2em,text height=1mm, inner sep=0]%
\begin{tikzpicture}[node distance = 2mm, auto]%
\node [data] (image) {in};%
\node [conv, below=1mm of image] (conv11) {};%
\node [conv, below of=conv11] (conv12) {};%
\node [pool, below of=conv12] (pool1) {};%
\node [conv, below of=pool1] (conv21) {};%
\node [conv, below of=conv21] (conv22) {};%
\node [pool, below of=conv22] (pool2) {};%
\node [conv, below of=pool2] (conv31) {};%
\node [conv, below of=conv31] (conv32) {};%
\node [conv, below of=conv32] (conv33) {};%
\node [pool, below of=conv33] (pool3) {};%
\node [conv, below of=pool3] (conv41) {};%
\node [conv, below of=conv41] (conv42) {};%
\node [conv, below of=conv42] (conv43) {};%
\node [pool, below of=conv43] (pool4) {};%
\node [conv, below of=pool4] (conv51) {};%
\node [conv, below of=conv51] (conv52) {};%
\node [conv, below of=conv52] (conv53) {};%
%
\node [unpool, right=2\thirdcolumn of pool1.center,anchor=center] (unpool1b) {};%
\node [conv, right=2\thirdcolumn of conv12.center,anchor=center] (deconv1b) {};%
\node [unpool, right=2\thirdcolumn of pool2.center,anchor=center] (unpool2b) {};%
\node [conv, right=2\thirdcolumn of conv22.center,anchor=center] (deconv2b) {};%
\node [unpool, right=2\thirdcolumn of pool3.center,anchor=center] (unpool3b) {};%
\node [conv, right=2\thirdcolumn of conv33.center,anchor=center] (deconv3b) {};%
\node [unpool, right=2\thirdcolumn of pool4.center,anchor=center] (unpool4b) {};%
\node [conv, right=2\thirdcolumn of conv43.center,anchor=center] (deconv4b) {};%
%
\node [unpool, right=\thirdcolumn of pool1.center,anchor=center] (unpool1) {};%
\node [conv, right=\thirdcolumn of conv12.center,anchor=center] (deconv1) {};%
\node [unpool, right=\thirdcolumn of pool2.center,anchor=center] (unpool2) {};%
\node [conv, right=\thirdcolumn of conv22.center,anchor=center] (deconv2) {};%
\node [unpool, right=\thirdcolumn of pool3.center,anchor=center] (unpool3) {};%
\node [conv, right=\thirdcolumn of conv33.center,anchor=center] (deconv3) {};%
\node [unpool, right=\thirdcolumn of pool4.center,anchor=center] (unpool4) {};%
\node [conv, right=\thirdcolumn of conv43.center,anchor=center] (deconv4) {};%
\node [data, right=\thirdcolumn of image.center,anchor=center] (contours) {out1};%
\node [data, right=2\thirdcolumn of image.center,anchor=center] (contoursb) {out2};%
\draw[-{Latex[length=1mm,width=1mm]}] (image) -- (conv11);%
\draw[-{Latex[length=1mm,width=1mm]}] (conv11) -- (conv12);
\draw[-{Latex[length=1mm,width=1mm]}] (conv12) -- (pool1);%
\draw[-{Latex[length=1mm,width=1mm]}] (pool1) -- (conv21);%
\draw[-{Latex[length=1mm,width=1mm]}] (conv21) -- (conv22);%
\draw[-{Latex[length=1mm,width=1mm]}] (conv22) -- (pool2);%
\draw[-{Latex[length=1mm,width=1mm]}] (pool2) -- (conv31);%
\draw[-{Latex[length=1mm,width=1mm]}] (conv31) -- (conv32);%
\draw[-{Latex[length=1mm,width=1mm]}] (conv32) -- (conv33);%
\draw[-{Latex[length=1mm,width=1mm]}] (conv33) -- (pool3);%
\draw[-{Latex[length=1mm,width=1mm]}] (pool3) -- (conv41);%
\draw[-{Latex[length=1mm,width=1mm]}] (conv41) -- (conv42);%
\draw[-{Latex[length=1mm,width=1mm]}] (conv42) -- (conv43);%
\draw[-{Latex[length=1mm,width=1mm]}] (conv43) -- (pool4);%
\draw[-{Latex[length=1mm,width=1mm]}] (pool4) -- (conv51);%
\draw[-{Latex[length=1mm,width=1mm]}] (conv51) -- (conv52);%
\draw[-{Latex[length=1mm,width=1mm]}] (conv52) -- (conv53);%
\draw[-{Latex[length=1mm,width=1mm]}] (conv53) -| (unpool4);%
\draw[-{Latex[length=1mm,width=1mm]}] (unpool4) -- (deconv4);%
\draw[-{Latex[length=1mm,width=1mm]}] (deconv4) -- (unpool3);%
\draw[-{Latex[length=1mm,width=1mm]}] (unpool3) -- (deconv3);%
\draw[-{Latex[length=1mm,width=1mm]}] (deconv3) -- (unpool2);%
\draw[-{Latex[length=1mm,width=1mm]}] (unpool2) -- (deconv2);%
\draw[-{Latex[length=1mm,width=1mm]}] (deconv2) -- (unpool1);%
\draw[-{Latex[length=1mm,width=1mm]}] (unpool1) -- (deconv1);%
\draw[-{Latex[length=1mm,width=1mm]}] (deconv1) -- (contours);%
\draw[-{Latex[length=1mm,width=1mm]}] (conv53) -| (unpool4b);%
\draw[-{Latex[length=1mm,width=1mm]}] (unpool4b) -- (deconv4b);%
\draw[-{Latex[length=1mm,width=1mm]}] (deconv4b) -- (unpool3b);%
\draw[-{Latex[length=1mm,width=1mm]}] (unpool3b) -- (deconv3b);%
\draw[-{Latex[length=1mm,width=1mm]}] (deconv3b) -- (unpool2b);%
\draw[-{Latex[length=1mm,width=1mm]}] (unpool2b) -- (deconv2b);%
\draw[-{Latex[length=1mm,width=1mm]}] (deconv2b) -- (unpool1b);%
\draw[-{Latex[length=1mm,width=1mm]}] (unpool1b) -- (deconv1b);%
\draw[-{Latex[length=1mm,width=1mm]}] (deconv1b) -- (contoursb);%
\draw[-{Latex[length=1mm,width=1mm]}] (conv12) -- (deconv1);%
\draw[-{Latex[length=1mm,width=1mm]}] (conv22) -- (deconv2);%
\draw[-{Latex[length=1mm,width=1mm]}] (conv33) -- (deconv3);%
\draw[-{Latex[length=1mm,width=1mm]}] (conv43) -- (deconv4);%
\draw[-{Latex[length=1mm,width=1mm]}] (conv12.east) to [out=30,in=165] (deconv1b);%
\draw[-{Latex[length=1mm,width=1mm]}] (conv22.east) to [out=30,in=165] (deconv2b);%
\draw[-{Latex[length=1mm,width=1mm]}] (conv33.east) to [out=30,in=165] (deconv3b);%
\draw[-{Latex[length=1mm,width=1mm]}] (conv43.east) to [out=30,in=165] (deconv4b);%
\draw[-{Latex[length=1mm,width=1mm]}] (deconv1) -- (deconv1b);%
\draw[-{Latex[length=1mm,width=1mm]}] (deconv2) -- (deconv2b);%
\draw[-{Latex[length=1mm,width=1mm]}] (deconv3) -- (deconv3b);%
\draw[-{Latex[length=1mm,width=1mm]}] (deconv4) -- (deconv4b);%
%
%
\end{tikzpicture}%
}&
\vcenteredinclude{
\tikzstyle{data} = [rectangle, draw, fill=white!20,%
    text width=2em, text centered, rounded corners, minimum height=.5mm]%
\tikzstyle{conv} = [rectangle, draw, fill=blue!20,%
    text width=2em, text centered, rounded corners=0.5mm, text height=1.5mm, inner sep=0]%
\tikzstyle{sum} = [circle, draw, fill=blue!20, text height=1mm, inner sep=0]%
\tikzstyle{relu} = [rectangle, draw, fill=red!20,minimum width=3em]%
\tikzstyle{pool} = [rectangle, draw, fill=orange!20,text width=2em,text height=1mm, inner sep=0]%
\tikzstyle{unpool} = [rectangle, draw, fill=orange!80,text width=2em,text height=1mm, inner sep=0]%
\begin{tikzpicture}[node distance = 2mm, auto]%
\node [data] (image) {in};%
\node [conv, below=1mm of image] (conv11) {};%
\node [conv, below of=conv11] (conv12) {};%
\node [pool, below of=conv12] (pool1) {};%
\node [conv, below of=pool1] (conv21) {};%
\node [conv, below of=conv21] (conv22) {};%
\node [pool, below of=conv22] (pool2) {};%
\node [conv, below of=pool2] (conv31) {};%
\node [conv, below of=conv31] (conv32) {};%
\node [conv, below of=conv32] (conv33) {};%
\node [pool, below of=conv33] (pool3) {};%
\node [conv, below of=pool3] (conv41) {};%
\node [conv, below of=conv41] (conv42) {};%
\node [conv, below of=conv42] (conv43) {};%
\node [pool, below of=conv43] (pool4) {};%
\node [conv, below of=pool4] (conv51) {};%
\node [conv, below of=conv51] (conv52) {};%
\node [conv, below of=conv52] (conv53) {};%
\node [unpool, right=2\thirdcolumn of pool1.center,anchor=center] (unpool1b) {};%
\node [conv, right=2\thirdcolumn of conv12.center,anchor=center] (deconv1b) {};%
\node [unpool, right=2\thirdcolumn of pool2.center,anchor=center] (unpool2b) {};%
\node [conv, right=2\thirdcolumn of conv22.center,anchor=center] (deconv2b) {};%
\node [unpool, right=2\thirdcolumn of pool3.center,anchor=center] (unpool3b) {};%
\node [conv, right=2\thirdcolumn of conv33.center,anchor=center] (deconv3b) {};%
\node [unpool, right=2\thirdcolumn of pool4.center,anchor=center] (unpool4b) {};%
\node [conv, right=2\thirdcolumn of conv43.center,anchor=center] (deconv4b) {};%
\node [unpool, right=\thirdcolumn of pool1.center,anchor=center] (unpool1) {};%
\node [conv, right=\thirdcolumn of conv12.center,anchor=center] (deconv1) {};%
\node [unpool, right=\thirdcolumn of pool2.center,anchor=center] (unpool2) {};%
\node [conv, right=\thirdcolumn of conv22.center,anchor=center] (deconv2) {};%
\node [unpool, right=\thirdcolumn of pool3.center,anchor=center] (unpool3) {};%
\node [conv, right=\thirdcolumn of conv33.center,anchor=center] (deconv3) {};%
\node [unpool, right=\thirdcolumn of pool4.center,anchor=center] (unpool4) {};%
\node [conv, right=\thirdcolumn of conv43.center,anchor=center] (deconv4) {};%
\node [data, right=\thirdcolumn of image.center,anchor=center] (contours) {out1};%
\node [data, right=2\thirdcolumn of image.center,anchor=center] (contoursb) {out2};%
\draw[-{Latex[length=1mm,width=1mm]}] (image) -- (conv11);%
\draw[-{Latex[length=1mm,width=1mm]}] (conv11) -- (conv12);%
\draw[-{Latex[length=1mm,width=1mm]}] (conv12) -- (pool1);%
\draw[-{Latex[length=1mm,width=1mm]}] (pool1) -- (conv21);%
\draw[-{Latex[length=1mm,width=1mm]}] (conv21) -- (conv22);%
\draw[-{Latex[length=1mm,width=1mm]}] (conv22) -- (pool2);%
\draw[-{Latex[length=1mm,width=1mm]}] (pool2) -- (conv31);%
\draw[-{Latex[length=1mm,width=1mm]}] (conv31) -- (conv32);%
\draw[-{Latex[length=1mm,width=1mm]}] (conv32) -- (conv33);%
\draw[-{Latex[length=1mm,width=1mm]}] (conv33) -- (pool3);%
\draw[-{Latex[length=1mm,width=1mm]}] (pool3) -- (conv41);%
\draw[-{Latex[length=1mm,width=1mm]}] (conv41) -- (conv42);%
\draw[-{Latex[length=1mm,width=1mm]}] (conv42) -- (conv43);%
\draw[-{Latex[length=1mm,width=1mm]}] (conv43) -- (pool4);%
\draw[-{Latex[length=1mm,width=1mm]}] (pool4) -- (conv51);%
\draw[-{Latex[length=1mm,width=1mm]}] (conv51) -- (conv52);%
\draw[-{Latex[length=1mm,width=1mm]}] (conv52) -- (conv53);%
\draw[-{Latex[length=1mm,width=1mm]}] (conv53) -| (unpool4);%
\draw[-{Latex[length=1mm,width=1mm]}] (unpool4) -- (deconv4);%
\draw[-{Latex[length=1mm,width=1mm]}] (deconv4) -- (unpool3);%
\draw[-{Latex[length=1mm,width=1mm]}] (unpool3) -- (deconv3);%
\draw[-{Latex[length=1mm,width=1mm]}] (deconv3) -- (unpool2);%
\draw[-{Latex[length=1mm,width=1mm]}] (unpool2) -- (deconv2);%
\draw[-{Latex[length=1mm,width=1mm]}] (deconv2) -- (unpool1);%
\draw[-{Latex[length=1mm,width=1mm]}] (unpool1) -- (deconv1);%
\draw[-{Latex[length=1mm,width=1mm]}] (deconv1) -- (contours);%
\draw[-{Latex[length=1mm,width=1mm]}] (conv53) -| (unpool4b);%
\draw[-{Latex[length=1mm,width=1mm]}] (unpool4b) -- (deconv4b);%
\draw[-{Latex[length=1mm,width=1mm]}] (deconv4b) -- (unpool3b);%
\draw[-{Latex[length=1mm,width=1mm]}] (unpool3b) -- (deconv3b);%
\draw[-{Latex[length=1mm,width=1mm]}] (deconv3b) -- (unpool2b);%
\draw[-{Latex[length=1mm,width=1mm]}] (unpool2b) -- (deconv2b);%
\draw[-{Latex[length=1mm,width=1mm]}] (deconv2b) -- (unpool1b);%
\draw[-{Latex[length=1mm,width=1mm]}] (unpool1b) -- (deconv1b);%
\draw[-{Latex[length=1mm,width=1mm]}] (deconv1b) -- (contoursb);%
\end{tikzpicture}
}\\
\vspace{-.1cm}\\
With skip connections& Without skip connections\\
\end{tabular}%
\hfill%
\vcenteredinclude{%
\setlength{\mywidth}{.14\textwidth}%
\setlength{\secondcolumn}{.2cm}%
\setlength{\thirdcolumn}{1cm}%
\tikzstyle{data} = [rectangle, draw, fill=white!20,%
    text width=2em, text centered, rounded corners, minimum height=.5mm]%
\tikzstyle{conv} = [rectangle, draw, fill=blue!20,%
    text width=2em, text centered, rounded corners=0.5mm, text height=1.5mm, inner sep=0]%
\tikzstyle{sum} = [circle, draw, fill=blue!20, text height=1mm, inner sep=0]%
\tikzstyle{relu} = [rectangle, draw, fill=red!20,minimum width=3em]%
\tikzstyle{pool} = [rectangle, draw, fill=orange!20,text width=2em,text height=1mm, inner sep=0]%
\tikzstyle{unpool} = [rectangle, draw, fill=orange!80,text width=2em,text height=1mm, inner sep=0]%
\begin{tikzpicture}[node distance = 2mm, auto]%
\node [data] (image) {in};%
\node [data, below=4mm of image] (out1) {out1};%
\node [data, below=4mm of out1] (out2) {out2};%
\node [conv, below=5mm of out2] (conv1) {};%
\node [pool, below=6mm of conv1] (pool1) {};%
\node [unpool, below=6mm of pool1] (unpool1) {};%
\node[draw=none,fill=none,text width=\mywidth,  align=left, right=.6\secondcolumn of image] {Input image};%
\node[draw=none,fill=none,text width=\mywidth, align=left, right=.6\secondcolumn of out1] {Conv+Sigmoid\\ (Boundaries)};%
\node[draw=none,fill=none, text width=\mywidth, align=left, right=.6\secondcolumn of out2] {Conv+Sigmoid\\ (Occlusions)};%
\node[draw=none,fill=none,text width=\mywidth, align=left, right=.6\secondcolumn of conv1] {Concat+\\ Conv+ReLU};%
\node[draw=none,fill=none,text width=\mywidth, align=left, right=.6\secondcolumn of pool1] {Spatial\\ pooling (.5$\times$)};%
\node[draw=none,fill=none,text width=\mywidth, align=left, right=.6\secondcolumn of unpool1] (unpoolname) {Spatial\\ unpooling (2$\times$)};%
\end{tikzpicture}
}
\hfill%
\vcenteredinclude{%
\fontsize{10}{11}\selectfont%
\setlength\mywidth{5mm}
\setlength\mywidthbis{2mm}%
\setlength\secondcolumn{3mm}%
\tikzstyle{convarrow} = [-{Latex[length=1mm,width=2mm]},line width=.4mm]%
\tikzstyle{feature1} = [rectangle, draw, fill=teal!20,text width=2em,text centered]%
\tikzstyle{feature2} = [rectangle, draw, fill=green!20,text width=2em,text centered]%
\tikzstyle{feature3} = [rectangle, draw, fill=red!20,text width=8em,text centered, text height=.8em,text depth=.4em,minimum height=.45cm]%
%

}
}

\subfloat[Comparative performances on PIOD and Mikado.]{%
\begin{tabular}{l|c|c|c|c|c|c|c|c|c|c|c|c}
Dataset: &\multicolumn{6}{c|}{\textbf{PIOD}}& \multicolumn{6}{c}{\textbf{Mikado}}\\
\hline
\textbf{Skip connections?} & \multicolumn{3}{c|}{\textbf{Boundaries}} & \multicolumn{3}{c|}{\textbf{Occlusions}} & \multicolumn{3}{c|}{\textbf{Boundaries}} & \multicolumn{3}{c}{\textbf{Occlusions}} \\
\textbf{(Type)} &ODS&AP&AP$_{60}$&ODS&AP&AP$_{60}$&ODS&AP&AP$_{60}$&ODS&AP&AP$_{60}$\\
\hline
No & .693 & \textbf{.744} & .495 & .692 & \textbf{.749} & .520 & .759 & .834 & .686 & .793 & .878 & .748 \\
\hline
Yes (Element-wise max) & .685 & .729 & .512 & .676 & .731 & .522 & .755 & .830 & .676 & .786 & .871 & .735 \\
Yes (Element-wise sum) & .687 & .730 & .505 & .678 & .731 & .514 & .761 & .838 & .685 & .791 & .876 & .743 \\
\hline
\textbf{Yes (Concatenation)} &  \textbf{.697} & .738 & \textbf{.517} & \textbf{.692} & .747 & \textbf{.532} & \textbf{.769} & \textbf{.847} & \textbf{.698} & \textbf{.801} & \textbf{.884} & \textbf{.758} \\
\end{tabular}%
}

\subfloat[From top to bottom: input and ground truth (i), activation map after the affine transformation on top of the first unpooling layer of the boundary branch (ii), inference (iii). Combining spatial information and higher-level semantics at each scale using skip connections between the encoder and decoders enables to detect instance boundaries earlier when decoding.]{%
\setlength{\mywidth}{.11\textwidth}
\setlength{\mywidthbis}{-0.25cm} 
\setlength{\tabcolsep}{1pt}
\begin{tabular}{cccp{2pt}ccp{4pt}ccp{2pt}cc}
& \multicolumn{5}{c}{\textbf{PIOD}} && \multicolumn{5}{c}{\textbf{Mikado}} \\
(i) &
\vcenteredinclude{\includegraphics[width=\mywidth]{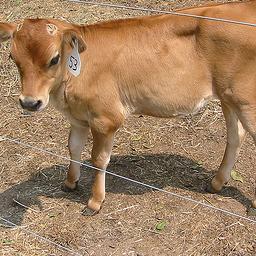}}&
\vcenteredinclude{\includegraphics[width=\mywidth]{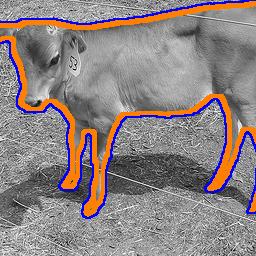}}&&
\vcenteredinclude{\includegraphics[width=\mywidth]{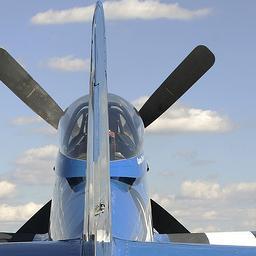}}&
\vcenteredinclude{\includegraphics[width=\mywidth]{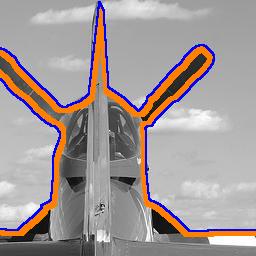}}&&
\vcenteredinclude{\includegraphics[width=\mywidth]{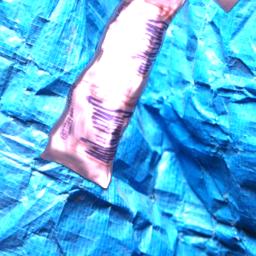}}&
\vcenteredinclude{\includegraphics[width=\mywidth]{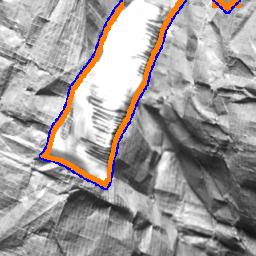}}&&
\vcenteredinclude{\includegraphics[width=\mywidth]{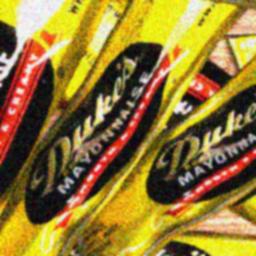}}&
\vcenteredinclude{\includegraphics[width=\mywidth]{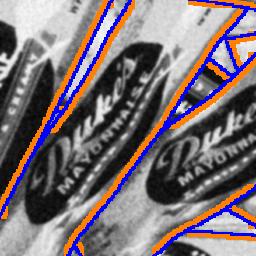}}\\
\vspace{0.5\mywidthbis}\\
(ii) & 
\vcenteredinclude{\includegraphics[width=\mywidth]{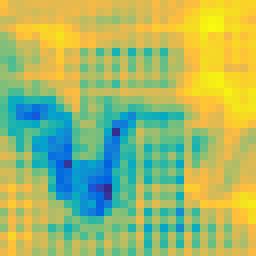}}&
\vcenteredinclude{\includegraphics[width=\mywidth]{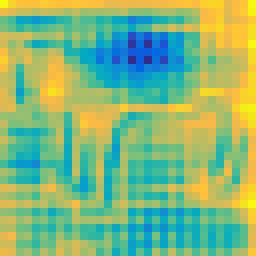}}&&
\vcenteredinclude{\includegraphics[width=\mywidth]{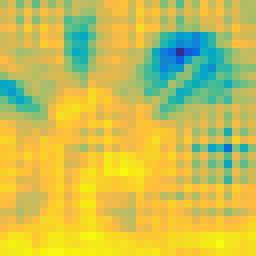}}&
\vcenteredinclude{\includegraphics[width=\mywidth]{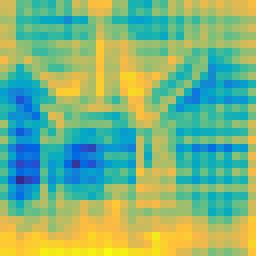}}&&
\vcenteredinclude{\includegraphics[width=\mywidth]{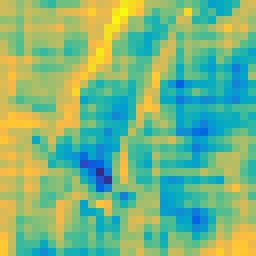}}&
\vcenteredinclude{\includegraphics[width=\mywidth]{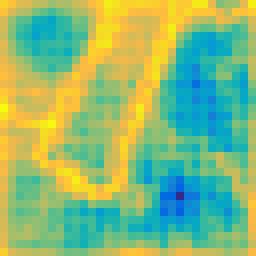}}&&
\vcenteredinclude{\includegraphics[width=\mywidth]{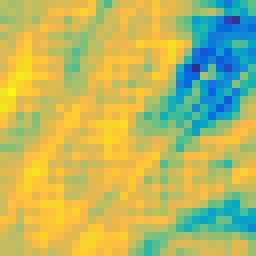}}&
\vcenteredinclude{\includegraphics[width=\mywidth]{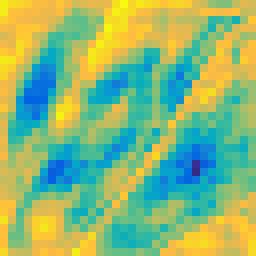}}\\
\vspace{\mywidthbis}\\
\end{tabular}%
} 

\subfloat[Plugging a bicameral decoder to a deeper encoder with DenseNet blocks \citep{densenet17} enables to capture better contextual representations of the image, thus improving the detection of occlusion-aware boundaries.]{%
\begin{tabular}{r|c|c|c|c|c|c|c|c}
\multirow{2}*{\textbf{Architecture}} & \textbf{Encoder} & \textbf{Number of} & \multicolumn{3}{c|}{\textbf{Boundaries}} & \multicolumn{3}{c}{\textbf{Occlusions}}  \\
&\textbf{backbone}&\textbf{parameters}&ODS&AP&AP$_{60}$&ODS&AP&AP$_{60}$\\
 \hline
Two streams (Baseline) & \multirow{2}*{VGG16} & 46,839,938 ($\times1.0$) & .673 & .708 & .476 & .681 & .733 & .518 \\
\cline{1-1}\cline{3-9}
\multirow{2}*{Bicameral decoder} &  & 34,301,250 ($\times.73$) &  .697 & .738 & .517 & .692 & .747 & .532 \\
\cline{2-2}
& DenseNet121 & 33,009,846 ($\times.70$) &  \textbf{.712} & \textbf{.761} & \textbf{.529} & \textbf{.714} & \textbf{.778}  & \textbf{.556} \\
\end{tabular}%
}

\caption{Comparative results for occlusion-aware boundary detection on PIOD and Mikado, using a bicameral structure: with and without skip connections, with different types of skip connections, with different encoder backbones. Best viewed in color.}

\label{fig:resultsskip}
\end{figure*}

\begin{figure*}
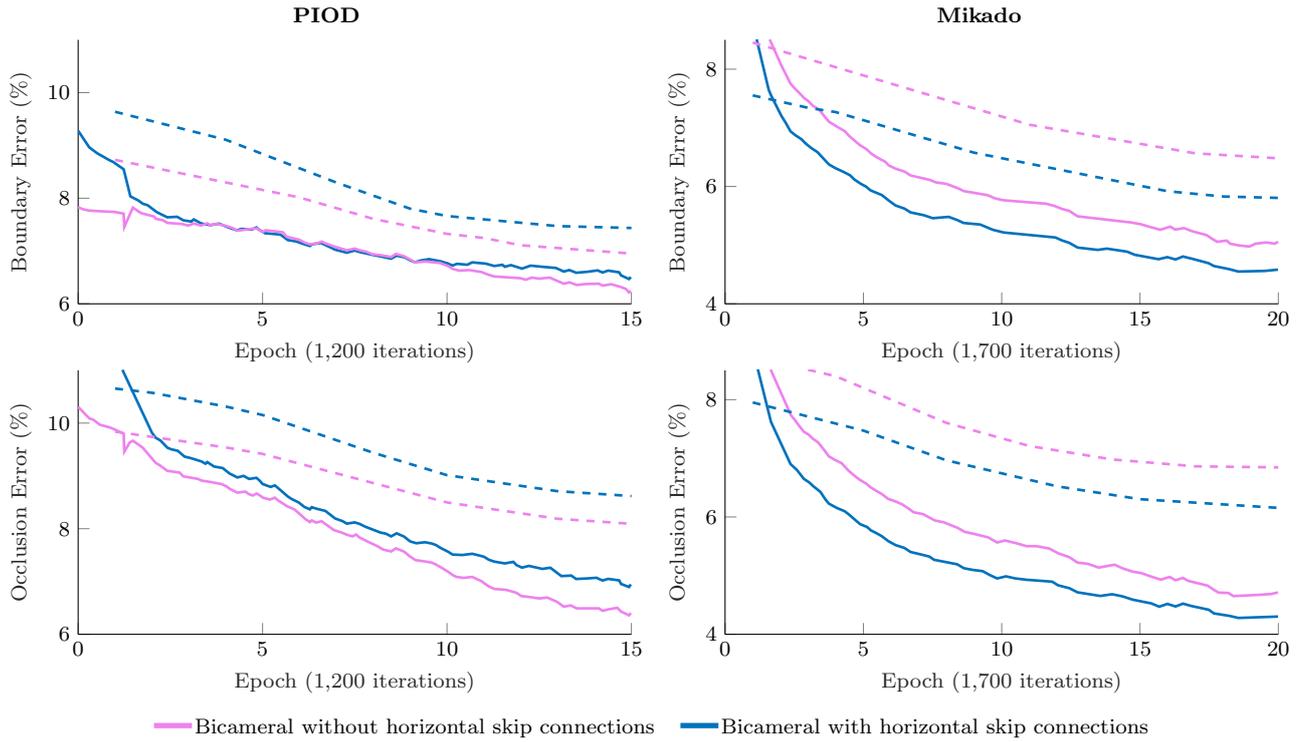

\centering
\fontsize{8}{7}\selectfont
\setlength\fwidth{.44\linewidth}
\setlength\fheight{3.5cm}
\setlength{\tabcolsep}{1pt}
\setlength{\mywidth}{.2cm}
\begin{tabular}{ccc}
 \textbf{PIOD} & &\textbf{Mikado}\\
\vspace{-0.1cm}\\
\end{tabular}

\caption{Training (solid) and test (dashed) errors for instance boundary (top) and occluding boundary side (bottom) detection on PIOD (left) and Mikado (right) using different network architectures. Lower boundary and occlusion errors are reached when jointly learning boundaries and occlusions (green, blue, yellow, purple) rather than independently (red). Best viewed in color.}
\label{fig:resultsskiploss}
\end{figure*}

\begin{figure*}
\centering
\centering

\subfloat{%
\fontsize{7}{8}\selectfont
\setlength{\tabcolsep}{5pt}
\setlength{\thirdcolumn}{.9cm}
%
}

\subfloat{\begin{tabular}{c|c|c|c|c|c|c|c}%
\textbf{Pretraining} & \textbf{Finetuning} & \textbf{Number of real} & \textbf{Frozen encoder} & \multicolumn{2}{c|}{\textbf{Boundaries}} & \multicolumn{2}{c}{\textbf{Occlusions}} \\
\textbf{on Mikado} & \textbf{on D2SA} & \textbf{training images} & \textbf{blocks$^\star$ (in pink)} &ODS&AP&ODS&AP\\
 \hline
\multirow{2}*{No} & \multirow{2}*{Yes} & 438 & \multirow{2}*{None} & .700 & .715 & .725 & .756 \\
 & & Augmented (D2SA+) & & .783 & .792 & .785 & .795 \\
\hline
\multirow{7}*{Yes} & No & 0  & -- & .652 & .649 & .458 & .400 \\
\cline{2-8}
  & \multirow{6}*{Yes} & \multirow{6}*{438} & None & .780 & .808 & .794 & .830 \\
& &  & 1 & .783 & .803 & .797 & .829 \\
& &  & 1, 2 & .780 & .802 & .793 & .827 \\
 &  &  & \textbf{1, 2, 3} & \textbf{.793} & \textbf{.819} & \textbf{.810} & \textbf{.849} \\
& &  & 1, 2, 3, 4 & .759 & .799 & .769 & .819 \\
& &  & 1, 2, 3, 4, 5 & .767 & .815 & .773 & .823 \\
\multicolumn{8}{c}{}\\
\end{tabular}}

\vspace{-.3cm}

\begin{flushleft}
\small$^\star$ A block is a set of convolutional layers between two pooling layers; a VGG16-based encoder is therefore composed of 5 blocks.
\end{flushleft}%

\vspace{-.3cm}

\caption{Comparative performances of a bicameral structure on D2SA using different pretraining conditions. Performances on both boundaries and occlusions are maximized when freezing at finetuning time the first three encoder blocks pretrained on Mikado. Best viewed in color.}
\label{fig:arch-bicameral-frozen}
\end{figure*}

\end{document}